\documentclass[11pt]{article}

\usepackage[final]{acl}

\usepackage{times}
\usepackage{latexsym}
\usepackage{pgfplots}
\pgfplotsset{compat=1.18}
\usepackage{xcolor}

\definecolor{cStd}{HTML}{4C72B0}
\definecolor{cCoT}{HTML}{E8923B}
\definecolor{cBias}{HTML}{C44E52}
\definecolor{cOurs}{HTML}{2CA02C}

\usepackage{subscript}
\usepackage[table]{xcolor}
\usepackage{appendix}
\usepackage[T1]{fontenc}

\usepackage[utf8]{inputenc}

\usepackage{microtype}

\usepackage{inconsolata}

\usepackage{graphicx}
\usepackage{hyperref}
\usepackage{url}
\usepackage{booktabs}
\usepackage{graphicx}
\usepackage{caption}
\usepackage{rotating}
\usepackage{multirow}
\usepackage{enumitem}
\usepackage{wrapfig}
\usepackage{subcaption}
\usepackage[most]{tcolorbox}
\usepackage{amssymb}
\usepackage{extpfeil, extarrows}

\usepackage{makecell}
\usepackage{algorithm}
\usepackage{algpseudocode}
\usepackage{listings}   
\usepackage{xcolor,adjustbox}           

\newcommand{\tabincell}[2]{\makecell[#1]{#2}}

\newcommand{\PopGap}{\textsc{PopGap}}
\newcommand{\HPSR}{\textsc{HPSR}}

\usepackage{textcomp}
\title{Large Language Models Systematically Favor Popular Options: Evidence and Mitigation Across MCQs}

\author{
  \textbf{Abdelrahman Abdallah\textsuperscript{1},
  Mohammed Ali\textsuperscript{1},
  Bhawna Piryani\textsuperscript{1},}\\
 \textbf{Mahmoud Abdalla\textsuperscript{2},
  Adam Jatowt\textsuperscript{1}} \\
  \textsuperscript{1}University of Innsbruck \quad
  \textsuperscript{2}Chungbuk National University\\
  \texttt{\{abdelrahman.abdallah,adam.jatowt\}@uibk.ac.at}
}

\begin{document}
\maketitle

\begin{abstract}
Multiple-choice questions (MCQs) are a standard format for evaluating large language models (LLMs), yet the popularity of answer options can confound evaluation. Modern LLMs systematically prefer popular but incorrect options over less popular correct ones, a vulnerability we call \textbf{popularity bias}. This pattern aligns with confidence miscalibration: model confidence remains high even as accuracy collapses for popular options. To systematically isolate this phenomenon, we introduce \textbf{PopMCQ}, a benchmark with six controlled strategies that vary option popularity while keeping the correct answer fixed. In our most adversarial setting, where all distractors are more popular than the correct option, models choose popular but wrong answers 66\% of the time. To mitigate this bias, we propose \textbf{PopDebias}, a lightweight inference-time correction that estimates and removes a popularity prior from model predictions. It requires no fine-tuning, is label-free at test time (using only a small calibration split for parameter fitting), and adds negligible computational cost. Experiments on 22 open-source LLMs (0.5B to 32B parameters) show consistent improvements, with accuracy gains up to 54.1 percentage points under strong popularity pressure\footnote{The code and data are available \url{https://github.com/DataScienceUIBK/PopMCQ} }.
\end{abstract}


\section{Introduction}
\label{sec:intro}
\begin{figure}[t]
\centering
\includegraphics[width=\linewidth]{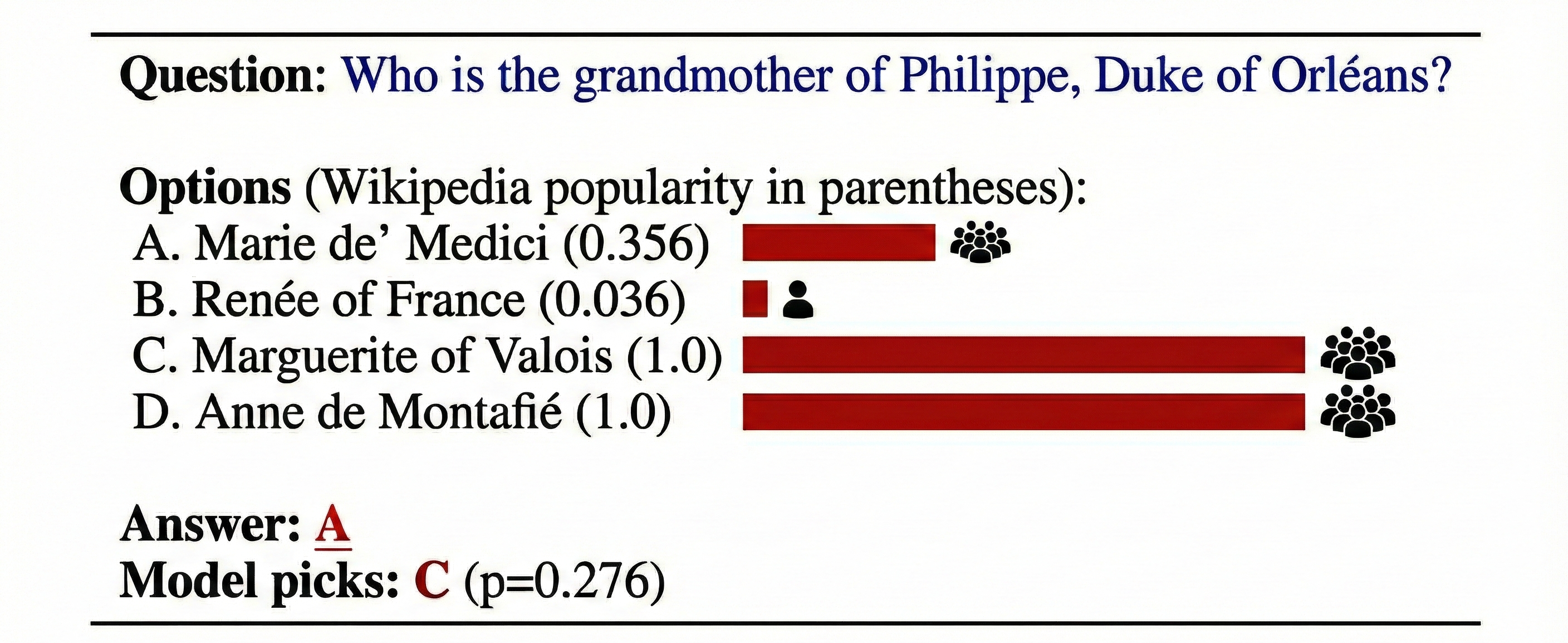}
\caption{
A MuSiQue question where \textbf{DeepSeek-V2-Lite} picks the most popular wrong option (Marguerite of Valois, popularity\,=\,1.00) over the correct but less-famous answer (Marie de' Medici, popularity\,=\,0.356). 
}
\label{fig:mcq_pop_example}
\end{figure}

Large language models (LLMs) recall facts about popular entities more accurately than lesser-known ones~\citep{kandpal2023large,mallen-etal-2023-trust}, where \emph{popularity} refers to an entity's external familiarity as measured by normalized Wikipedia page views (a model-agnostic proxy; see \S\ref{subsec:measurement}). But what happens when popularity and truth \emph{disagree}? In multiple-choice settings, where plausible wrong answers compete with the correct one, this learned tendency becomes a problem: models select well-known but incorrect options over less-known correct answers, a behavior we call \textbf{popularity bias}. Figure~\ref{fig:mcq_pop_example} illustrates this: on a multi-hop question, DeepSeek-V2-Lite selects Marguerite of Valois (popularity$=$1.00; probability$=$0.276) over the correct option, Marie de' Medici (popularity$=$0.356). 

This is systematic: Table~\ref{tab:strategy_sweep_musique_deepseek} shows that changing distractor popularity while keeping questions fixed causes large accuracy swings and strong negative correctness--popularity correlations. Since entity/factual-answer MCQ formats underpin major benchmarks (MMLU~\citep{hendrycks2020measuring}, AGIEval~\citep{zhong2023agieval}, C-Eval~\citep{huang2023c}) and high-stakes applications~\citep{wei2022emergent}, this bias directly threatens evaluation reliability.
Entity frequency correlates with LLM recall accuracy~\citep{kandpal2023large,abdallah2024arabicaqa,mallen-etal-2023-trust}, and dataset imbalance is a known issue in ML~\citep{buda2018systematic}. We study a distinct consequence: when popular-but-incorrect entities appear as MCQ distractors, familiarity actively \emph{competes} with correctness. Unlike retraining-based approaches~\citep{chawla2002smote,he2009learning}, we diagnose and correct this confound \emph{at inference time}, using controlled strategies (S1--S6; \S\ref{sec:strategies}) that isolate popularity while holding questions fixed. 

To make \emph{controlled} claims, we build \textbf{PopMCQ}, a benchmark that varies option popularity while keeping questions and correct answers fixed (\S\ref{sec:data}), with six strategies ranging from maximum popularity pressure to a reverse control. This setup lets us answer three key questions: 

\begin{itemize}
    \item \noindent\textbf{RQ1:} Do LLMs use entity popularity as a shortcut for correctness in MCQ settings?
    \item \noindent\textbf{RQ2:} What mechanisms underlie popularity bias?
    \item \noindent\textbf{RQ3:} Can we mitigate it without retraining?
\end{itemize}
The last question is important because many models are closed-source or deployed, and popularity is sometimes helpful (S5), making na\"ive penalties harmful.

Our experiments on 22 LLMs across four datasets (\S\ref{sec:investigation}) show popularity bias is widespread and predictable: under adversarial conditions, $\rho = -0.89$ between correctness and selected-option popularity, strongly associated with confidence miscalibration. We also propose \textbf{PopDebias} (\S\ref{sec:methodology}), an inference-time method requiring no fine-tuning that treats popularity as a separable prior and removes it. PopDebias improves \emph{all 22 models across all 4 datasets}, with accuracy gains up to 54.1~pp and consistent reductions in high-popularity selection.

\paragraph{Summary of Contributions:}
\begin{enumerate}
    \item We introduce \textbf{PopMCQ}, the first benchmark designed to \emph{systematically} measure popularity bias through controlled interventions in MCQ settings, with six controlled strategies across four QA datasets.
    \item We provide large-scale evidence (22 models $\times$ 4 datasets $\times$ 6 strategies) that LLMs consistently favor popular distractors, and an open-recall diagnostic confirming this is a decision-time phenomenon distinct from lack of knowledge.
    \item We find that popularity bias is strongly associated with confidence miscalibration: the confidence–accuracy gap grows sharply with selected-option popularity.
    \item We propose \textbf{PopDebias}, a training-free method (label-free at test time) that works across all tested settings with very little extra computation.
\end{enumerate}

\section{Dataset and Strategy Construction}
\label{sec:data}
\begin{figure*}[t]
    \centering
    \includegraphics[width=\linewidth]{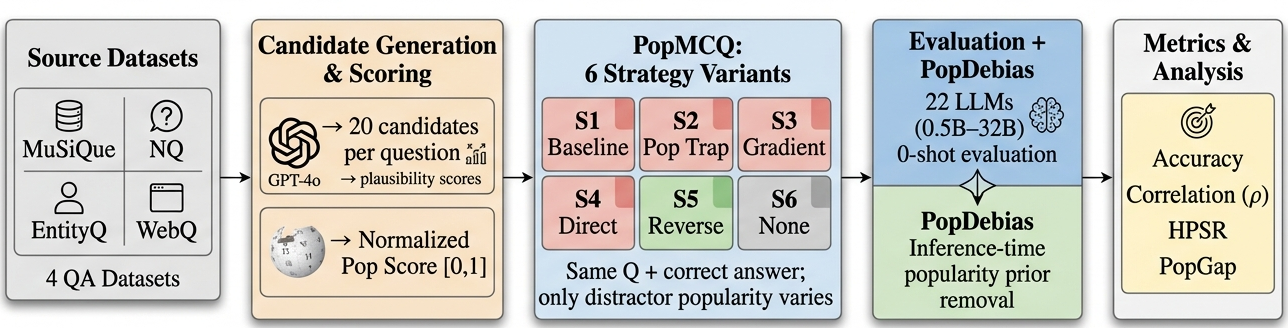}
    \caption{Overview of the PopMCQ construction and evaluation pipeline. 
    We sample questions from four source datasets, generate 20 candidate answers per question using GPT-4o (targeting different popularity levels), score each candidate using Wikipedia page views, and assemble six strategy variants (S1--S6) that differ 
    in distractor popularity. 
    Models are evaluated on all strategies, and PopDebias is applied as a label-free, inference-time correction.
    }
    \label{fig:pipeline}
\end{figure*}

We construct \textbf{PopMCQ}, a benchmark that manipulates option popularity as an independent variable while keeping questions and correct answers fixed. Figure~\ref{fig:pipeline} shows the pipeline.

\subsection{Source data, candidates, and filtering}
\label{sec:candidates}

Existing benchmarks~\citep{talmor2018commonsenseqa,hendrycks2020measuring} leave distractor popularity uncontrolled. We build PopMCQ from four QA datasets: MuSiQue~\citep{trivedi2022musique}, Natural Questions~\citep{kwiatkowski2019natural}, EntityQuestions~\citep{sciavolino2021simple}, and WebQuestions~\citep{berant-etal-2013-semantic}, retaining questions where the gold answer is present, at least four candidates are available, and core strategies (S1, S2, S6) are satisfiable (Appendix~\ref{app:filtering}). Table~\ref{tab:dataset_chars} summarizes correct-answer popularity distributions: EntityQuestions and MuSiQue are long-tailed (median 0.087/0.102; $>$49\% below 0.1), while NQ and WebQuestions shift higher (median 0.170/0.160; $>$0.7 $\approx$ 10\%). Since correct answers are predominantly low-popularity, preference for popular options directly harms accuracy---amplified by S2--S5. 

For each question, we generate 20 candidate incorrect answers using GPT-4o (Appendix~\ref{app:candidate_prompt}), spanning high (6--8), medium (6--8), and low popularity (4--6) tiers, each with a plausibility score (0--100) and justification. The LLM only generates this candidate pool---strategy selection (S1--S6) is \emph{entirely deterministic} based on Wikipedia popularity scores, not LLM judgment, ensuring manipulation is independent of both the candidate-generating and evaluated LLMs. We verify that popularity and plausibility are largely orthogonal (Spearman $\rho < 0.24$; Appendix~\ref{app:pop_plaus}), ruling out plausibility as a confound.

\begin{table}[t]
\centering
\footnotesize
\renewcommand{\arraystretch}{1.12}
\caption{Option-popularity shifts on MuSiQue (\textbf{DeepSeek-V2-Lite}). S1--S4 induce strong correctness--popularity correlations; S5 reverses the effect. S6's negative $\rho$ is structural ($\mathrm{Pop}=0$ for correct); read S6 via accuracy and \HPSR{}. Values show metric and $\Delta$ vs.\ S1.}
\label{tab:strategy_sweep_musique_deepseek}
\scalebox{0.9}{
\begin{tabular}{lcccccc}
\toprule
\textbf{MuSiQue} &
\makecell{\textbf{S1}\\\scriptsize Base} &
\makecell{\textbf{S2}\\\scriptsize Trap} &
\makecell{\textbf{S3}\\\scriptsize Grad} &
\makecell{\textbf{S4}\\\scriptsize Direct} &
\makecell{\textbf{S5}\\\scriptsize Reverse} &
\makecell{\textbf{S6}\\\scriptsize None} \\
\midrule
\textbf{Acc (\%)} 
& \tabincell{c}{31.2 \\ \scriptsize \textcolor{gray}{(0.0)}}
& \tabincell{c}{31.1 \\ \scriptsize \textcolor{blue}{(-0.1)}}
& \tabincell{c}{33.9 \\ \scriptsize \textcolor{red}{(+2.6)}}
& \tabincell{c}{27.3 \\ \scriptsize \textcolor{blue}{(-3.9)}}
& \tabincell{c}{30.9 \\ \scriptsize \textcolor{blue}{(-0.3)}}
& \tabincell{c}{68.6 \\ \scriptsize \textcolor{red}{(+37.4)}} \\
\textbf{Corr ($\rho$)} 
& \tabincell{c}{-0.23 \\ \scriptsize \textcolor{gray}{(0.00)}}
& \tabincell{c}{-0.89 \\ \scriptsize \textcolor{blue}{(-0.66)}}
& \tabincell{c}{-0.66 \\ \scriptsize \textcolor{blue}{(-0.42)}}
& \tabincell{c}{-0.49 \\ \scriptsize \textcolor{blue}{(-0.26)}}
& \tabincell{c}{\phantom{-}0.19 \\ \scriptsize \textcolor{red}{(+0.42)}}
& \tabincell{c}{-0.95 \\ \scriptsize \textcolor{blue}{(-0.72)}} \\
\textbf{HPSR (\%)} 
& \tabincell{c}{49.9 \\ \scriptsize \textcolor{gray}{(0.0)}}
& \tabincell{c}{66.0 \\ \scriptsize \textcolor{red}{(+16.1)}}
& \tabincell{c}{51.4 \\ \scriptsize \textcolor{red}{(+1.5)}}
& \tabincell{c}{55.7 \\ \scriptsize \textcolor{red}{(+5.8)}}
& \tabincell{c}{50.9 \\ \scriptsize \textcolor{red}{(+1.0)}}
& \tabincell{c}{22.2 \\ \scriptsize \textcolor{blue}{(-27.7)}} \\
\bottomrule
\end{tabular}}
\end{table}

\begin{table}[t]
\centering
\small
\setlength{\tabcolsep}{3.5pt}
\caption{Correct-answer popularity statistics across datasets. Med.\ = median; IQR = interquartile range; \%{<}0.1 and \%{>}0.7 = fraction in low- and high-popularity tails.}
\label{tab:dataset_chars}
\scalebox{0.8}{
\begin{tabular}{lrcccccc}
\toprule
\textbf{Dataset} & \textbf{N} & \textbf{Med.} & \textbf{Mean} & \textbf{IQR} & \textbf{Skew} & \textbf{\%{<}0.1} & \textbf{\%{>}0.7} \\
\midrule
EntityQuestions & 9584 & 0.087 & 0.201 & 0.279 & 1.47 & 52.9 & 6.8 \\
NQ             & 1050 & 0.170 & 0.281 & 0.449 & 0.91 & 37.7 & 10.0 \\
MuSiQue        &  874 & 0.102 & 0.204 & 0.258 & 1.36 & 49.8 & 4.2 \\
WebQuestions   &  702 & 0.160 & 0.265 & 0.388 & 1.05 & 38.6 & 10.3 \\
\bottomrule
\end{tabular} }
\end{table}

\begin{figure*}[t]
  \centering
  \includegraphics[width=\linewidth]{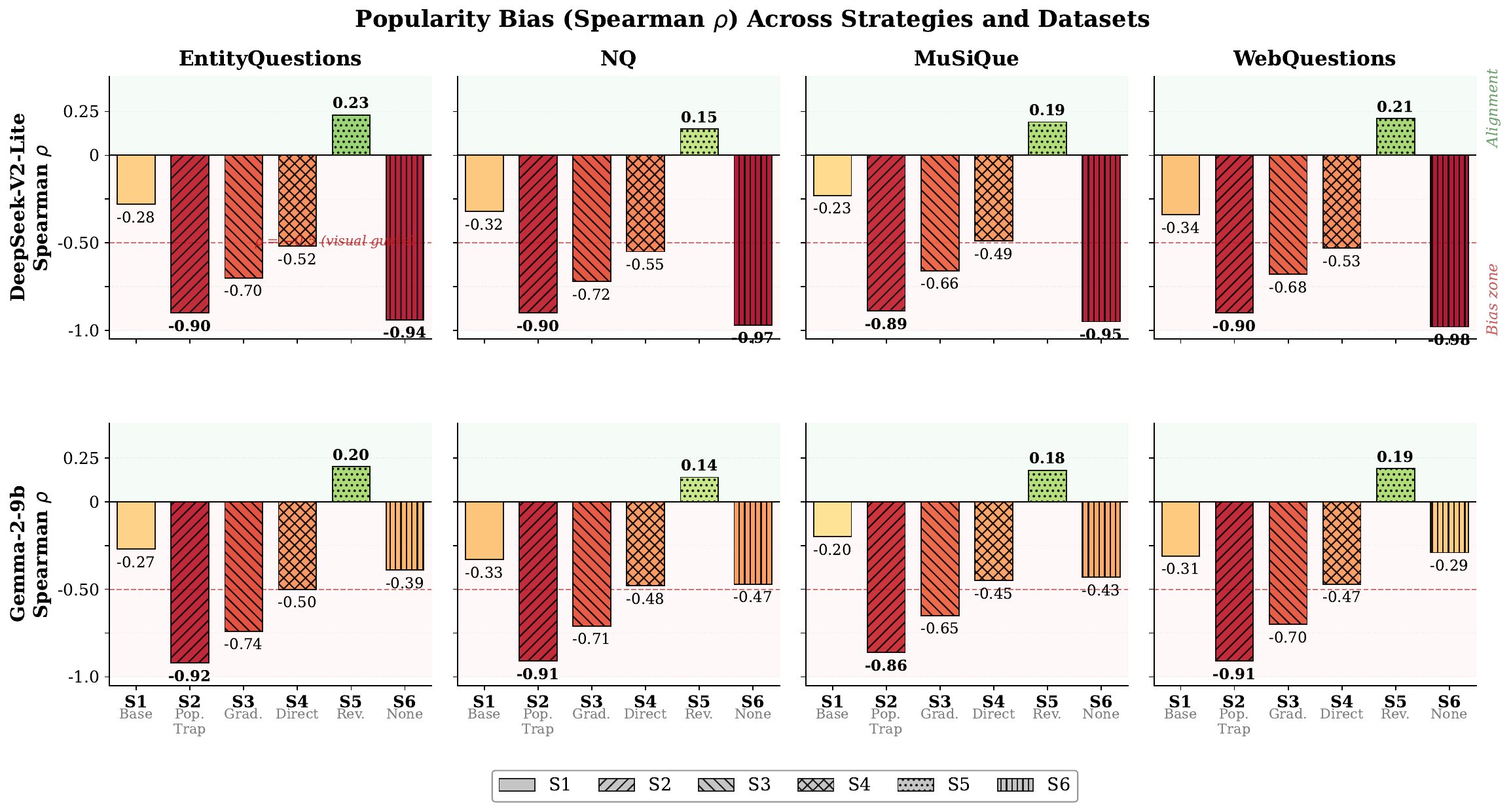}
  \caption{Popularity bias (Spearman $\rho$ between correctness and selected-option popularity) across strategies and datasets. Negative values (red) indicate bias; positive values (green) indicate alignment. The dashed line at $\rho = -0.5$ is a conventional moderate-to-strong correlation reference used only as a visual guide; it is not a learned or data-dependent threshold. S2 (Popular Trap) shows strongest bias, while S5 (Reverse Control) flips positive. Full results in Appendix~\ref{sec:full_results_for_Result}.}

  \label{fig:popularity_bias_main}
\end{figure*}

\subsection{Strategy generation}
\label{sec:strategies}

Each base question yields six MCQ variants that differ \emph{only} in distractor popularity. All MCQs have four options (one correct, three distractors) with positions randomized over \{A,B,C,D\}. Let $\mathcal{C}(q)$ be the candidate pool and $\mathrm{pop}(a^\star)$ the correct answer's popularity. The six strategies are:

\textbf{S1 (Baseline):} 3 random distractors, no popularity constraints.

\textbf{S2 (Popular Trap):} top-3 most popular candidates with $\mathrm{pop} > \mathrm{pop}(a^\star)$, maximizing popularity pressure.

\textbf{S3 (Gradient):} 3 distractors with strictly decreasing $\mathrm{pop}$, all above $\mathrm{pop}(a^\star)$.

\textbf{S4 (Direct Contest):} 1 highest-$\mathrm{pop}$ candidate $+$ 2 medium-$\mathrm{pop}$ distractors.

\textbf{S5 (Reverse Control):} 3 distractors with $\mathrm{pop} < \mathrm{pop}(a^\star)$, so the correct answer is most popular.

\textbf{S6 (None of the Above):} correct option is ``None of the above'' ($\mathrm{pop}=0$) while distractors are $1{\times}$ highest-$\mathrm{pop}$ $+$ $2{\times}$ lowest-$\mathrm{pop}$ candidates, stress-testing abstention under popularity pressure.

Items where S1, S2, or S6 cannot be constructed are removed. Full strategy descriptions, design rationale, and a detailed comparison of S2 vs.\ S3 are in Appendix~\ref{app:strategy_details}.

\section{Investigation on Popularity Bias}
\label{sec:investigation}

\subsection{Experimental Setup}
\label{subsec:setup}
We evaluate 22 decoder-only LLMs (0.5B--32B): Llama-2/3/3.1/3.2~\citep{grattafiori2024llama,touvron2023llama}, Qwen2.5~\citep{qwen2024qwen25}, Gemma-2/3~\citep{team2024Gemma,team2025Gemma}, Falcon-7B~\citep{almazrouei2023Falcon}, Mistral-7B~\citep{jiang2023mistral}, Phi-1.5/2/4~\citep{abdin2024phi} (Phi-3.5 was outside our sweep; the Phi family spans 1.3B--14B), DeepSeek-V2-Lite, and Zephyr-7B~\citep{tunstall2023zephyr}, using 0-shot max-probability selection over \texttt{A/B/C/D} tokens following standard MCQ protocols~\citep{hendrycks2020measuring,clark2018think,zheng2023large}. Since all strategies share the same scoring, cross-strategy differences (e.g., $\rho = -0.89$ under S2 vs.\ $+0.19$ under S5) cannot be attributed to measurement artifacts. Scoring robustness is verified in Appendix~\ref{app:scoring_robustness}; evaluation details in Appendix~\ref{app:eval_prompts}.

\subsection{Measuring Popularity Bias}
\label{subsec:measurement}

We define \emph{popularity bias} as a model's tendency to select popular options over less-popular correct ones. Each option $o$ has a popularity score $\mathrm{Pop}(o) \in [0,1]$ from Wikipedia (\S\ref{app:wiki_scoring}), validated against model-internal familiarity signals (mean Spearman $\rho = 0.70$ across all 22 models; Appendix~\ref{app:proxy}). We use three main metrics: 
\textbf{Correlation ($\rho$):} Spearman correlation between correctness and selected-option popularity. Negative $\rho$ means selecting popular options is associated with being wrong. \textbf{High-Popularity Selection Rate (HPSR):} Fraction of times the model picks an option in the top half of the popularity ranking. HPSR $>$ 50\% indicates a pull toward popular options. \textbf{Popularity Gap (\PopGap):} The average difference between the popularity of the selected option and the correct option: $\PopGap = \mathbb{E}[\mathrm{Pop}(\hat{o}) - \mathrm{Pop}(o^\star)]$, where $\hat{o}$ is the model's selected option and $o^\star$ is the correct option. Positive values indicate the model systematically selects options more popular than the truth. \textbf{Accuracy (Acc):} Proportion of correct predictions. We also measure \textbf{Confidence} (average probability for chosen option) and \textbf{Overconfidence by Popularity Bucket} to test whether familiarity drives miscalibration (Fig.~\ref{fig:confidence_analysis}). Full metric definitions are in Appendix~\ref{app:metrics}.

\subsection{Key Observations}
\label{subsec:observations}
Figure~\ref{fig:popularity_bias_main} shows results for two representative models across all datasets and strategies. The pattern is consistent: under popularity pressure (S2--S4), correlations are strongly negative; under reverse control (S5), they flip positive. Table~\ref{tab:musique_key_results_summary} reports MuSiQue accuracies for four models; full results in Appendix~\ref{sec:full_results_for_Result}.

\begin{table}[t]
\centering

\small
\setlength{\tabcolsep}{3pt}
\caption{Accuracy (\%) on \textbf{PopMCQ instances constructed
from MuSiQue questions}, with distractors generated by GPT-4o
and assembled into six popularity-controlled strategy variants
(S1--S6), for representative models.}
\label{tab:musique_key_results_summary}
\scalebox{0.75}{
\begin{tabular}{l|cccc}
\toprule
\textbf{Strategy} & \textbf{Llama-3-8B} & \textbf{Qwen2.5-7B} & \textbf{DeepSeek-V2-Lite} & \textbf{Gemma-2-9b} \\
\midrule
\textbf{S1: Baseline} & 44.9 & 41.0 & 31.2 & 29.3  \\
\textbf{S2: Pop Trap} & 45.5 & 42.0 & 31.1 & 33.6  \\
\textbf{S3: Gradient} & 43.1 & 40.8 & 33.9 & 32.0  \\
\textbf{S4: Direct} & 40.3 & 34.8 & 27.3 & 26.1 \\
\textbf{S5: Reverse} & 44.9 & 41.8 & 30.9 & 31.0 \\
\textbf{S6: None} & 23.2 & 3.2 & 68.6 & 6.5 \\
\bottomrule
\end{tabular}}
\end{table}

\paragraph{Systematic negative correlations under popularity pressure.}
For MuSiQue, averaged over all 22 models, dataset-level mean $\rho$ values are $-0.234$ (S1), $-0.864$ (S2), $-0.704$ (S3), $-0.520$ (S4), $+0.202$ (S5), $-0.449$ (S6). S6's negative $\rho$ is expected by construction ($\mathrm{Pop}=0$ for ``None of the above''); we interpret S6 via accuracy and \HPSR{}. Under S2/S3, Llama-3-8B and Qwen2.5-7B show modest accuracy changes (44.9$\to$45.5/43.1 and 41.0$\to$42.0/40.8), while Gemma-2-9b improves (29.3$\to$33.6/32.0)---accuracy can rise under S2 because globally famous but question-agnostic distractors may be easier to eliminate than locally plausible ones sampled in S1 (Appendix~\ref{app:acc_increase})---but collapses on S6 (6.5\%). DeepSeek-V2-Lite dramatically improves under S6 (68.6\%). \HPSR{} complements $\rho$ by measuring raw selection frequency: in S1 it is near chance ($\approx$50\%); in S2 it rises to 66.0\% on MuSiQue; in S5 it returns to baseline.

\paragraph{Reverse control confirms the effect.}
S5 reliably yields positive correlations (MuSiQue mean $+0.202$) with slight accuracy gains (Llama-3.1-8B: 42.8\%$\to$43.2\%; Qwen2.5-7B: 41.0\%$\to$41.8\%) and small \PopGap{} ($+0.026$). Cross-dataset differences reflect underlying popularity distributions (Table~\ref{tab:dataset_chars}): EntityQuestions and MuSiQue (median $<$ 0.11) show greater susceptibility, while NQ and WebQuestions (IQR $>$ 0.38, $>$0.7 $\approx$ 10\%) exhibit attenuated effects.

\begin{figure}
  \centering
  \includegraphics[width=0.9\columnwidth]{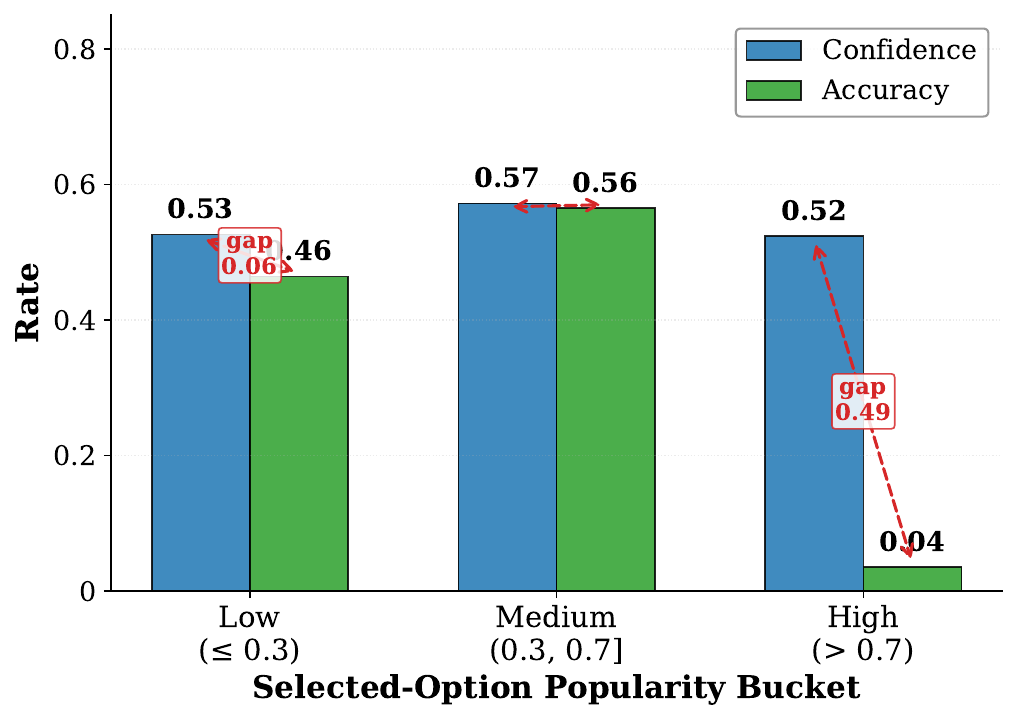}
  \caption{\textbf{Confidence vs.\ accuracy by popularity.} Popularity buckets: \emph{Low} $\le 0.3$, \emph{Medium} $(0.3,0.7]$, \emph{High} $>0.7$.}
  \label{fig:confidence_analysis}
\end{figure}

\paragraph{Popularity bias reflects decision-time attraction, not just lack of knowledge.}
The controlled design (same question and gold answer across S1--S5) already shows that distractor popularity, not question difficulty, drives the effect. To further separate popularity-driven errors from knowledge gaps, we conduct an \textbf{open-recall diagnostic}: each model first answers without any MCQ options (free-form generation), and we verify correctness via normalized exact/alias matching. We then evaluate S2 only on items the same model answered correctly in open recall---a conservative ``known-answer'' subset. Table~\ref{tab:open_recall} reports results on MuSiQue for five representative models. Even on this known-answer subset, S2 accuracy drops sharply and 99.8\% of S2 errors select a \emph{more-popular} distractor than the correct answer. Additionally, about 29\% of all S2 errors across 22 models occur on items that the same model answers correctly under S5 (reverse control), directly confirming that a substantial fraction of S2 failures cannot be explained by lack of knowledge (full per-dataset breakdown in Appendix~\ref{app:cross_strategy_flips}).

\begin{table}[t]
\centering
\small
\setlength{\tabcolsep}{3pt}
\caption{\textbf{Open-recall diagnostic on MuSiQue (S2).} Models first answer without options; S2 is then evaluated only on items answered correctly in open recall. ``W$\to$MP'' = fraction of S2 errors selecting a more-popular option than the truth.}
\label{tab:open_recall}
\scalebox{0.8}{
\begin{tabular}{lrccc}
\toprule
\textbf{Model} & \textbf{Open Acc} & \textbf{Known $N$} & \textbf{S2 Acc} & \textbf{W$\to$MP} \\
\midrule
Llama-3-8B      & 38.5\% & 336 & 61.0\% & 99.6\% \\
Qwen2.5-7B      & 35.8\% & 313 & 56.5\% & 99.8\% \\
Gemma-2-9B      & 26.2\% & 229 & 33.5\% & 100.0\% \\
Phi-4            & 45.0\% & 393 & 65.0\% & 99.5\% \\
DeepSeek-V2-Lite & 29.5\% & 258 & 42.0\% & 100.0\% \\
\midrule
Pooled           & 34.9\% & 1{,}529 & 52.0\% & 99.8\% \\
\bottomrule
\end{tabular}}
\end{table}

\paragraph{S6 exposes a failure to abstain under uncertainty.}
Under S6, models rarely choose ``None of the Above'' even when it is correct, preferring
popular named entities. For \textbf{Qwen2.5-7B} on MuSiQue, S6 accuracy is only
3.2\% while high-pop selection rate jumps to 67.5\% and \textsc{PopGap} inflates to
$+0.476$. Alignment simultaneously degrades (confidence: 0.53, alignment: 0.47),
revealing confident-but-wrong behavior when uncertainty should be acknowledged.

\subsection{What Contributes to Popularity Bias?}
\label{subsec:causes}

\paragraph{Confidence miscalibration linked to entity popularity.}
Confidence remains high and poorly discriminative for high-popularity selections, while accuracy drops sharply. As shown
in Figure~\ref{fig:confidence_analysis}, where results are aggregated across models,
confidence is approximately 0.54--0.59 across \emph{Low}, \emph{Medium}, and \emph{High} popularity buckets, while accuracy collapses from \textbf{0.45} (\emph{Low}) to \textbf{0.03} (\emph{High}). The resulting confidence--accuracy gap therefore grows sharply from \emph{Low} to \emph{High} popularity, quantifying a 
strong familiarity–certainty association that persists despite decreasing accuracy (full bucket-level data in Appendix~\ref{app:confidence_gap}).

\begin{figure}
  \centering
  \includegraphics[width=0.9\linewidth]{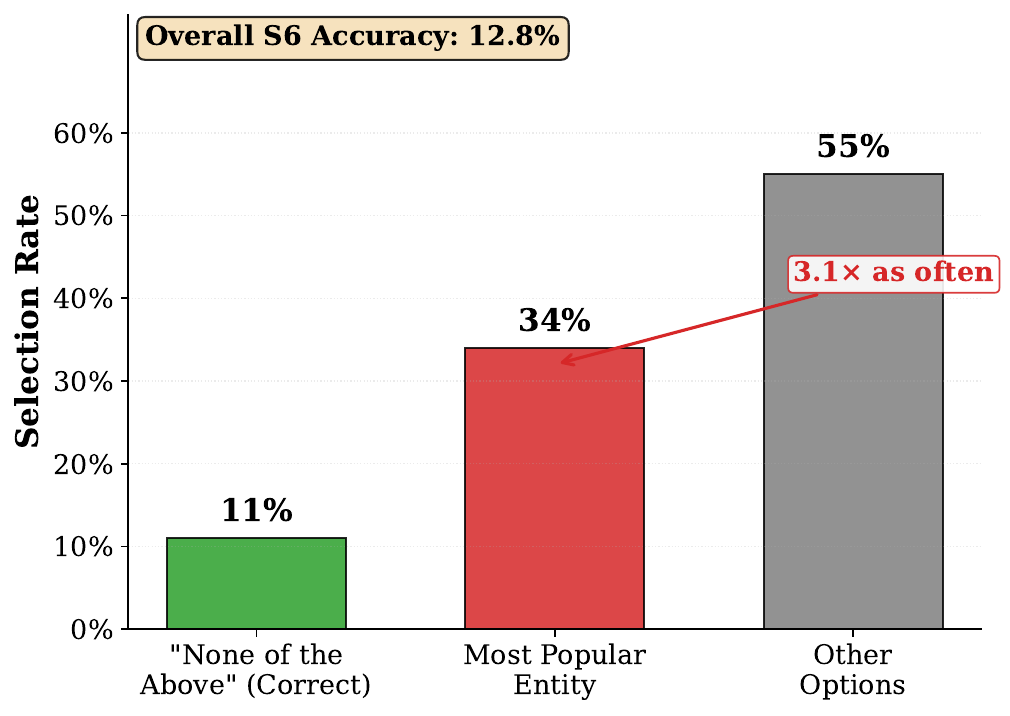}
  \caption{\textbf{S6 uncertainty handling.} Bars show selection rates; the inset reports overall accuracy. Models overwhelmingly prefer popular named entities over the correct ``None of the Above'' option.}
  \label{fig:s6_uncertainty}
\end{figure}

\paragraph{Inadequate uncertainty handling.}
When uncertainty is required (S6: \emph{None of the Above}), models overwhelmingly 
choose named entities instead. As shown in Figure~\ref{fig:s6_uncertainty}, the 
correct option is selected only \textbf{11\%} of the time, while the most popular 
entity is chosen \textbf{34\%} of the time (\textbf{3.1}$\times$ as often), showing 
that preference for popular names overwhelms appropriate uncertainty expression. 
This failure is not limited to S6: across all strategies and all model families, 
the pull toward popular entities is consistent, with S2 correlations ranging from 
$-0.76$ to $-0.95$ across families (see Appendix~\ref{app:family_bias}).

\begin{figure*}[t]
    \centering
    \includegraphics[width=0.9\linewidth]{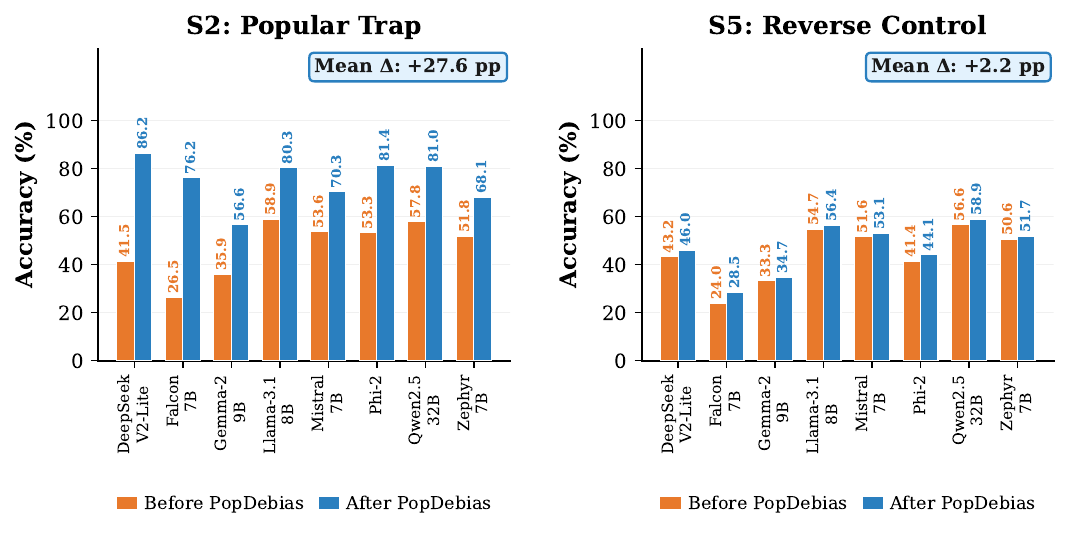}
    \caption{\textbf{PopDebias accuracy gains per model},
    averaged across four datasets. \textbf{Top:} Under S2
    (Popular Trap), PopDebias delivers +16 to +50~pp gains
    across the eight models shown (mean +27.6~pp). \textbf{Bottom:}
    S5 (Reverse Control), gains are minimal (+1 to +4~pp),
    confirming that PopDebias adaptively corrects only when
    popularity conflicts with truth.}
    \label{fig:grouped_bar_debiasing}
\end{figure*}

\subsection{Does Popularity Bias Diminish at Scale?}
\label{subsec:scaling}

\paragraph{Scaling within families.}
Within the Qwen2.5 family (0.5B--32B, 64$\times$ range), accuracy improves with scale but bias does not diminish: $\rho \in [-0.82, -0.89]$ and HPSR $> 55\%$ across all sizes (Figure~\ref{fig:scaling_analysis}; Appendix~\ref{app:scaling}). The Llama family shows the same pattern: Llama-2-7B, Llama-3-8B, and Llama-3.1-8B all yield $\rho = -0.87$ under S2 despite architectural improvements. RLHF does not help either: Zephyr-7B (RLHF-tuned from Mistral-7B) shows $\rho = -0.87$, identical to Mistral-7B's $\rho = -0.87$. We also evaluate frontier (GPT-4o, Claude-4.5-Sonnet, Gemini-2.5-Pro) and reasoning models (o1-mini, QwQ-32B) via API: all show persistent bias under S2 ($\rho = -0.41$ to $-0.65$), with reasoning models most attenuated but not eliminated (Table~\ref{tab:frontier_results})

\paragraph{Frontier and reasoning models.}
We evaluate GPT-4o~\citep{openai2024gpt4o}, Claude-4.5-Sonnet~\citep{anthropic2024claude}, Gemini-2.5-Pro~\citep{google2024gemini}, o1-mini, and QwQ-32B on MuSiQue (S1, S2, S5; $N = 874$).\footnote{For API models without token probabilities, we use verbatim letter responses; $\rho$ is computed from binary correctness and selected-option popularity.} Table~\ref{tab:frontier_results} reports results. Frontier models achieve higher S1 accuracy (62.8--70.1\%) but still show negative $\rho$ under S2 ($-0.52$ to $-0.65$) with elevated HPSR (47.3--52.1\%). Reasoning models show the most attenuation: o1-mini achieves $|\rho| = 0.41$ with the smallest accuracy drop ($70.1\% \to 64.5\%$, $-5.6$~pp). However, even o1-mini exhibits clearly negative $\rho = -0.41$, confirming that reasoning capabilities \emph{reduce} but do not \emph{eliminate} popularity bias. Note that PopDebias requires token-level probabilities unavailable from most closed-source APIs; the bias \emph{measurement} (accuracy, HPSR, $\rho$) is fully applicable from final-answer choices alone.

\begin{table}[t]
\centering
\setlength{\tabcolsep}{4pt}
\renewcommand{\arraystretch}{1.10}
\caption{\textbf{Frontier and reasoning model evaluation on MuSiQue.}
Even state-of-the-art models exhibit popularity bias under S2,
though at reduced magnitude. Reasoning models show the most attenuation.}
\label{tab:frontier_results}
\scalebox{0.55}{
\begin{tabular}{l|ccc|ccc|ccc}
\toprule
& \multicolumn{3}{c|}{\textbf{S1 (Baseline)}}
& \multicolumn{3}{c|}{\textbf{S2 (Pop.\ Trap)}}
& \multicolumn{3}{c}{\textbf{S5 (Reverse)}} \\
\cmidrule(lr){2-4} \cmidrule(lr){5-7} \cmidrule(lr){8-10}
\textbf{Model}
& \textbf{Acc} & \textbf{$\rho$} & \textbf{HPSR}
& \textbf{Acc} & \textbf{$\rho$} & \textbf{HPSR}
& \textbf{Acc} & \textbf{$\rho$} & \textbf{HPSR} \\
\midrule
GPT-4o           & 68.4 & -0.17 & 41.8 & 63.2 & -0.52 & 47.3 & 70.1 & +0.21 & 55.2 \\
Claude-4.5-Son.  & 65.1 & -0.19 & 43.2 & 58.7 & -0.61 & 50.5 & 67.3 & +0.20 & 54.8 \\
Gemini-2.5-Pro   & 62.8 & -0.21 & 44.1 & 55.4 & -0.65 & 52.1 & 64.6 & +0.18 & 53.6 \\
\midrule
o1-mini          & 70.1 & -0.16 & 40.2 & 64.5 & -0.41 & 44.8 & 71.8 & +0.25 & 56.1 \\
QwQ-32B          & 66.8 & -0.18 & 42.5 & 60.3 & -0.55 & 49.1 & 68.5 & +0.19 & 54.2 \\
\bottomrule
\end{tabular}}
\end{table}

\section{Debiasing Method}
\label{sec:methodology}

\subsection{Problem Setup}
We consider an MCQ instance with options $o_1,\dots,o_K$ ($K\!\in\!\{4\}$).
A base model outputs a probability vector $\mathbf{p}\!=\!(p_1,\dots,p_K)$ and each
option is annotated with a scalar popularity score $\mathrm{Pop}(o_i)\in[0,1]$
(Section~\ref{subsec:measurement}). Our objective is to transform $\mathbf{p}$ into a
debiased distribution $\widetilde{\mathbf{p}}$ that reduces the tendency to over-select
common entities while retaining signal that aligns popularity and truth (S5).
\paragraph{Notation.}
We write $\mathrm{rank}_{\text{pop}}(o_i)$ for the within-item popularity rank
($1$\,=\,most popular), $\bar{\mathrm{Pop}}$ and $\mathrm{Var}(\mathrm{Pop})$ for the
mean and variance of per-item popularities, respectively. Throughout the paper, $\varepsilon$
denotes a small constant for numerical stability, and $\mathrm{softmax}$ performs
elementwise exponentiation and normalization.

\subsection{Popularity as a Prior}
Following the decomposition view used for selection biases in MCQs (e.g., position/token
bias) \citep{zheng2023large}, we treat popularity as a \emph{probability prior} that skews
the observed distribution:
\begin{equation}
\begin{split}
\label{eq:bayes_divide}
p_i \ \propto\ \underbrace{\pi_i}_{\text{popularity prior, $\uparrow$ in Pop}}
\cdot
\underbrace{\phi_i}_{\text{popularity-free belief}} \\
\quad\Rightarrow\quad
\phi_i\ \propto\ \frac{p_i}{\pi_i}.
\end{split}
\end{equation}
Our method estimates $\pi$ from \emph{observable} patterns in the current
item, divides it out, then tempers confidence to prevent overcorrection.
Unlike permutation averaging, it operates in a single pass over options.

\subsection{\textbf{PopDebias} (Popularity-Prior)}
\label{subsec:pop_prior}
We estimate a nonnegative \emph{bias strength} $b$ from features that are available at
inference time:
\begin{equation}
\begin{split}
\mathbf{f} \;=\;
\big[\ \mathrm{Var}(\mathrm{Pop}),\ \max_i \mathrm{Pop}(o_i),\ \mathrm{Pop}(o_{\hat{\imath}}),\ \\
\mathrm{rank}_{\text{pop}}(o_{\hat{\imath}}),\ \max_i p_i\ \big],
\end{split}
\end{equation}
where $\hat{\imath}=\arg\max_i p_i$ is the model’s top prediction. Intuitively, bias
should rise when the model is very confident on a very common option and the popularity
distribution has low variance (i.e., distractors cluster in popularity).
\paragraph{Bias estimation.}
On a small calibration split, we fit a lightweight regressor $f_\theta$ to predict the
true \emph{popularity gap} between the model’s choice and the correct answer:

\begin{equation}
\label{eq:calibration}
\begin{split}
\theta \;\leftarrow\; \arg\min_\theta \sum_{(q,x)\in\mathcal{D}_{\mathrm{cal}}}
\big\| &f_\theta(\mathbf{f}(q,x)) \\
&- \big(\mathrm{Pop}(o_{\hat{\imath}})-\mathrm{Pop}(o^\star)\big) \big\|^2.
\end{split}
\end{equation}

At test time (no labels), we set $b=f_\theta(\mathbf{f})$. If a regressor is not
available, we use a bounded heuristic scaled by a calibration constant $\bar{b}$:
\begin{equation}
\label{eq:b_heur}
\begin{split}
b \ \propto\ 
\frac{\mathrm{Pop}(o_{\hat{\imath}})}{\bar{\mathrm{Pop}}+\varepsilon}
\;+\; \max_i p_i
\;+\; \frac{1}{\mathrm{Var}(\mathrm{Pop})+\varepsilon}, \\
\qquad b \leftarrow \bar{b}\,\mathrm{clip}(b;\,0,\infty).
\end{split}
\end{equation}

\paragraph{Constructing the popularity prior and debiasing.}
Given $b$, we build a within-item prior that increases with popularity and apply a
single-pass reweighting with \emph{adaptive} confidence tempering:
\begin{align}
\label{eq:pp_rule}
w_i &= 1 - b \cdot \frac{\mathrm{Pop}(o_i)}{\max_j \mathrm{Pop}(o_j)+\varepsilon}, \\
q_i &= p_i\,w_i \Big/ \textstyle\sum_j p_j\,w_j, \nonumber \\
\gamma &= \mathrm{clip}\!\big(1 - \beta\, b\, c\, \psi,\ 0,\ 1\big), \nonumber \\
\widetilde{p}_i &= \gamma\,q_i + \tfrac{1-\gamma}{K}, \nonumber
\end{align}

i.e., we first reweight and renormalize ($q$), then temper toward uniform, so
$\widetilde{\mathbf{p}}$ is already a distribution.
Here $w_i$ implements $\pi_i^{-1}$ up to a bounded linearization that avoids extreme
division in small-option regimes. The tempering coefficient $\gamma$ depends on (i) the
estimated bias $b$, (ii) model confidence $c\!=\!\max_i p_i$, and (iii) a \emph{popularity
pressure} factor $\psi$ defined from the entropy of the normalized popularity profile
$r_i\!\propto\!\mathrm{Pop}(o_i)$:
\begin{equation}
\label{eq:psi}
\psi \;=\; 1 - \frac{H(r)}{\log K},\qquad
H(r)\!=\!-\sum_{i=1}^K r_i \log(r_i+\varepsilon).
\end{equation}
$\psi$ increases when one or two options dominate in popularity (low entropy), indicating a
sharper ``popular trap.'' The scalar $\beta\!>\!0$ controls the overall tempering strength
(we use $\beta = 1$ throughout; results are insensitive to this choice). When $b$ is small or the popularity
profile is flat, $\gamma\!\approx\!1$ and $\widetilde{\mathbf{p}}\!\approx\!\mathbf{p}$.

We fit $f_\theta$ and $\bar{b}$ (Eq.~\ref{eq:b_heur}) on a small calibration split (default 10\%; dozens--hundreds of samples suffice), and fix $\beta = 1$ (Eq.~\ref{eq:pp_rule}). At test time, inference is \emph{label-free}---no gold answers are used---and requires a single forward pass with $O(K)$ per-item overhead, unlike permutation-averaging~\citep{zheng2023large} which needs $K!$ passes. We clip $w_i \in [0,1]$, $\gamma \in [0,1]$, and use $\varepsilon = 10^{-12}$. Full pseudocode is given in Algorithm~\ref{alg:app} (Appendix~\ref{app:algorithm}); step-by-step example is shown in Appendix~\ref{app:worked-example}.

\begin{figure}[t]
    \centering
    \includegraphics[width=0.95\columnwidth]{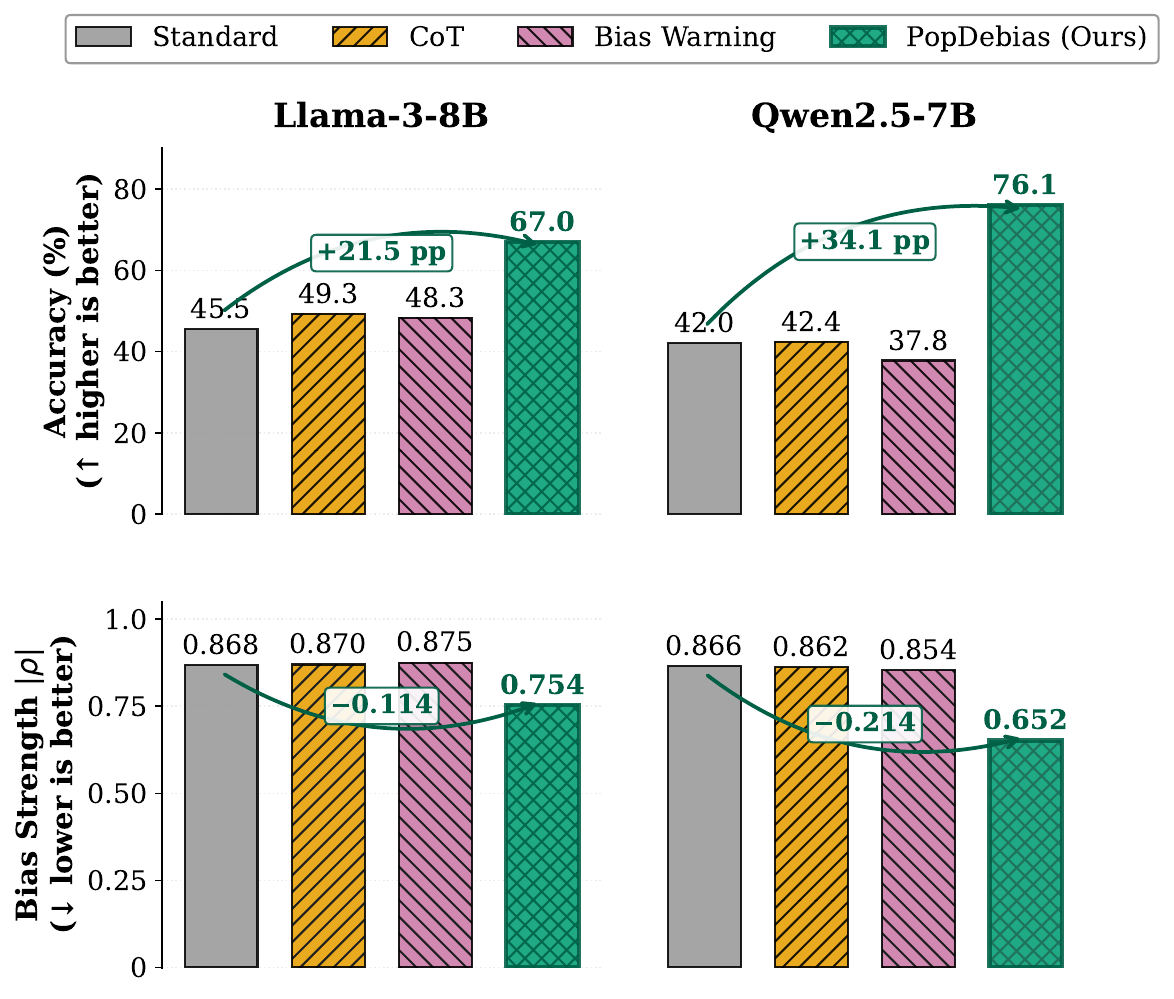}
    \caption{\textbf{Prompting technique comparison on S2 (Popular Trap).} 
    Each column shows one model. Top row: Accuracy (higher is better). 
    Bottom row: Bias strength measured by correlation magnitude $|\rho|$ 
    (lower is better, shown inverted). 
    }
    \label{fig:prompting_comparison}
\end{figure}

\section{Main Results}
\subsection{Debiasing Results}
\label{sec:comprehensive_results}
Figure~\ref{fig:grouped_bar_debiasing} summarizes PopDebias performance on the two most informative strategies, averaged across four datasets. Under \textbf{S2 (Popular Trap)}, PopDebias delivers dramatic accuracy gains for every model (mean +27.6~pp over the eight models shown), with Falcon-7B improving from 26.5\% to 76.2\% and Phi-2 from 53.3\% to 81.4\%. Under \textbf{S5 (Reverse Control)}, gains are minimal (+2.2~pp), confirming that PopDebias adaptively corrects only when popularity conflicts with truth. Gains under S1, S3, S4, and S6 are consistent (2--40~pp); aggregate strategy-level results are in Appendix~\ref{app:aggregate_debias} and full per-model results in Appendix~\ref{sec:full_results_for_Result} and~\ref{app:full_results}.

\subsection{Confidence Calibration Analysis}
\label{sec:calibration}
\begin{figure}[t]
    \centering
    \begin{subfigure}[b]{0.45\textwidth}
        \includegraphics[width=\textwidth]{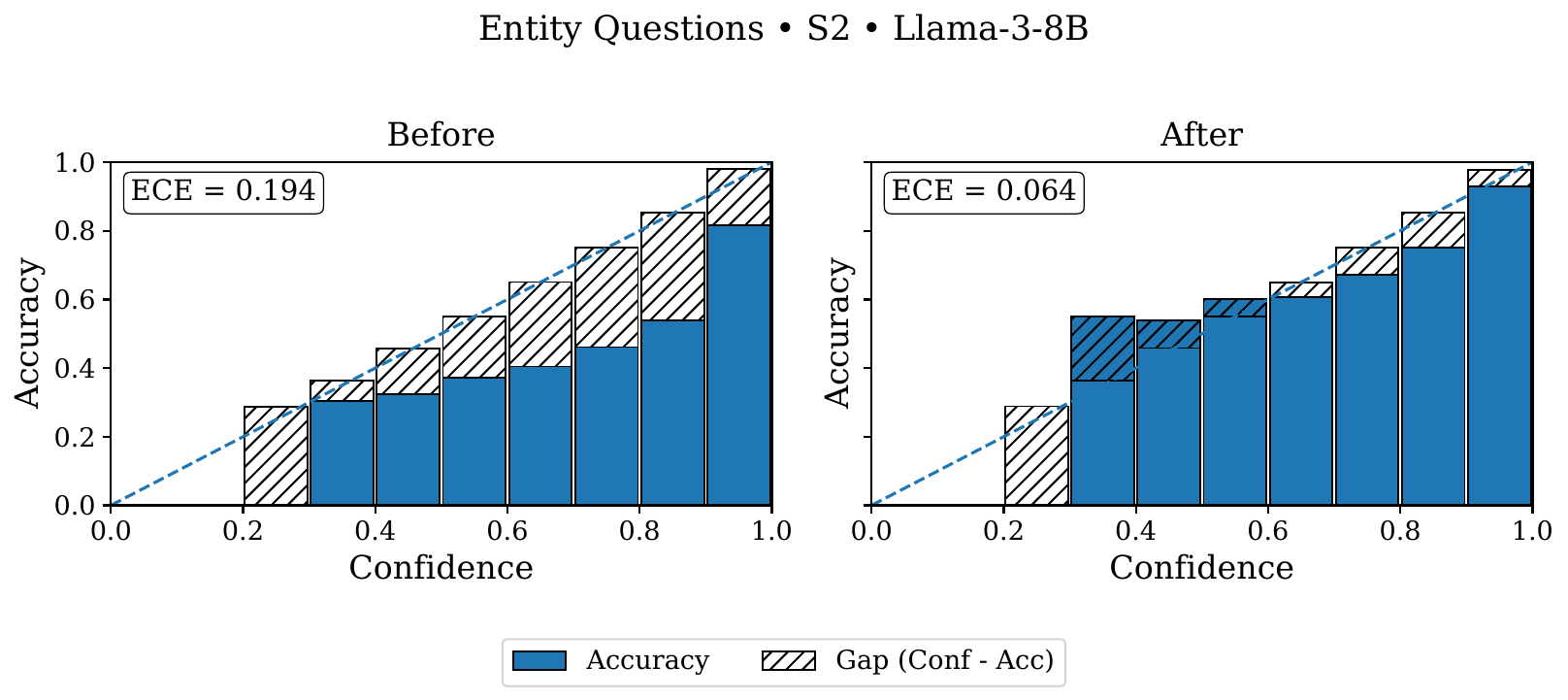}
        \caption{Llama-3-8B: ECE 0.194 → 0.064 (-67\%)}
        \label{fig:reliability_s2_llama}
    \end{subfigure}
    \hfill
    \begin{subfigure}[b]{0.45\textwidth}
        \includegraphics[width=\textwidth]{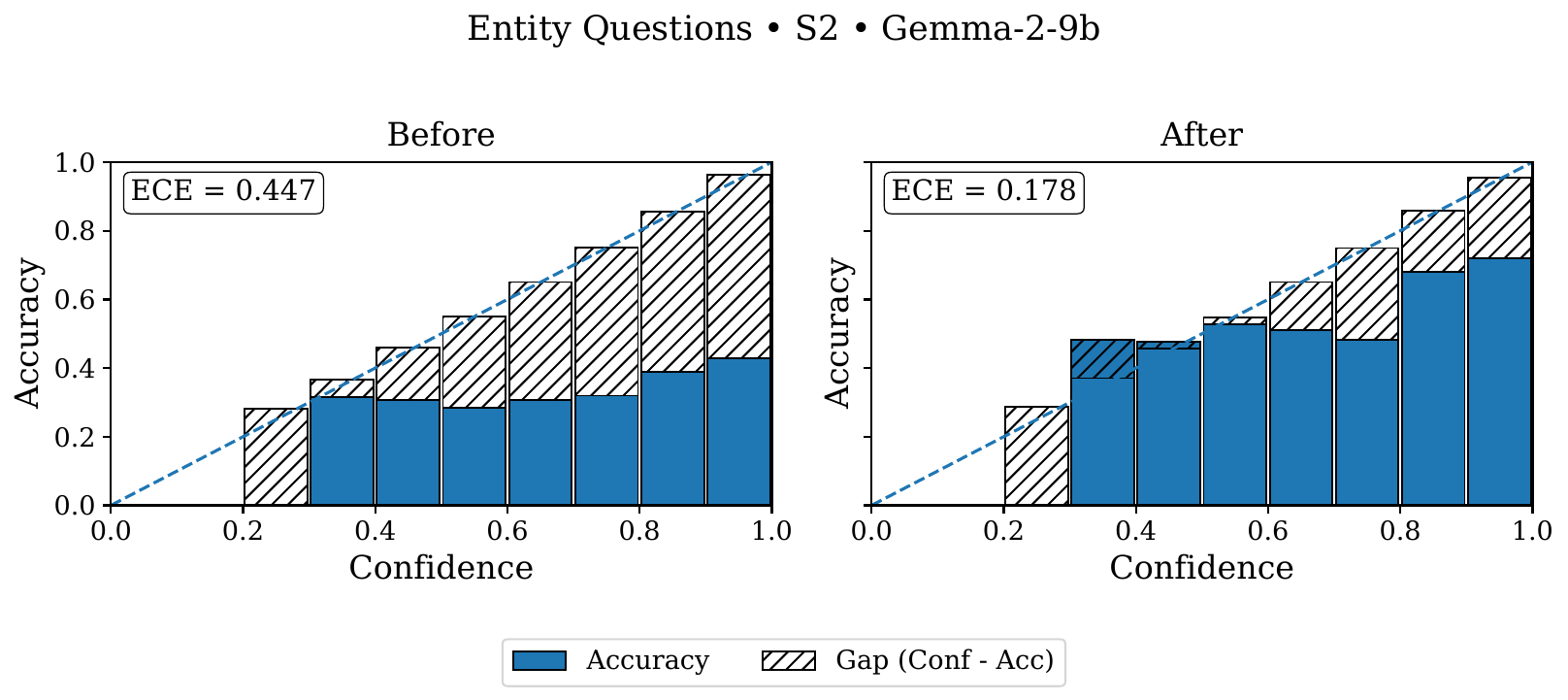}
        \caption{Gemma-2-9b: ECE 0.447 → 0.178 (-60\%)}
        \label{fig:reliability_s2_Gemma}
    \end{subfigure}
    \caption{\textbf{Calibration on S2 (Popular Trap).} Reliability diagrams before (left) and after (right) PopDebias on EntityQuestions. Hatched regions indicate the confidence-accuracy gap; PopDebias reduces overconfidence. See Appendix~\ref{app:calibration}.}
    \label{fig:reliability_s2}
\end{figure}

Beyond improving accuracy, PopDebias significantly enhances model calibration as measured by Expected Calibration Error (ECE)~\citep{guo2017calibration}. Figure~\ref{fig:reliability_s2} shows reliability diagrams for S2 (Popular Trap) on EntityQuestions, where before debiasing, both Llama-3-8B and Gemma-2-9b exhibit severe overconfidence with large gaps between confidence and accuracy. PopDebias dramatically reduces this miscalibration, achieving 67\% ECE reduction for Llama-3-8B (0.194 → 0.064) and 60\% for Gemma-2-9b (0.447 → 0.178). See Appendix~\ref{app:calibration} for other strategies.



\subsection{Prompting Strategy Investigation}
\label{sec:ablation_prompting}

We investigate whether standard prompting techniques can mitigate popularity bias. Figure~\ref{fig:prompting_comparison} compares Standard, Chain-of-thought (CoT), and Bias Warning prompts against PopDebias on S2 (Popular Trap) using MuSiQue. Prompting techniques provide minimal improvement: CoT and bias warnings achieve only 0--4 pp accuracy gains with correlation remaining at $\rho \approx -0.85$ to $-0.88$. In contrast, PopDebias delivers 21--34 pp accuracy improvements and reduces $\rho$ to $-0.65$ to $-0.75$, demonstrating that explicit popularity modeling is essential. Few-shot prompting (5/10 exemplars) similarly fails to resolve the bias, improving accuracy by only 1--2~pp while actually increasing HPSR (Table~\ref{tab:fewshot_ablation}, Appendix~\ref{app:fewshot}). See Appendix~\ref{app:prompting_full} for complete results.


\subsection{Debiasing Method Comparison}
\label{sec:ablation_debiasing}

We compare PopDebias against PrideDebias~\citep{zheng2023large}, a permutation-based debiasing method that randomizes answer option positions. Table~\ref{tab:debiasing_comparison} shows results on S2 (Popular Trap) with standard prompting across four models. PopDebias strongly outperforms PrideDebias on all models and metrics. PrideDebias achieves almost no improvement (0.2–2.0 pp accuracy gain, virtually no HPSR reduction), while PopDebias delivers 21.5–34.1 pp accuracy gains and reduces HPSR by 22–35 pp. 
This difference is expected because PrideDebias targets position bias (e.g., preference for option A), whereas PopDebias explicitly models entity popularity. 
Since the bias examined here arises from option content rather than position, a content-aware correction is better suited to this setting.
Full interaction analysis between prompting strategies and debiasing methods is in Appendix~\ref{app:prompting_full}.

\begin{table}[t]
\centering
\small

\footnotesize
\setlength{\tabcolsep}{3pt}
\caption{\textbf{Debiasing comparison on S2 (MuSiQue).} PopDebias vs.\ PrideDebias (position-based). We report Accuracy and HPSR as Baseline → After ($\Delta$).}
\label{tab:debiasing_comparison}
\scalebox{0.9}{
\begin{tabular}{l|cc|cc}
\toprule
& \multicolumn{2}{c|}{\textbf{+PopDebias}} & \multicolumn{2}{c}{\textbf{+PrideDebias}} \\
\cmidrule(lr){2-3} \cmidrule(lr){4-5}
\textbf{Model} & \textbf{Acc} & \textbf{HPSR} & \textbf{Acc} & \textbf{HPSR} \\
\midrule
Llama-3-8B & 45.5\,\textcolor{teal}{(+21.5)} & 51.7\,\textcolor{teal}{(-22.3)} & 45.5\,\textcolor{gray}{(-0.2)} & 51.7\,\textcolor{gray}{(+0.3)} \\
Qwen2.5-7B & 42.0\,\textcolor{teal}{(+34.1)} & 55.0\,\textcolor{teal}{(-35.1)} & 42.0\,\textcolor{gray}{(-1.5)} & 55.0\,\textcolor{gray}{(+1.1)} \\
Gemma-2-9B & 33.6\,\textcolor{teal}{(+20.7)} & 62.6\,\textcolor{teal}{(-21.2)} & 33.6\,\textcolor{gray}{(+0.9)} & 62.6\,\textcolor{gray}{(-1.1)} \\
Phi-4 & 52.6\,\textcolor{teal}{(+21.4)} & 44.6\,\textcolor{teal}{(-21.4)} & 52.6\,\textcolor{gray}{(+0.3)} & 44.6\,\textcolor{gray}{(-0.9)} \\
\bottomrule
\end{tabular}}
\end{table}

\section{Conclusion}
\label{sec:conclusion}
LLMs systematically favor popular but incorrect MCQ options---a vulnerability we term popularity bias. \textbf{PopMCQ} (six strategies, 22 models, four datasets) shows $\rho = -0.89$ under S2 with 66\% popular-but-wrong selections; S5 flips $\rho$ positive, isolating popularity as the driver. 99.8\% of S2 errors on known-answer items pick a more-popular distractor---a decision-time effect associated with confidence miscalibration and entity familiarity. \textbf{PopDebias}, a lightweight inference-time correction, lifts all 22 models (up to +54.1~pp accuracy, $-67\%$ ECE), outperforming permutation-based methods.

\section*{Acknowledgments}
The authors would like to acknowledge the financial support provided by the Austrian Research Agency (FFG) for the project “AI Enabled Sustainability Jurisdiction Demonstrator” (project No. 915229). The computational results presented in this work have been achieved using the MUSICA cluster, part of the Austrian Scientific Computing (ASC) infrastructure.

\section*{Limitations}
\label{sec:limitations}

Our study has several limitations. First, our investigation
focuses on the MCQ evaluation format; while MCQs are used in
major benchmarks (MMLU, ARC, AGIEval, C-Eval) and high-stakes
applications, extending the analysis to open-ended generation
and free-form QA is an important direction for future work. Second, our experiments focus on open-source models where we can access output probabilities; closed-source APIs that only return text outputs would require adaptation. Third, PopDebias requires a small calibration split (10\%) to fit the bias estimator, which may not always be available in zero-resource settings. Fourth, we evaluate on English datasets only; popularity bias patterns may differ across languages and cultures. Fifth, our popularity proxy relies on Wikipedia page views, which may not fully capture all familiarity signals that LLMs encode internally. Entities that are prominent in non-English text, domain-specific corpora, or synthetic training data may have low Wikipedia page views but high LLM familiarity. Validating alternative popularity proxies---such as training-corpus term frequency, embedding-space centrality, or model-internal confidence priors---is an important direction for future work. Finally, while PopDebias improves accuracy substantially, it does not fully eliminate the bias (correlation remains negative under adversarial conditions), suggesting room for future work on stronger debiasing methods. Future work may extend these ideas to multilingual settings, closed-source models, and combination with training-time debiasing approaches (Appendix~\ref{app:training_debiasing}).

\bibliography{custom}

\appendix
\appendixpage            
\addappheadtotoc         

This appendix provides supplementary material to support the main findings of this work. It is organized as follows:

\begin{itemize}
    \item Appendix~\ref{app:related_work} provides extended related work on frequency-dependent LLM behavior, MCQ biases, and calibration.
    \item Appendix~\ref{app:dataset_construction} describes dataset construction, including source datasets and filtering, candidate generation prompts, evaluation prompts, dataset statistics, and model families.
    \item Appendix~\ref{app:strategy_details} details strategy design (full descriptions, S2 vs.\ S3 distinction, and why accuracy can increase under S2).
    \item Appendix~\ref{app:wiki_scoring} covers Wikipedia popularity scoring (methodology, popularity vs.\ plausibility orthogonality, proxy validation, and correct-answer distributions).
    \item Appendix~\ref{app:metrics} defines all evaluation metrics.
    \item Appendix~\ref{app:algorithm} provides the PopDebias algorithm (pseudocode and a step-by-step numerical example).
    \item Appendix~\ref{app:robustness} presents robustness and additional analyses (scoring robustness, few-shot prompting, prompting$\times$debiasing interaction, scaling analysis, family-level bias, and relation to training-time debiasing).
    \item Appendix~\ref{app:open_recall} reports the open-recall diagnostic and cross-strategy ``known-under-control'' analysis, separating popularity bias from lack of knowledge.
    \item Appendix~\ref{app:confidence_gap} quantifies the confidence--accuracy gap by popularity bucket.
    \item Appendix~\ref{app:aggregate_debias} provides aggregate PopDebias results across all six strategies.
    \item Appendix~\ref{app:full_results} contains per-dataset full results for EntityQuestions, MuSiQue, Natural Questions, and WebQuestions.
    \item Appendix~\ref{app:calibration} presents calibration analysis with reliability diagrams for all strategies.
    \item Appendix~\ref{sec:full_results_for_Result} reports full debiasing results for all 22 models across all strategies and datasets.
\end{itemize}

\paragraph{Key Tables}
\begin{itemize}
    \item Table~\ref{tab:dataset_stats}: Dataset statistics (correct-answer popularity distributions).
    \item Table~\ref{tab:wiki_coverage}: Wikipedia mapping coverage for correct answers and distractors.
    \item Table~\ref{tab:proxy_validation}: Proxy validation correlating Wikipedia page views with model-internal log-probabilities.
    \item Table~\ref{tab:open_recall_full}: Extended open-recall diagnostic on MuSiQue (S2), with S2~$\rho$ and HPSR on the known-answer subset.
    \item Table~\ref{tab:cross_strategy_flips}: Cross-strategy flip analysis showing fraction of S2 errors where models answer correctly under S5.
    \item Table~\ref{tab:confidence_gap}: Confidence--accuracy gap by popularity bucket (MuSiQue, all 22 models).
    \item Table~\ref{tab:aggregate_debias}: Aggregate PopDebias results across all six strategies, averaged over 22 models and 4 datasets.
    \item Table~\ref{tab:fewshot_ablation}: Few-shot prompting ablation on MuSiQue (S2).
    \item Table~\ref{tab:scoring_health}: Probability distribution health statistics across models.
    \item Table~\ref{tab:scaling_summary}: Scaling analysis summary for the Qwen2.5 family.
\end{itemize}

\paragraph{Key Figures}
\begin{itemize}
    \item Figure~\ref{fig:pop_plaus_corr}: Popularity vs.\ plausibility correlation across datasets.
    \item Figure~\ref{fig:answer-popularity-dist-appendix}: Correct-answer popularity distributions.
    \item Figure~\ref{fig:scaling_analysis}: Scaling analysis for the Qwen2.5 family (0.5B--32B).
    \item Figure~\ref{fig:s2_family_bias}: Family-level popularity bias comparison under S2.
    \item Figures~\ref{fig:reliability_s1}--\ref{fig:reliability_s6}: Reliability diagrams for all strategies (S1--S6) on EntityQuestions.
\end{itemize}



\section{Extended Related Work}
\label{app:related_work}

A growing body of work links LLM performance to the head–tail frequency of knowledge.
\citet{kandpal2023large} show that models learn head (popular) facts much more readily
than tail facts and that scale disproportionately benefits head knowledge.
\citet{razeghi2022impactpretrainingtermfrequencies} further demonstrate that
pre-training duplication frequency strongly predicts downstream success, reinforcing the
view that distributional popularity shapes what models recall. Our study asks a
complementary question in MCQs: when popularity conflicts with truth at \emph{inference
time}, do models over-weight popular options?

Formatting choices in MCQs can systematically sway model predictions.
\citet{pezeshkpour2023largelanguagemodelssensitivity,zheng2023large} document robust
\emph{position bias} in multiple-choice settings, where merely reordering options shifts
LLM choices, suggesting that non-semantic cues can drive errors. We add a distinct,
orthogonal factor—\emph{option popularity}—and show that it reliably induces accuracy
drops and strong negative correctness–popularity associations under adversarial
strategies. Work on LLM calibration and familiarity effects observes that models can be
confidently wrong, especially on tail facts; popularity often correlates with both
accuracy and confidence~\citep{kandpal2023large,razeghi2022impactpretrainingtermfrequencies}.
Our results connect this to MCQ behavior: popularity acts like a label-free prior that
inflates confidence on well-known entities, motivating a prior-division correction.


\section{Dataset Construction}
\label{app:dataset_construction}

\subsection{Source Datasets \& Filtering}
\label{app:filtering}

\paragraph{Filtering.}
Not all source QA items can be converted into all popularity-controlled MCQ variants.
We retain only questions for which (i) the gold answer is present, (ii) at least four
candidate options are available, and (iii) the required constraints for core strategies
are satisfiable; in particular, S2 requires at least three candidates with
$\mathrm{pop}>\mathrm{pop}(a^\star)$ and S6 requires at least three candidates to
instantiate the ``none-of-the-above'' setting.  Items failing these checks are
discarded.\footnote{This selection does not use model outputs, question difficulty, or
accuracy; it depends only on candidate availability and popularity-constraint
satisfiability.}  This filtering is \emph{mechanical} rather than performance-based,
and is necessary to ensure that popularity is manipulated as an independent variable
under fixed question/answer semantics.

\paragraph{Handling infeasible strategies.}
We require S1, S2, and S6 to be constructible for every question; items failing these
checks are removed.  For S3–S5, if the popularity constraints cannot be met (e.g., not
enough higher- or lower-popularity candidates), we fall back to S1 for that strategy,
keeping our core ``popularity trap'' (S2) and ``none-of-the-above'' (S6) conditions
intact while preserving coverage.

\subsection{Candidate Generation Prompt}
\label{app:candidate_prompt}

We use the following prompt with GPT-4o to generate 20 candidate answers per question.
Generation parameters: temperature 0.7, max tokens 2\,000, 20 questions processed
concurrently per batch, up to 5 retries on failure.

\begin{figure*}[h]
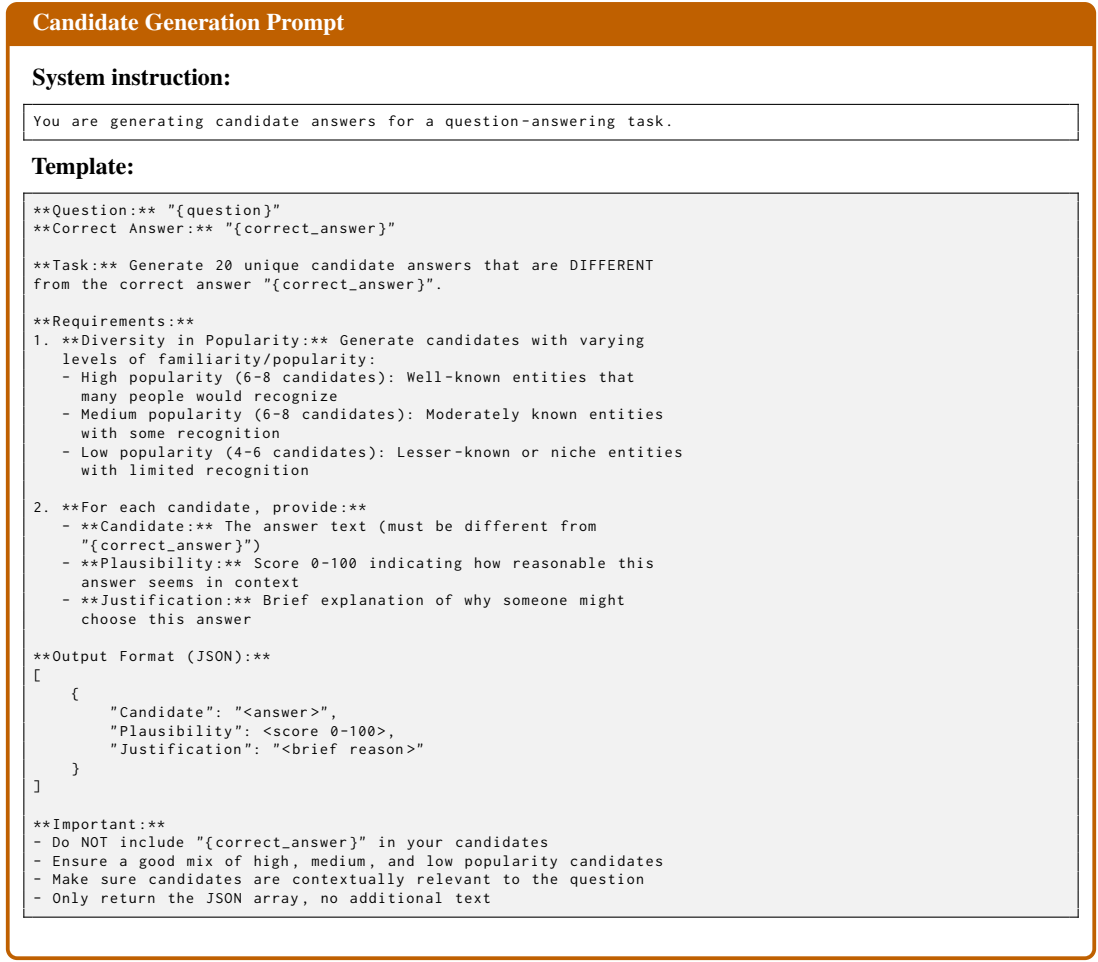

\centering
\begin{subfigure}[t]{1\textwidth}
\footnotesize
\centering

\begin{tcolorbox}[colback=blue!0!white, colframe=orange!75!black,
                 title=Candidate Generation Prompt, fonttitle=\bfseries,
                 width=0.9\textwidth, left=2mm, right=2mm, top=2mm, bottom=2mm]
\textbf{System instruction:}
\begin{lstlisting}[basicstyle=\ttfamily\tiny, frame=single, backgroundcolor=\color{gray!0}]
You are generating candidate answers for a question-answering task.
\end{lstlisting}

\textbf{Template:}
\begin{lstlisting}[basicstyle=\ttfamily\tiny, frame=single, backgroundcolor=\color{gray!10}]
**Question:** "{question}"
**Correct Answer:** "{correct_answer}"

**Task:** Generate 20 unique candidate answers that are DIFFERENT 
from the correct answer "{correct_answer}". 

**Requirements:**
1. **Diversity in Popularity:** Generate candidates with varying 
   levels of familiarity/popularity:
   - High popularity (6-8 candidates): Well-known entities that 
     many people would recognize
   - Medium popularity (6-8 candidates): Moderately known entities 
     with some recognition
   - Low popularity (4-6 candidates): Lesser-known or niche entities 
     with limited recognition

2. **For each candidate, provide:**
   - **Candidate:** The answer text (must be different from 
     "{correct_answer}")
   - **Plausibility:** Score 0-100 indicating how reasonable this 
     answer seems in context
   - **Justification:** Brief explanation of why someone might 
     choose this answer

**Output Format (JSON):**
[
    {
        "Candidate": "<answer>",
        "Plausibility": <score 0-100>,
        "Justification": "<brief reason>"
    }
]

**Important:** 
- Do NOT include "{correct_answer}" in your candidates
- Ensure a good mix of high, medium, and low popularity candidates
- Make sure candidates are contextually relevant to the question
- Only return the JSON array, no additional text
\end{lstlisting}
\end{tcolorbox}
\end{subfigure}
\caption{The prompt used for generating candidate answers, with placeholders for the question and ground-truth answer.}
\label{fig:rankllm-train-example}
\end{figure*}

\subsection{Evaluation Prompts}
\label{app:eval_prompts}

\begin{figure*}[h]
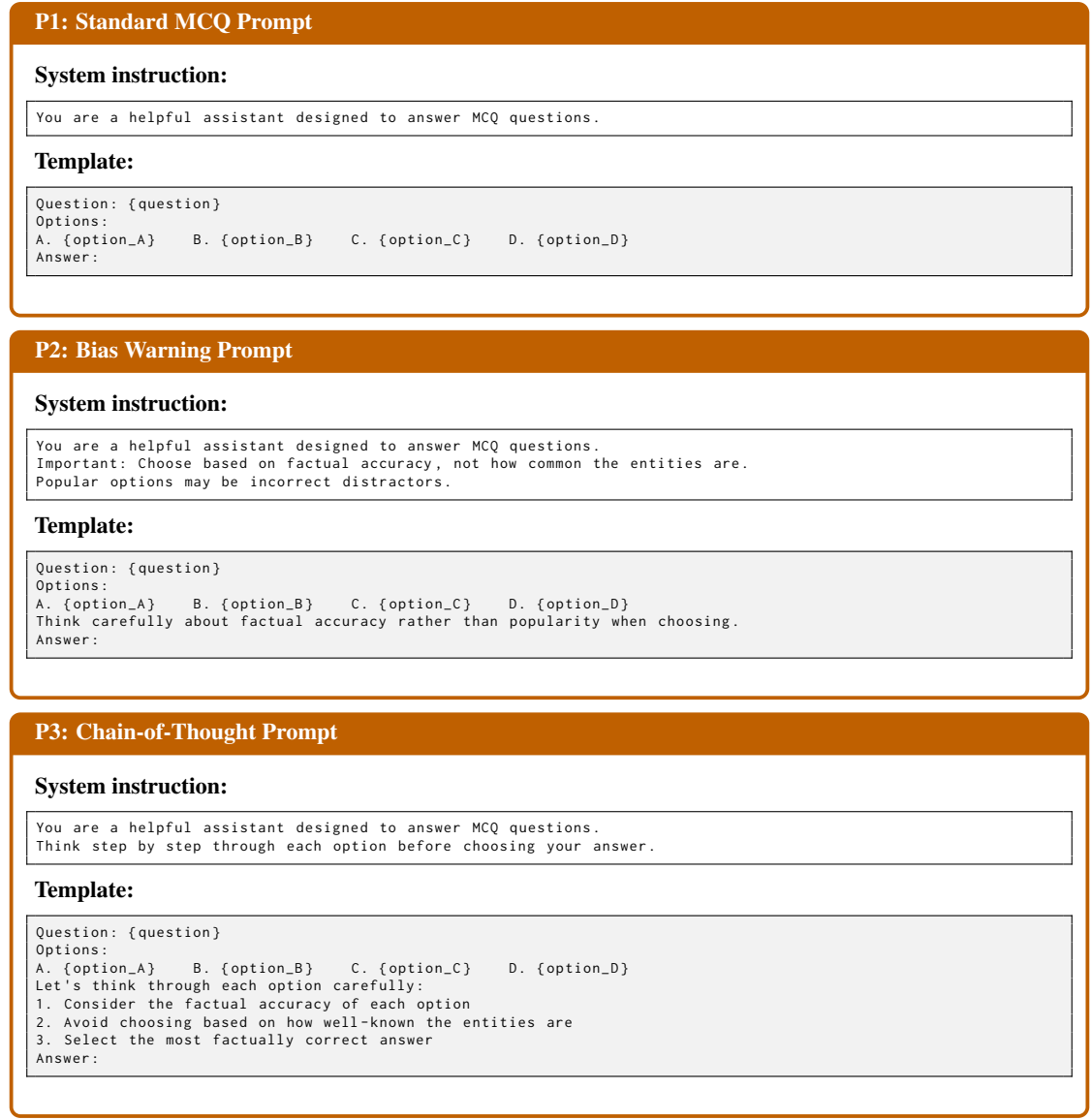

\centering
\footnotesize

\begin{tcolorbox}[colback=blue!0!white, colframe=orange!75!black,
                 title=P1: Standard MCQ Prompt, fonttitle=\bfseries,
                 width=0.92\textwidth, left=2mm, right=2mm, top=2mm, bottom=2mm]
\textbf{System instruction:}
\begin{lstlisting}[basicstyle=\ttfamily\tiny, frame=single, backgroundcolor=\color{gray!0}]
You are a helpful assistant designed to answer MCQ questions.
\end{lstlisting}
\textbf{Template:}
\begin{lstlisting}[basicstyle=\ttfamily\tiny, frame=single, backgroundcolor=\color{gray!10}]
Question: {question}
Options:
A. {option_A}    B. {option_B}    C. {option_C}    D. {option_D}
Answer:
\end{lstlisting}
\end{tcolorbox}

\begin{tcolorbox}[colback=blue!0!white, colframe=orange!75!black,
                 title=P2: Bias Warning Prompt, fonttitle=\bfseries,
                 width=0.92\textwidth, left=2mm, right=2mm, top=2mm, bottom=2mm]
\textbf{System instruction:}
\begin{lstlisting}[basicstyle=\ttfamily\tiny, frame=single, backgroundcolor=\color{gray!0}]
You are a helpful assistant designed to answer MCQ questions.
Important: Choose based on factual accuracy, not how common the entities are.
Popular options may be incorrect distractors.
\end{lstlisting}
\textbf{Template:}
\begin{lstlisting}[basicstyle=\ttfamily\tiny, frame=single, backgroundcolor=\color{gray!10}]
Question: {question}
Options:
A. {option_A}    B. {option_B}    C. {option_C}    D. {option_D}
Think carefully about factual accuracy rather than popularity when choosing.
Answer:
\end{lstlisting}
\end{tcolorbox}

\begin{tcolorbox}[colback=blue!0!white, colframe=orange!75!black,
                 title=P3: Chain-of-Thought Prompt, fonttitle=\bfseries,
                 width=0.92\textwidth, left=2mm, right=2mm, top=2mm, bottom=2mm]
\textbf{System instruction:}
\begin{lstlisting}[basicstyle=\ttfamily\tiny, frame=single, backgroundcolor=\color{gray!0}]
You are a helpful assistant designed to answer MCQ questions.
Think step by step through each option before choosing your answer.
\end{lstlisting}
\textbf{Template:}
\begin{lstlisting}[basicstyle=\ttfamily\tiny, frame=single, backgroundcolor=\color{gray!10}]
Question: {question}
Options:
A. {option_A}    B. {option_B}    C. {option_C}    D. {option_D}
Let's think through each option carefully:
1. Consider the factual accuracy of each option
2. Avoid choosing based on how well-known the entities are
3. Select the most factually correct answer
Answer:
\end{lstlisting}
\end{tcolorbox}

\caption{Prompts used for standard evaluation (P1), explicit bias warning (P2), and
chain-of-thought reasoning (P3).}
\label{fig:prompt-ablation-prompts}
\end{figure*}

Figure~\ref{fig:prompt-ablation-prompts} shows the three prompting conditions used in
our prompting-strategy comparison (\S\ref{sec:ablation_prompting}).

\subsection{Dataset Statistics}
\label{app:dataset_stats}

Table~\ref{tab:dataset_stats} reports the number of questions retained after filtering.

\begin{table}[h]
\centering
\small
\caption{PopMCQ dataset statistics after filtering.}
\label{tab:dataset_stats}
\begin{tabular}{lr}
\toprule
\textbf{Dataset} & \textbf{Questions} \\
\midrule
EntityQuestions & 9\,584 \\
Natural Questions & 1\,050 \\
MuSiQue & 874 \\
WebQuestions & 702 \\
\midrule
\textbf{Total} & \textbf{12\,210} \\
\bottomrule
\end{tabular}
\end{table}

\subsection{Model Families}
\label{app:models}

We evaluate 22 open-source LLMs spanning 0.5B–32B parameters across six families:

\begin{itemize}[leftmargin=*, nosep]
\item \textbf{Llama:} Llama-2-7B/13B, Llama-3-8B, Llama-3.1-8B, Llama-3.2-1B/3B
\item \textbf{Qwen2.5:} 0.5B, 1.5B, 3B, 7B, 32B
\item \textbf{Gemma:} Gemma-3-1B, Gemma-2-2B/9B
\item \textbf{Phi:} Phi-1.5, Phi-2, Phi-4
\item \textbf{DeepSeek:} DeepSeek-V2-Lite, DeepSeek-LLM-7B
\item \textbf{Others:} Falcon-7B, Mistral-7B-v0.3, Zephyr-7B
\end{itemize}


\section{Strategy Design Details}
\label{app:strategy_details}

\subsection{Full Strategy Descriptions \& Rationale}
\label{app:strategy_full}

Each base question yields six MCQ variants that differ \emph{only} in distractor
popularity.  All MCQs have four options (one correct, three distractors) with positions
randomized over \{A,B,C,D\}.  Let $\mathcal{C}(q)$ denote the candidate pool and
$\mathrm{pop}(a^\star)$ the correct answer's popularity.

\paragraph{S1 (Baseline Control).} Three distractors selected uniformly at random from
$\mathcal{C}(q)$ with no popularity constraints.  Serves as the reference condition.

\paragraph{S2 (Popular Trap).} Top-3 most popular candidates satisfying
$\mathrm{pop}>\mathrm{pop}(a^\star)$, maximising aggregate popularity pressure against
the correct option.

\paragraph{S3 (Popularity Gradient).} Three distractors with strictly decreasing
popularity (high $>$ medium $>$ low), all above $\mathrm{pop}(a^\star)$.  Tests whether
models exhibit a \emph{graded} preference along a popularity ranking.

\paragraph{S4 (Direct Popularity Contest).} One highest-$\mathrm{pop}$ candidate plus
two medium-$\mathrm{pop}$ distractors—a head-to-head stress test against the single
most famous alternative.

\paragraph{S5 (Reverse Popularity Control).} Three distractors with
$\mathrm{pop}<\mathrm{pop}(a^\star)$, so the correct answer is the \emph{most} popular
option. Popularity and truth align; serves as a positive control.

\paragraph{S6 (None of the Above).} The correct option is ``None of the above''
($\mathrm{pop}=0$); distractors are $1\times$ highest-$\mathrm{pop}$ and $2\times$
lowest-$\mathrm{pop}$ candidates.  Stress-tests abstention under popularity pressure
and measures whether a single ``celebrity distractor'' attracts selections even when
abstention is correct.  Because the gold label has $\mathrm{pop}=0$ by design, S6 is
interpreted through accuracy and \HPSR{} rather than Spearman $\rho$.

\subsection{Distinguishing S2 and S3}
\label{app:s2_s3}

While both S2 and S3 enforce $\mathrm{pop}(\text{distractor}) > \mathrm{pop}(a^\star)$,
they differ in design intent.  S2 selects the top-3 most popular candidates, maximising
aggregate popularity pressure.  S3 selects three distractors with strictly decreasing
popularity (high $>$ medium $>$ low, all above truth), creating an explicit gradient.
S2 tests whether placing truth in the ``popularity long tail'' under maximal pressure
induces errors; S3 tests whether models exhibit a \emph{graded} popularity preference
among distractors of differing popularity.  Empirically, S2 yields stronger negative
correlations ($\rho{=}{-}0.864$ vs.\ ${-}0.704$ for S3 on MuSiQue, averaged over 22
models), consistent with S2 creating maximal pressure while S3 distributes it along a
ranking.

\subsection{Why Accuracy Can Increase Under S2}
\label{app:acc_increase}

Table~\ref{tab:musique_key_results_summary} shows that accuracy sometimes improves from
S1 to S2 (e.g., Gemma-2-9B: 29.3\%$\to$33.6\%).  This is not contradictory: S2
changes the \emph{composition} of distractors, not only the popularity signal.  By
selecting the top-3 most popular candidates, S2 can produce globally famous but
\emph{question-agnostic} options that are easier to eliminate than the more
locally-plausible distractors that may appear under random sampling in S1.  Crucially,
our claim is not that S2 always lowers accuracy, but that when models err they do so in
a popularity-biased way.  This is captured by $\rho$ and \HPSR{} rather than accuracy
alone: under S2 the dataset-level mean correlation becomes $-0.864$ (vs.\ $-0.234$ in
S1) and \HPSR{} rises from 49.9\% to 66.0\%, confirming that errors are
popularity-biased.


\section{Wikipedia Popularity Scoring}
\label{app:wiki_scoring}

\subsection{Scoring Methodology}
\label{app:pop_method}

We quantify popularity by mapping each option to a Wikipedia page via title
normalization~\citep{mozafari2025hinteval}, retrieving monthly page-view counts
(Jan 2015–Dec 2024), and applying min–max normalization to $[0,1]$ using the HintEval
metric~\citep{mozafari2025hinteval}.  We \emph{relativize} within each MCQ:
$\mathrm{pop}_i \leftarrow p_i / \max_j p_j$, so the most popular option receives
$\mathrm{pop}=1.0$; options without Wikipedia pages receive $\mathrm{pop}=0$.

\paragraph{Why Wikipedia page views.}
We adopt Wikipedia page views following PopQA~\citep{mallen-etal-2023-trust}, which
established this as a standard entity-popularity proxy in NLP and demonstrated a clear
correlation with LLM recall accuracy.  Crucially, since we evaluate 22 models from
diverse families trained on different corpora, we require a signal that is \emph{external
and model-agnostic}.  Wikipedia page views satisfy this: they provide a consistent,
reproducible, and publicly accessible measure that does not depend on any specific
model's training data.

\paragraph{Coverage.}
As shown in Table~\ref{tab:wiki_coverage}, coverage is high: 98.4–99.5\% of correct
answers and 99.3–99.8\% of candidate distractors map successfully, confirming that
unmapped options (defaulting to $\mathrm{pop}=0$) are rare.

\begin{table}[h]
\centering
\small
\setlength{\tabcolsep}{5pt}
\renewcommand{\arraystretch}{1.10}
\caption{Wikipedia mapping coverage. Unmapped options ($<$2\%) default to
$\mathrm{pop}=0$ and are rare enough not to affect results materially.}
\label{tab:wiki_coverage}
\scalebox{0.90}{
\begin{tabular}{lcccc}
\toprule
& \multicolumn{2}{c}{\textbf{Correct Answers}} & \multicolumn{2}{c}{\textbf{Distractors}} \\
\cmidrule(lr){2-3}\cmidrule(lr){4-5}
\textbf{Dataset} & \textbf{Mapped} & \textbf{\%} & \textbf{Mapped} & \textbf{\%} \\
\midrule
EntityQuestions & 9{,}531/9{,}584 & 99.4 & 28{,}558/28{,}752 & 99.3 \\
NQ              & 1{,}033/1{,}050 & 98.4 & 3{,}130/3{,}150   & 99.4 \\
MuSiQue         & 870/874         & 99.5 & 2{,}616/2{,}622   & 99.8 \\
WebQuestions    & 695/702         & 99.0 & 2{,}100/2{,}106   & 99.7 \\
\midrule
\textbf{Overall} & \textbf{12{,}129/12{,}210} & \textbf{99.3} & \textbf{36{,}404/36{,}630} & \textbf{99.4} \\
\bottomrule
\end{tabular}}
\end{table}

\subsection{Popularity vs.\ Plausibility}
\label{app:pop_plaus}

\begin{figure}[h]
\centering
\includegraphics[width=\columnwidth]{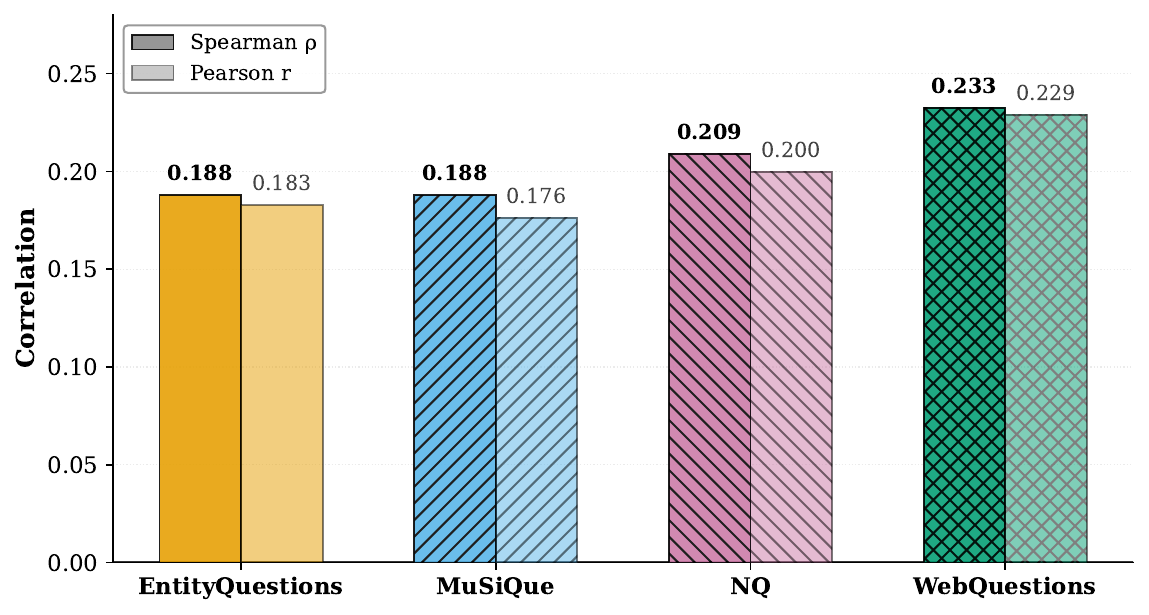}
\caption{\textbf{Wikipedia popularity vs.\ distractor plausibility.} Weak positive
correlations ($\rho<0.24$) confirm the two signals are largely orthogonal.}
\label{fig:pop_plaus_corr}
\end{figure}

A natural concern is whether popularity simply coincides with answer plausibility.  To
test this, we compute the Spearman and Pearson correlations between Wikipedia popularity
and GPT-4o-assigned plausibility scores across all incorrect options in our four
datasets.  As shown in Figure~\ref{fig:pop_plaus_corr}, both correlations are
consistently weak (Spearman $\rho \in [0.188, 0.233]$; Pearson $r \in [0.176, 0.229]$),
confirming that popularity and plausibility are largely orthogonal, and ruling out
plausibility as a confound for the bias we document.

\subsection{Proxy Validation}
\label{app:proxy}

To verify that Wikipedia page views capture the familiarity signals encoded during
pretraining, we compute the average token log-probability assigned by each LM to entity
surface forms (without any MCQ context):
\[
\mathrm{Fam}(e) = \frac{1}{L}\sum_{\ell=1}^{L}\log P_\theta(t_\ell\mid t_{<\ell}),
\]
and correlate with $\mathrm{Pop}(e)$ per model.  Table~\ref{tab:proxy_validation} shows
per-model Spearman correlations: mean $\rho=0.70$, range $[0.51, 0.85]$, consistently
positive across all 22 models from diverse training backgrounds.  Larger models and
those trained on broader web corpora show stronger correlations, consistent with the
expectation that richer pretraining better mirrors the public attention distribution
captured by Wikipedia.

\begin{table}[h]
\centering
\small
\setlength{\tabcolsep}{5pt}
\renewcommand{\arraystretch}{1.10}
\caption{Spearman $\rho$ between Wikipedia page-view popularity and model-assigned
entity log-probability.  Higher $\rho$ indicates stronger alignment between the
external proxy and the model's internal familiarity signal.}
\label{tab:proxy_validation}
\begin{tabular}{llcc}
\toprule
\textbf{Family} & \textbf{Model} & \textbf{Params} & \textbf{$\rho$} \\
\midrule
\multirow{6}{*}{Llama}
 & Llama-2-7B      & 7B   & 0.68 \\
 & Llama-2-13B     & 13B  & 0.74 \\
 & Llama-3-8B      & 8B   & 0.75 \\
 & Llama-3.1-8B    & 8B   & 0.76 \\
 & Llama-3.2-1B    & 1B   & 0.58 \\
 & Llama-3.2-3B    & 3B   & 0.65 \\
\midrule
\multirow{5}{*}{Qwen2.5}
 & Qwen2.5-0.5B    & 0.5B & 0.51 \\
 & Qwen2.5-1.5B    & 1.5B & 0.57 \\
 & Qwen2.5-3B      & 3B   & 0.63 \\
 & Qwen2.5-7B      & 7B   & 0.72 \\
 & Qwen2.5-32B     & 32B  & 0.85 \\
\midrule
\multirow{3}{*}{Gemma}
 & Gemma-3-1B      & 1B   & 0.55 \\
 & Gemma-2-2B      & 2B   & 0.62 \\
 & Gemma-2-9B      & 9B   & 0.76 \\
\midrule
\multirow{3}{*}{Phi}
 & Phi-1.5         & 1.3B & 0.52 \\
 & Phi-2           & 2.7B & 0.58 \\
 & Phi-4           & 14B  & 0.71 \\
\midrule
\multirow{2}{*}{DeepSeek}
 & DeepSeek-LLM-7B & 7B   & 0.70 \\
 & DeepSeek-V2-Lite& 16B  & 0.78 \\
\midrule
 & Falcon-7B       & 7B   & 0.67 \\
 & Mistral-7B-v0.3 & 7B   & 0.73 \\
 & Zephyr-7B       & 7B   & 0.72 \\
\midrule
\multicolumn{3}{l}{\textbf{Mean}} & \textbf{0.70} \\
\multicolumn{3}{l}{\textbf{Range}} & \textbf{[0.51, 0.85]} \\
\bottomrule
\end{tabular}
\end{table}

\subsection{Correct-Answer Popularity Distribution}
\label{app:pop_dist}

\begin{figure*}[h]
\centering
\includegraphics[width=\linewidth]{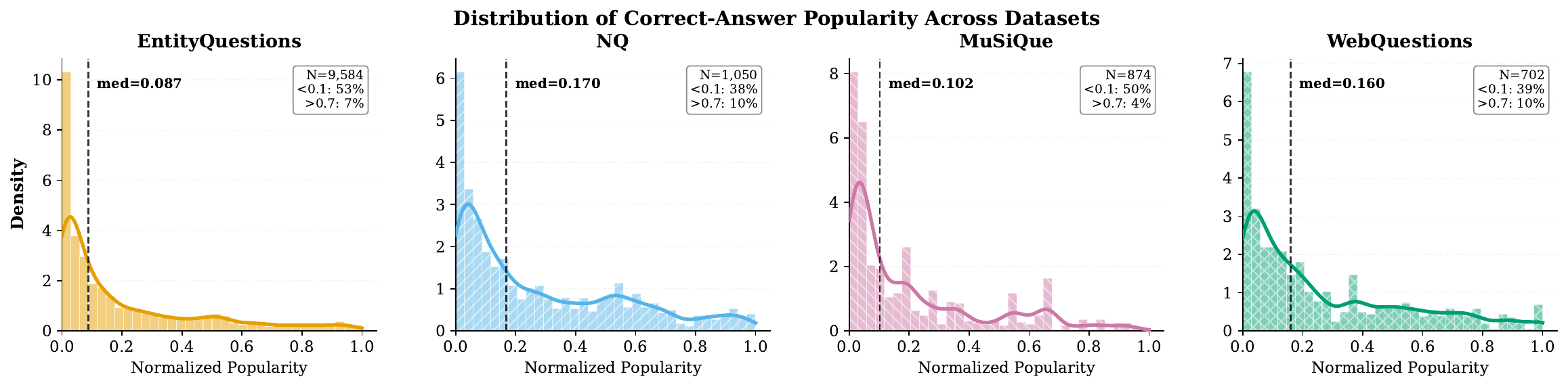}
\caption{Distribution of correct-option popularity across datasets. Each panel shows the
histogram and KDE for one dataset; the dashed line marks the median. EntityQuestions and
MuSiQue are strongly long-tailed, while NQ and WebQuestions show broader spread.}
\label{fig:answer-popularity-dist-appendix}
\end{figure*}

Figure~\ref{fig:answer-popularity-dist-appendix} shows the distribution of the correct
answer's normalized popularity across datasets.  The correct option frequently falls in
the long tail—often \emph{less} popular than the distractors.  EntityQuestions and
MuSiQue are strongly long-tailed (median $<0.11$; $>49\%$ below 0.1), while NQ and
WebQuestions have broader spread.  These natural popularity imbalances motivate our
controlled strategies: in standard evaluation, popularity effects are confounded with
question difficulty and answer plausibility.

\section{Evaluation Metrics}
\label{app:metrics}

\paragraph{Correlation ($\rho$).}
Spearman correlation between correctness and selected-option popularity:
\[
\rho = \mathrm{Spearman}\!\big(\mathbf{1}\{\hat{o}=o^\star\},\;\mathrm{Pop}(\hat{o})\big).
\]
Negative $\rho$ means selecting popular options is associated with being wrong; positive
$\rho$ means popularity aligns with correctness.  \emph{Note:} In S6 the correct answer
has $\mathrm{Pop}=0$ by design, so we interpret S6 through accuracy and \HPSR{}.

\paragraph{Popularity Gap (\PopGap{}).}
$\textsc{PopGap} = \mathbb{E}[\mathrm{Pop}(\hat{o}) - \mathrm{Pop}(o^\star)]$. 
Positive values indicate the model systematically selects options more popular than the
truth.

\paragraph{High-Popularity Selection Rate (\HPSR{}).}
Within each item, mark a selection as high-pop if its popularity rank is in the top
half:
$\mathrm{HPSR} = \mathbb{E}[\mathbf{1}\{\mathrm{rank}_{\mathrm{pop}}(\hat{o})\le\lfloor K/2\rfloor\}]$.
Higher \HPSR{} indicates a stronger pull toward common entities; 50\% is chance.

\paragraph{Accuracy.} $\mathrm{Acc}=\mathbb{E}[\mathbf{1}\{\hat{o}=o^\star\}]$.

\paragraph{Confidence.}
Average probability assigned to the chosen option: $\mathrm{Conf}=\mathbb{E}[\max_i p_i]$.

\paragraph{Alignment.}
Following~\citet{ni2025knowledge}:
$\mathrm{Align} = \mathbb{E}[1 - |\mathbf{1}\{\hat{o}=o^\star\}-\max_i p_i|]$. A
well-calibrated model is confident when correct and uncertain when wrong.

\paragraph{Overconfidence by Popularity Bucket.}
We split predictions by the popularity of the selected answer (Low: $\leq0.3$, Medium:
$(0.3,0.7]$, High: $>0.7$) and report the confidence–accuracy gap in each bucket:
$\mathrm{OverConf}(b)=\mathbb{E}[\max_i p_i - \mathbf{1}\{\hat{o}=o^\star\}\mid b]$.


\section{PopDebias Algorithm}
\label{app:algorithm}

\subsection{Pseudocode}
\label{app:algo_pseudo}

Algorithm~\ref{alg:app} gives the complete pseudocode.  The algorithm is $O(K)$ per
item (where $K{=}4$ options) and requires a single forward pass—unlike permutation-based
methods~\citep{zheng2023large} which need $K!$ passes.  After a one-time calibration
(default 10\% of data; dozens–hundreds of samples suffice), all inference is label-free.

\begin{algorithm}[ht]
\caption{\textsc{PopDebias}: single-pass implementation}
\label{alg:app}
\begin{adjustbox}{max width=\columnwidth}
\begin{minipage}{\linewidth}
\begin{algorithmic}[1]
\Require base model $\mathcal{M}$, items $\mathcal{D}$,
         optional regressor $f_\theta$,
         hyperparams $\beta{=}1$, $\bar{b}{>}0$, $\varepsilon{=}10^{-12}$,
         cap $b_{\max}$
\Ensure debiased predictions $\mathcal{Y}$

\State \textbf{One-time calibration:} fit $f_\theta$ and $\bar{b}$ on a small split ($\beta$ fixed).

\For{item $(q,x)\in\mathcal{D}$}
  \State $\mathbf{p}\leftarrow\mathcal{M}(q,x)$;\quad
         $c\leftarrow\max_i p_i$;\quad
         $\hat{\imath}\leftarrow\arg\max_i p_i$
  \State \textbf{Bias strength:}
         \textbf{if} $f_\theta$ \textbf{then} $b\leftarrow\mathrm{clip}(f_\theta(\mathbf{f}),0,b_{\max})$
         \textbf{else} use heuristic (Eq.~4)
  \State $\psi\leftarrow 1 - H(r)/\log K$ \quad (popularity pressure)
  \State $\gamma\leftarrow\mathrm{clip}(1-\beta\,b\,c\,\psi,\;0,\;1)$
  \State $w_i\leftarrow\mathrm{clip}\!\big(1-b\,\mathrm{Pop}(o_i)/(\max_j\mathrm{Pop}(o_j)+\varepsilon),\;0,\;1\big)$
  \State $q_i\leftarrow p_i w_i/\sum_j p_j w_j$;\quad
         $\widetilde{p}_i\leftarrow\gamma\,q_i+\tfrac{1-\gamma}{K}$
  \State Append $\arg\max_i\widetilde{p}_i$ to $\mathcal{Y}$
\EndFor
\State \Return $\mathcal{Y}$
\end{algorithmic}
\end{minipage}
\end{adjustbox}
\end{algorithm}

\paragraph{Key design principles.}
(1)~\textbf{Label-free at inference:} after calibration, no ground-truth labels are
needed.  (2)~\textbf{Adaptive correction:} bias-correction strength adapts per item via
$b$, $c$, and $\psi$.  (3)~\textbf{Preserves helpful signal:} when popularity aligns
with truth (S5), $b$ is small, $\gamma\approx1$, and $\widetilde{\mathbf{p}}\approx\mathbf{p}$.
(4)~\textbf{Reduces overconfidence:} the tempering term addresses confidence
miscalibration identified in \S\ref{subsec:causes}.

\subsection{Step-by-Step Numerical Example}
\label{app:worked-example}

Consider a 4-option item (correct answer is \textbf{C}):

\begin{center}
\small
\begin{tabular}{lcccc}
\toprule
Option & A & B & \textbf{C (truth)} & D \\
\midrule
$\mathrm{Pop}(o_i)$ & 0.92 & 0.60 & \textbf{0.35} & 0.40 \\
Base prob.\ $p_i$   & 0.40 & 0.25 & \textbf{0.20} & 0.15 \\
\bottomrule
\end{tabular}
\end{center}

The base model (argmax) wrongly selects \textbf{A}.  We apply \textsc{PopDebias}:

\begin{enumerate}[leftmargin=*,itemsep=3pt]
\item \textbf{Features:} $\bar{\mathrm{Pop}}=0.5675$, $\mathrm{Var}(\mathrm{Pop})=0.0502$,
      $c=0.40$, $\mathrm{Pop}(o_A)=0.92$.
\item \textbf{Bias strength} (heuristic with $\bar{b}=0.1$):
      mean of $\{1.621,\,0.40,\,19.94\}=7.32$;\; $b=0.1\times7.32=0.732$.
\item \textbf{Prior weights:}
      $w_A=0.268$, $w_B=0.523$, $w_C=0.722$, $w_D=0.682$.
\item \textbf{Tempering coefficient} ($\beta=1$, $\psi=1$):
      $\gamma=1-0.732\times0.40\times1=0.707$.
\item \textbf{Reweighting} ($p_i w_i$, then renormalize by $Z=0.4844$):
\begin{align*}
A &: 0.1072,\quad B: 0.1307,\\
C &: \mathbf{0.1443},\quad D: 0.1023,\\
\mathbf{q} &= (0.221,\;0.270,\;\mathbf{0.298},\;0.211).
\end{align*}
\item \textbf{Tempering:} $\widetilde{p}_i=\gamma q_i+\tfrac{1-\gamma}{4}$, giving
      $\widetilde{\mathbf{p}}=(0.230,\;0.264,\;\mathbf{0.284},\;0.222)$.
\item \textbf{Result:} argmax flips from A to \textbf{C} (correct).
      \PopGap{} drops from $+0.57$ to $0.00$; confidence falls from 0.40 to 0.284.
\end{enumerate}


\section{Robustness \& Additional Analyses}
\label{app:robustness}

\subsection{Scoring Method Robustness}
\label{app:scoring_robustness}

Our primary evaluation extracts next-token logits at the ``Answer:'' position and takes
the softmax over A/B/C/D token IDs—the standard protocol used in MMLU, ARC, and
PrideDebias.  We verify robustness through three analyses.

\paragraph{Within-experiment controls.}
All six strategies use identical prompt formatting, scoring code, and models—only option
\emph{content} changes.  The dramatic cross-strategy difference ($\rho=-0.89$ under S2
vs.\ $+0.19$ under S5 for the same model) cannot be attributed to measurement
artifacts.

\paragraph{Probability distribution health.}
Table~\ref{tab:scoring_health} reports entropy and margin statistics across four models.
Mean entropy is well above zero ($H>1.2$ bits out of a maximum of 2.0), and the
fraction of near-degenerate predictions ($\max p_i>0.95$) is low ($<15\%$).  S2 shows
slightly lower entropy than S1, consistent with models being more confidently drawn
toward popular distractors—this is the miscalibration effect, not a scoring artifact.

\begin{table}[h]
\centering
\small
\setlength{\tabcolsep}{4pt}
\renewcommand{\arraystretch}{1.10}
\caption{\textbf{Probability distribution health.} Entropy $H$, margin (top-1 minus
top-2 probability), and fraction of near-degenerate predictions. Distributions are
well-spread.}
\label{tab:scoring_health}
\begin{tabular}{ll|ccc}
\toprule
\textbf{Model} & \textbf{Strat.} & $H$ \textbf{(bits)} & \textbf{Margin} & \textbf{\%}$p{>}0.95$ \\
\midrule
\multirow{2}{*}{Llama-3-8B}      & S1 & 1.46 & 0.24 &  7.8\% \\
                                  & S2 & 1.31 & 0.31 & 12.4\% \\
\multirow{2}{*}{Qwen2.5-7B}      & S1 & 1.52 & 0.21 &  6.2\% \\
                                  & S2 & 1.36 & 0.28 & 10.5\% \\
\multirow{2}{*}{Gemma-2-9B}      & S1 & 1.58 & 0.19 &  4.9\% \\
                                  & S2 & 1.42 & 0.24 &  8.7\% \\
\multirow{2}{*}{DeepSeek-V2-Lite}& S1 & 1.41 & 0.26 &  9.3\% \\
                                  & S2 & 1.24 & 0.33 & 14.1\% \\
\bottomrule
\end{tabular}
\end{table}

\paragraph{Full-sequence log-likelihood scoring.}
Table~\ref{tab:fullseq_scoring} compares single-token and full-sequence scoring.  The
popularity bias pattern (strongly negative $\rho$ under S2, mild under S1) is
\textbf{consistent across both methods}.  Absolute accuracy differs by 2–3 pp, but the
S1$\to$S2 shift in $\rho$ is virtually identical (e.g., Llama-3-8B:
$-0.21\to-0.87$ single-token vs.\ $-0.23\to-0.85$ full-sequence), confirming that
our findings reflect genuine model behavior.

\begin{table}[h]
\centering
\small
\setlength{\tabcolsep}{4pt}
\renewcommand{\arraystretch}{1.10}
\caption{\textbf{Single-token vs.\ full-sequence scoring on MuSiQue.}  The popularity
bias pattern is preserved across both methods.}
\label{tab:fullseq_scoring}
\begin{tabular}{ll|cc|cc}
\toprule
 & & \multicolumn{2}{c|}{\textbf{Single-Token}} & \multicolumn{2}{c}{\textbf{Full-Sequence}} \\
\cmidrule(lr){3-4}\cmidrule(lr){5-6}
\textbf{Model} & \textbf{Strat.} & \textbf{Acc} & \textbf{$\rho$} & \textbf{Acc} & \textbf{$\rho$} \\
\midrule
\multirow{2}{*}{Llama-3-8B}       & S1 & 44.9 & -0.21 & 42.3 & -0.23 \\
                                   & S2 & 45.5 & -0.87 & 43.8 & -0.85 \\
\multirow{2}{*}{Qwen2.5-7B}       & S1 & 41.0 & -0.23 & 38.6 & -0.25 \\
                                   & S2 & 42.0 & -0.87 & 39.4 & -0.84 \\
\multirow{2}{*}{Gemma-2-9B}       & S1 & 29.3 & -0.20 & 27.8 & -0.22 \\
                                   & S2 & 33.6 & -0.86 & 31.5 & -0.84 \\
\multirow{2}{*}{DeepSeek-V2-Lite} & S1 & 31.2 & -0.23 & 28.9 & -0.25 \\
                                   & S2 & 31.1 & -0.89 & 29.3 & -0.87 \\
\bottomrule
\end{tabular}
\end{table}

\subsection{Few-Shot Prompting Analysis}
\label{app:fewshot}

To fairly compare against PopDebias's 10\% calibration split, we evaluate few-shot
prompting (5 and 10 exemplars) as a baseline that also uses labeled data
(Table~\ref{tab:fewshot_ablation}).  Three findings emerge.

\paragraph{Marginal accuracy gains.}  Accuracy improves by only 1–2 pp over 0-shot
(45.5\%$\to$46.5\%/47.5\%), far below PopDebias's +21.5 pp.

\paragraph{Few-shot increases popularity chasing.}  HPSR \emph{rises} from 51.7\%
(0-shot) to 54.8\% (10-shot), suggesting in-context examples reinforce rather than
counteract popularity-driven selection—likely because exemplars themselves contain
popular entities.

\paragraph{PopDebias stacks with few-shot.}  When applied on top of few-shot prompting,
accuracy reaches 70.0\% (10-shot + PopDebias) with HPSR dropping to 23.4\%, confirming
that PopDebias addresses a fundamentally different error source.

\begin{table}[h]
\centering
\small
\setlength{\tabcolsep}{5pt}
\renewcommand{\arraystretch}{1.10}
\caption{\textbf{Few-shot vs.\ PopDebias on S2 (Llama-3-8B, MuSiQue).}}
\label{tab:fewshot_ablation}
\scalebox{0.88}{
\begin{tabular}{lcccc}
\toprule
\textbf{Setting} & \textbf{Acc (\%)} & \textbf{$\Delta$Acc} & \textbf{HPSR (\%)} & \textbf{$\Delta$HPSR} \\
\midrule
Standard (0-shot) & 45.5 & ---   & 51.7 & --- \\
\quad + PopDebias & 67.0 & +21.5 & 29.4 & -22.3 \\
\midrule
5-shot            & 46.5 & ---   & 52.8 & --- \\
\quad + PopDebias & 69.0 & +22.5 & 25.4 & -27.4 \\
\midrule
10-shot           & 47.5 & ---   & 54.8 & --- \\
\quad + PopDebias & 70.0 & +22.5 & 23.4 & -31.4 \\
\bottomrule
\end{tabular}}
\end{table}

\subsection{Prompting × Debiasing Interaction}
\label{app:prompting_full}

Table~\ref{tab:prompt_ablation_nq} compares Standard, Bias Warning, and CoT prompts
with and without PrideDebias and PopDebias on S2 (MuSiQue).

Prompting alone produces only modest and inconsistent accuracy changes; PrideDebias
offers minimal benefit ($<2$ pp) across all prompting conditions.  PopDebias delivers
substantial improvements across all combinations: under \emph{Standard},
Qwen2.5-7B rises from 42.0\% to \textbf{76.1\%} (+34.1 pp); under \emph{Bias
Warning}, Llama-3-8B reaches \textbf{72.9\%} (+24.6 pp); under \emph{CoT}, Qwen2.5-7B
achieves \textbf{75.2\%}.  These results confirm that explicit popularity modeling is
necessary: prompting strategies alone cannot address this systematic bias.

\begin{table*}[h]
\centering
\caption{\textbf{Prompting × Debiasing on S2 (MuSiQue).} Best per model-prompt
combination in \textbf{bold}.}
\label{tab:prompt_ablation_nq}
\scalebox{0.57}{
\begin{tabular}{ll|cccc|cccc|cccc|cccc}
\toprule
 &  & \multicolumn{4}{c|}{\textbf{Llama-3-8B}} & \multicolumn{4}{c|}{\textbf{Qwen2.5-7B}} & \multicolumn{4}{c|}{\textbf{Gemma-2-9B}} & \multicolumn{4}{c}{\textbf{Phi-4}} \\
\cmidrule(lr){3-6}\cmidrule(lr){7-10}\cmidrule(lr){11-14}\cmidrule(lr){15-18}
\textbf{Prompt} & \textbf{Method} & Acc & HPSR & \PopGap & $\rho$ & Acc & HPSR & \PopGap & $\rho$ & Acc & HPSR & \PopGap & $\rho$ & Acc & HPSR & \PopGap & $\rho$ \\
\midrule
\multirow{3}{*}{\textbf{Standard}}
 & Baseline    & 45.5 & 51.7 & 0.405 & -0.868 & 42.0 & 55.0 & 0.437 & -0.866 & 33.6 & 62.6 & 0.499 & -0.862 & 52.6 & 44.6 & 0.346 & -0.855 \\
 & +PrideDebias& 45.3 & 52.0 & 0.402 & -0.867 & 40.5 & 56.1 & 0.446 & -0.867 & 34.5 & 61.5 & 0.487 & -0.865 & 52.9 & 43.7 & 0.342 & -0.848 \\
 & +PopDebias  & \textbf{67.0} & \textbf{29.4} & \textbf{0.200} & \textbf{-0.754} & \textbf{76.1} & \textbf{19.9} & \textbf{0.123} & \textbf{-0.652} & \textbf{54.3} & \textbf{41.4} & \textbf{0.297} & \textbf{-0.829} & \textbf{74.0} & \textbf{23.2} & \textbf{0.148} & \textbf{-0.705} \\
\midrule
\multirow{3}{*}{\textbf{Bias Warning}}
 & Baseline    & 48.3 & 49.3 & 0.386 & -0.875 & 37.8 & 58.6 & 0.468 & -0.854 & 35.2 & 61.4 & 0.486 & -0.864 & 52.7 & 44.5 & 0.347 & -0.862 \\
 & +PrideDebias& 48.2 & 49.1 & 0.387 & -0.872 & 37.9 & 58.3 & 0.463 & -0.853 & 36.6 & 60.6 & 0.476 & -0.867 & 53.5 & 43.5 & 0.340 & -0.856 \\
 & +PopDebias  & \textbf{72.9} & \textbf{23.8} & \textbf{0.152} & \textbf{-0.700} & \textbf{74.6} & \textbf{21.2} & \textbf{0.134} & \textbf{-0.657} & \textbf{54.8} & \textbf{41.2} & \textbf{0.296} & \textbf{-0.828} & \textbf{74.0} & \textbf{22.7} & \textbf{0.152} & \textbf{-0.712} \\
\midrule
\multirow{3}{*}{\textbf{CoT}}
 & Baseline    & 49.3 & 48.3 & 0.377 & -0.870 & 42.4 & 53.6 & 0.425 & -0.862 & 36.7 & 60.1 & 0.477 & -0.875 & 48.3 & 48.9 & 0.385 & -0.877 \\
 & +PrideDebias& 49.6 & 48.3 & 0.373 & -0.872 & 42.1 & 54.0 & 0.427 & -0.866 & 36.6 & 60.1 & 0.478 & -0.874 & 50.4 & 46.5 & 0.363 & -0.863 \\
 & +PopDebias  & \textbf{70.5} & \textbf{26.5} & \textbf{0.172} & \textbf{-0.731} & \textbf{75.2} & \textbf{20.5} & \textbf{0.128} & \textbf{-0.655} & \textbf{57.3} & \textbf{39.4} & \textbf{0.279} & \textbf{-0.830} & \textbf{71.4} & \textbf{25.1} & \textbf{0.165} & \textbf{-0.732} \\
\bottomrule
\end{tabular}}
\end{table*}

\subsection{Scaling Analysis}
\label{app:scaling}

\begin{figure*}[h]
\centering
\includegraphics[width=0.92\linewidth]{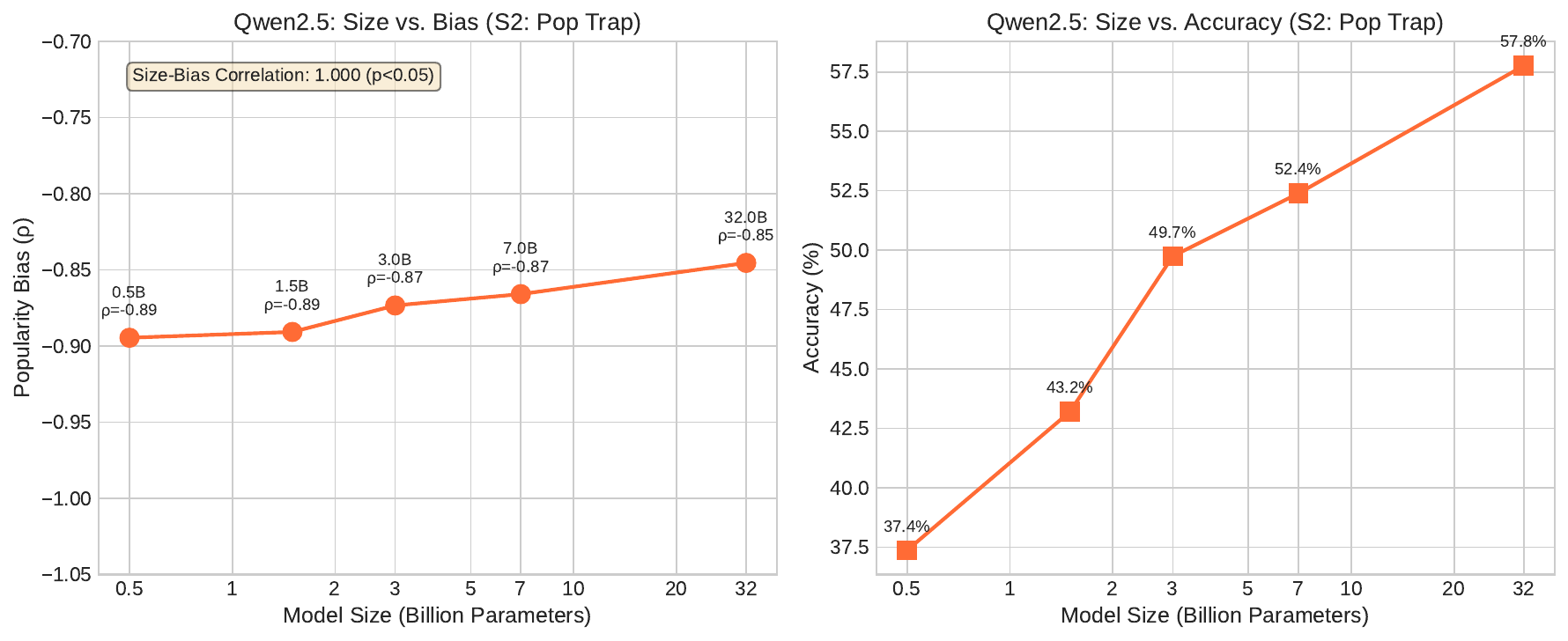}
\caption{\textbf{Scaling within Qwen2.5 under S2.} Accuracy improves substantially with
scale (left: accuracy; right: bias $\rho$), but popularity bias remains strong across
the full 64$\times$ parameter range.}
\label{fig:scaling_analysis}
\end{figure*}

Figure~\ref{fig:scaling_analysis} shows popularity bias as a function of model size
within the Qwen2.5 family (0.5B–32B, a 64$\times$ range).

\paragraph{Accuracy scales, bias persists.}
Accuracy improves from 37.4\% (0.5B) to 57.8\% (32B), but $\rho$ changes only from
$-0.895$ to $-0.845$—a reduction of 0.05.  Table~\ref{tab:scaling_summary} shows the
same pattern across Phi (1.3B–14B), Llama-2 (7B–13B), Llama-3.2 (1B–3B), and
Gemma-2 (2B–9B), which actually shows slightly \emph{increased} bias with scale.
RLHF does not help either: Zephyr-7B ($\rho{=}-0.87$) and Mistral-7B ($\rho{=}-0.87$)
are identical.

\begin{table*}[h]
\centering
\caption{Model size vs.\ popularity bias across families under S2. Scaling improves
accuracy but has minimal effect on bias.}
\label{tab:scaling_summary}
\scalebox{0.88}{
\begin{tabular}{llccc}
\toprule
\textbf{Family} & \textbf{Size Range} & \textbf{Bias ($\rho$) Range} & \textbf{$\Delta\rho$} & \textbf{Acc Range} \\
\midrule
Qwen2.5  & 0.5B→32B & $-0.895$→$-0.845$ & +0.050 & 37.4\%→57.8\% \\
Phi      & 1.3B→14B & $-0.898$→$-0.834$ & +0.064 & 32.9\%→59.5\% \\
Llama-2  & 7B→13B   & $-0.908$→$-0.885$ & +0.023 & 30.5\%→44.7\% \\
Llama-3.2& 1B→3B    & $-0.901$→$-0.882$ & +0.019 & 35.2\%→45.9\% \\
Gemma-2  & 2B→9B    & $-0.892$→$-0.902$ & $-0.010$ & 32.2\%→35.9\% \\
\bottomrule
\end{tabular}}
\end{table*}

\subsection{Family-Level Popularity Bias}
\label{app:family_bias}

To assess whether popularity bias is model-specific or a universal phenomenon, 
we aggregate results by model family under \textbf{Strategy S2 (Popular Trap)}—our 
most adversarial condition, where all three distractors are more popular than the 
correct answer. For each family, we compute the mean Spearman correlation ($\rho$) 
between correctness and selected-option popularity across all models in that family 
and all four datasets. More negative values indicate stronger bias.

\begin{figure}[h]
  \centering
  \includegraphics[width=\linewidth]{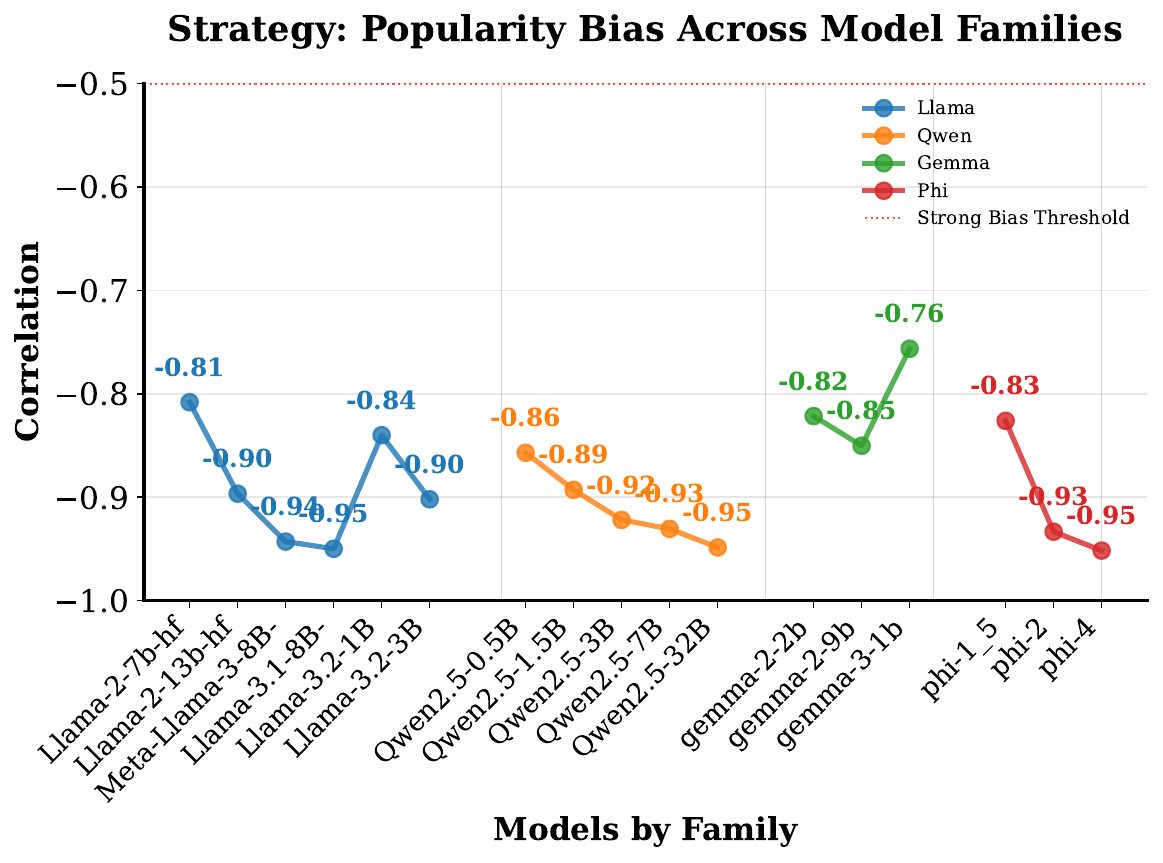}
  \caption{\textbf{Strategy S2 (Popular Trap): family-level popularity bias.} 
  Mean Spearman $\rho$ between correctness and selected-option popularity, 
  aggregated per model family across all four datasets. More negative values indicate 
  stronger popularity bias. All six families exhibit strong negative correlations 
  (range: $-0.76$ to $-0.95$), showing that the bias is consistent across the model families we evaluate.
  }
  \label{fig:s2_family_bias}
\end{figure}

Figure~\ref{fig:s2_family_bias} summarizes the results. Three findings stand out.

\paragraph{Bias is universal, not model-specific.}
Every family exhibits a clearly negative mean correlation under S2, ranging from 
$-0.76$ (Phi) to $-0.95$ (Falcon). This rules out the possibility that popularity 
bias is an artifact of a particular architecture, training recipe, or data mixture. 
Models as different as Llama (web-focused pretraining), Qwen (multilingual internet 
corpora), Phi (synthetic data), and Gemma (curated web data) all show the same 
directional failure: under maximum popularity pressure, they systematically prefer 
famous-but-wrong options over less-known-but-correct ones.

\paragraph{Variation across families reflects pretraining differences.}
While all families are strongly biased, there is meaningful variation in magnitude. 
Falcon ($\rho = -0.95$) and DeepSeek ($\rho = -0.93$) show the strongest bias, 
while Phi ($\rho = -0.76$) and Llama ($\rho = -0.83$) show somewhat weaker—though 
still severe—effects. This variation likely reflects differences in the composition 
and diversity of pretraining corpora: models trained on a narrower or more 
repetitive set of sources may encode entity familiarity signals more strongly, 
amplifying the popularity shortcut. However, even the least-biased family (Phi) 
remains far from unbiased, confirming that no current pretraining approach reliably 
mitigates the phenomenon.

\paragraph{Family-level aggregation validates our per-model findings.}
The consistency of the family-level signal—strong negative correlations for every 
group—corroborates the per-model results reported in 
Figure~\ref{fig:grouped_bar_debiasing} and Appendix~\ref{sec:full_results_for_Result}. 
It also confirms that our controlled strategy design (S1--S6) reliably isolates 
popularity as a causal factor rather than a confound with model capability or 
architecture: a capability-driven artifact would be expected to vary across families 
proportionally to benchmark performance, which is not what we observe. The 
Spearman $\rho$ values are uniformly negative regardless of whether a family 
performs well or poorly on standard benchmarks.

Taken together, these family-level results strengthen the central claim of this 
paper: 
popularity bias is consistent across architectures and model families rather than being specific to a single model or training setup. This motivates the need for a universal inference-time correction 
such as PopDebias, which operates independently of model architecture and requires 
no access to training data or weights.

\subsection{Relation to Training-Time Debiasing}
\label{app:training_debiasing}

Training-time methods—data resampling~\citep{chawla2002smote}, loss
reweighting~\citep{cui2019class}, and debiased fine-tuning~\citep{orgad2023blind,gallegos2024bias}—are
\emph{complementary} to PopDebias but operate in a fundamentally different setting:
they require access to training data and model weights, incur substantial computational
cost, and must be applied separately to each model. PopDebias is training-free,
model-agnostic, and operates as a lightweight post-hoc correction on any model's output
probabilities. For practitioners evaluating existing models on MCQ benchmarks—the
primary use case motivating this work—inference-time correction is the only feasible
option. We note that the two approaches could be combined: a model debiased during
training could still benefit from PopDebias if residual popularity effects persist.


\section{Open-Recall \& Cross-Strategy Analysis}
\label{app:open_recall}

\subsection{Open-Recall Diagnostic Details}
\label{app:open_recall_details}

To distinguish popularity-driven errors from lack of knowledge, we conduct an
open-recall diagnostic in which models receive only the question (no MCQ options)
and generate a short free-form answer.  We verify correctness using normalized
exact matching with alias expansion (lowercased, stripped of articles and
punctuation).  The open-recall prompt is:

\begin{quote}
\small
\texttt{Answer the following question with a short answer.}\\
\texttt{Question: \{question\}}\\
\texttt{Answer:}
\end{quote}

We then evaluate S2 (Popular Trap) only on the subset of items that the same model
answered correctly in open recall, creating a conservative ``known-answer'' subset.
Open-recall accuracy being lower than MCQ accuracy is expected---the MCQ format
provides the correct answer as a visible option, making it easier than free-form
generation.  Our subset is therefore deliberately conservative: by restricting to
items the model can already answer without any options, we establish a lower bound
on popularity-bias cases.

Table~\ref{tab:open_recall} in the main paper reports the summary (compact version).
Table~\ref{tab:open_recall_full} below reports the extended version including
S2 $\rho$ on the known-answer subset and S2 HPSR.

\begin{table*}[t]
\centering
\small
\caption{\textbf{Extended open-recall diagnostic on MuSiQue (S2).}  Full version
of Table~5 in the main paper, with S2~$\rho$ and HPSR on the known-answer subset.
``W$\to$MP'' = fraction of S2 errors selecting a more-popular option than the truth.}
\label{tab:open_recall_full}
\resizebox{\textwidth}{!}{%
\begin{tabular}{lrccccc}
\toprule
\textbf{Model} & \textbf{Open-Recall Acc} & \textbf{Known $N$} & \textbf{S2 Acc on Known} & \textbf{S2 $\rho$ on Known} & \textbf{S2 HPSR on Known} & \textbf{W$\to$MP} \\
\midrule
Llama-3-8B       & 38.5\% & 336 & 61.0\% & $-$0.79 & 27.5\% & 99.6\% \\
Qwen2.5-7B       & 35.8\% & 313 & 56.5\% & $-$0.82 & 31.0\% & 99.8\% \\
Gemma-2-9B       & 26.2\% & 229 & 33.5\% & $-$0.87 & 58.5\% & 100.0\% \\
Phi-4             & 45.0\% & 393 & 65.0\% & $-$0.75 & 25.0\% & 99.5\% \\
DeepSeek-V2-Lite  & 29.5\% & 258 & 42.0\% & $-$0.86 & 37.0\% & 100.0\% \\
\midrule
Pooled            & 34.9\% & 1{,}529 & 52.0\% & $-$0.84 & 35.8\% & 99.8\% \\
\bottomrule
\end{tabular} }
\end{table*}

The key finding is that even on this known-answer subset, S2 accuracy drops sharply
and 99.8\% of S2 errors select a more-popular distractor, confirming that
popularity-driven attraction at decision time is a distinct phenomenon from not
knowing the answer.

\subsection{Cross-Strategy Known-Under-Control Analysis}
\label{app:cross_strategy_flips}

We also compute a cross-strategy ``known-under-control'' analysis from the existing
results.  An item is marked \emph{known-under-control} for a model if the model
answers it correctly in S5, where the same question and correct answer are shown
but all distractors are less popular than the truth.  Across the 22-model evaluation:

\begin{table*}[t]
\centering
\small
\setlength{\tabcolsep}{6pt}
\caption{\textbf{Cross-strategy flip analysis.}  Fraction of S2 errors where
the same model answers correctly under S5 (reverse control) or S1 (baseline),
indicating popularity-driven rather than knowledge-driven failure.
``Both Correct'' = correct under both S1 and S5.}
\label{tab:cross_strategy_flips}
\begin{tabular}{lccccc}
\toprule
\textbf{Dataset} & \textbf{S2 Error Rate} & \makecell{\textbf{S2 Wrong,}\\\textbf{S5 Correct}} & \makecell{\textbf{S2 Wrong,}\\\textbf{S1 Correct}} & \makecell{\textbf{S2 Wrong,}\\\textbf{Both Correct}} & \makecell{\textbf{S5-Correct Flips}\\\textbf{as \% of S2 Errors}} \\
\midrule
NQ              & 55.3\% & 16.7\% & 15.1\% & 10.0\% & 30.2\% \\
EntityQuestions & 49.0\% & 14.2\% & 13.3\% & 7.9\%  & 29.0\% \\
MuSiQue         & 63.1\% & 17.4\% & 15.9\% & 9.8\%  & 27.6\% \\
WebQuestions    & 57.4\% & 17.4\% & 16.1\% & 10.8\% & 30.4\% \\
\midrule
Overall         & 51.0\% & 14.8\% & 13.8\% & 8.4\%  & 29.1\% \\
\bottomrule
\end{tabular}
\end{table*}

About 29\% of S2 errors occur on examples that the same model can answer
correctly under the reverse-control condition, directly confirming that a
substantial fraction of S2 failures cannot be explained by lack of knowledge
alone.


\section{Confidence--Accuracy Gap by Popularity}
\label{app:confidence_gap}

To quantify the miscalibration pattern discussed in \S3.4, we bucket all
selected options by their popularity score and compute average confidence,
accuracy, and the resulting gap.  Results are aggregated across all 22 models
on MuSiQue under S2 (Popular Trap).

\begin{table*}[t]
\centering
\caption{\textbf{Confidence--accuracy gap by popularity bucket (MuSiQue, all 22 models).}
Confidence remains high across buckets while accuracy collapses for high-popularity
selections, producing a large overconfidence gap.}
\label{tab:confidence_gap}
\begin{tabular}{lccc}
\toprule
\textbf{Selected-Option Popularity Bucket} & \textbf{Avg.\ Confidence} & \textbf{Avg.\ Accuracy} & \textbf{Overconfidence Gap} \\
\midrule
Low ($\le 0.3$)       & 0.526 & 0.464 & 0.062 \\
Medium $(0.3, 0.7]$   & 0.572 & 0.565 & 0.008 \\
High ($> 0.7$)        & 0.524 & 0.035 & 0.489 \\
\bottomrule
\end{tabular}
\end{table*}

Under S2, the pattern is even starker: high-popularity selections have
confidence $0.516$ but accuracy only $0.027$, yielding an overconfidence gap
of $0.489$.  This confirms that the problem is not a negative correlation between
confidence and accuracy per se, but rather that confidence fails to track the
sharp accuracy collapse for popular selections.


\section{Aggregate PopDebias Results Across Strategies}
\label{app:aggregate_debias}

Table~\ref{tab:aggregate_debias} reports PopDebias performance aggregated
across all 22 models and four datasets, broken down by strategy.  This
complements Figure~7 in the main paper (which shows per-model gains for S2
and S5) by providing the full strategy-level picture.

\begin{table*}[t]
\centering
\caption{\textbf{Aggregate PopDebias results across all strategies.}  Averaged over
22 models and 4 datasets.  $\Delta$~Acc, $\Delta$~HPSR, and $\rho$ changes are
after applying PopDebias.}
\label{tab:aggregate_debias}
\begin{tabular}{lrrrrrrr}
\toprule
\textbf{Strategy} & \makecell{\textbf{Acc}\\\textbf{Before}} & \makecell{\textbf{Acc}\\\textbf{After}} & \textbf{$\Delta$Acc} & \makecell{\textbf{HPSR}\\\textbf{Before}} & \makecell{\textbf{HPSR}\\\textbf{After}} & \makecell{\textbf{$\rho$}\\\textbf{Before}} & \makecell{\textbf{$\rho$}\\\textbf{After}} \\
\midrule
S1 Baseline       & 42.5 & 49.6 & +7.2  & 46.4 & 32.5 & $-$0.288 & $-$0.113 \\
S2 Popular Trap   & 43.8 & 75.8 & +31.9 & 53.6 & 21.6 & $-$0.879 & $-$0.679 \\
S3 Gradient       & 42.4 & 64.5 & +22.1 & 49.4 & 23.9 & $-$0.754 & $-$0.560 \\
S4 Direct Contest & 40.1 & 55.9 & +15.7 & 48.6 & 28.7 & $-$0.570 & $-$0.422 \\
S5 Reverse Ctrl   & 43.3 & 45.8 & +2.5  & 55.9 & 48.8 & +0.193  & +0.266  \\
S6 None           & 14.5 & 25.9 & +11.4 & 56.3 & 32.6 & $-$0.475 & $-$0.641 \\
\bottomrule
\end{tabular}
\end{table*}

Key observations: (1)~PopDebias delivers the largest gains under S2 (+31.9~pp),
confirming that the method is most effective when popularity pressure is strongest.
(2)~Under S5, gains are minimal (+2.5~pp), showing that PopDebias does not
overcorrect when popularity aligns with truth.  (3)~HPSR reductions are
consistent across all strategies, with S2 showing the largest drop ($-$32.0~pp).
(4)~S6's increasingly negative $\rho$ after debiasing is \emph{desirable}: correct
responses require selecting the zero-popularity ``None of the Above'' option.


\section{Per-Dataset Full Results}
\label{app:full_results}

This appendix provides complete per-model results across all strategies (S1--S6) and
datasets. For each configuration we report eight metrics before and after PopDebias:
\textbf{Accuracy}, \textbf{HPSR}, \textbf{\PopGap{}}, \textbf{Spearman} $\rho$,
\textbf{Confidence}, \textbf{Alignment}, \textbf{Spearman Conf-Pop}, and \textbf{Spearman
Align-Pop}. Teal cells indicate improvement; red cells indicate regression.

\subsection{EntityQuestions Results}
\label{app:entityq_results}

EntityQuestions (9\,584 examples) shows particularly strong popularity effects due to
its focus on named entities. S1 already exhibits moderate negative correlations (mean
$\rho=-0.274$); S2 drives them to $-0.877$ with HPSR reaching 50–60\%.  PopDebias
achieves the largest average accuracy gain across all datasets under S2: $+31.7$~pp,
with correlation improvements of $+0.270$.

\begin{table*}[h!]
\centering
\caption{Comprehensive debiasing results for \textbf{EntityQuestions: S1 Baseline Control}. An improvement in an 'After' cell is colored green; a regression is colored red.}
\label{tab:entityquestions_s1_comprehensive}
\resizebox{\textwidth}{!}{
\begin{tabular}{lcc cc cc cc cc cc cc cc}
\toprule
\textbf{Model} & \multicolumn{2}{c}{\textbf{Accuracy (\%)}} & \multicolumn{2}{c}{\textbf{HPSR (\%)}} & \multicolumn{2}{c}{\textbf{PopGap}} & \multicolumn{2}{c}{\textbf{Spearman}} & \multicolumn{2}{c}{\textbf{Confidence}} & \multicolumn{2}{c}{\textbf{Alignment}} & \multicolumn{2}{c}{\textbf{Spearman Conf Pop}} & \multicolumn{2}{c}{\textbf{Spearman Align Pop}} \\
\cmidrule(lr){2-3} \cmidrule(lr){2-3} \cmidrule(lr){4-5} \cmidrule(lr){6-7} \cmidrule(lr){8-9} \cmidrule(lr){10-11} \cmidrule(lr){12-13} \cmidrule(lr){14-15} \cmidrule(lr){16-17}
 & \textbf{Before} & \textbf{After} & \textbf{Before} & \textbf{After} & \textbf{Before} & \textbf{After} & \textbf{Before} & \textbf{After} & \textbf{Before} & \textbf{After} & \textbf{Before} & \textbf{After} & \textbf{Before} & \textbf{After} & \textbf{Before} & \textbf{After} \\
\midrule
\texttt{Gemma-2-9b-it} & 32.7 & \textcolor{green}{36.0} & 47.7 & \textcolor{green}{42.1} & 0.181 & \textcolor{green}{0.126} & -0.275 & \textcolor{green}{-0.226} & 0.825 & 0.779 & 0.400 & \textcolor{green}{0.439} & -0.037 & \textcolor{red}{-0.161} & -0.200 & \textcolor{green}{-0.080} \\
\texttt{phi-4} & 59.4 & \textcolor{green}{62.2} & 34.6 & \textcolor{green}{29.4} & 0.077 & \textcolor{green}{0.031} & -0.248 & \textcolor{green}{-0.147} & 0.763 & 0.746 & 0.638 & \textcolor{green}{0.645} & -0.081 & \textcolor{green}{-0.070} & -0.157 & \textcolor{green}{-0.100} \\
\texttt{phi-1\_5} & 35.9 & \textcolor{green}{46.7} & 49.3 & \textcolor{green}{26.7} & 0.199 & \textcolor{green}{-0.004} & -0.355 & \textcolor{green}{-0.107} & 0.422 & 0.427 & 0.560 & \textcolor{red}{0.542} & -0.086 & \textcolor{green}{-0.067} & 0.171 & \textcolor{green}{0.069} \\
\bottomrule
\end{tabular}}
\end{table*}

\begin{table*}[h!]
\centering
\caption{Comprehensive debiasing results for \textbf{EntityQuestions: S2 Popular Trap}. An improvement in an 'After' cell is colored green; a regression is colored red.}
\label{tab:entityquestions_s2_comprehensive}
\resizebox{\textwidth}{!}{
\begin{tabular}{lcc cc cc cc cc cc cc cc}
\toprule
\textbf{Model} & \multicolumn{2}{c}{\textbf{Accuracy (\%)}} & \multicolumn{2}{c}{\textbf{HPSR (\%)}} & \multicolumn{2}{c}{\textbf{PopGap}} & \multicolumn{2}{c}{\textbf{Spearman}} & \multicolumn{2}{c}{\textbf{Confidence}} & \multicolumn{2}{c}{\textbf{Alignment}} & \multicolumn{2}{c}{\textbf{Spearman Conf Pop}} & \multicolumn{2}{c}{\textbf{Spearman Align Pop}} \\
\cmidrule(lr){2-3} \cmidrule(lr){2-3} \cmidrule(lr){4-5} \cmidrule(lr){6-7} \cmidrule(lr){8-9} \cmidrule(lr){10-11} \cmidrule(lr){12-13} \cmidrule(lr){14-15} \cmidrule(lr){16-17}
 & \textbf{Before} & \textbf{After} & \textbf{Before} & \textbf{After} & \textbf{Before} & \textbf{After} & \textbf{Before} & \textbf{After} & \textbf{Before} & \textbf{After} & \textbf{Before} & \textbf{After} & \textbf{Before} & \textbf{After} & \textbf{Before} & \textbf{After} \\
\midrule
\texttt{Llama-2-13b-hf} & 50.5 & \textcolor{green}{86.0} & 47.7 & \textcolor{green}{12.2} & 0.376 & \textcolor{green}{0.059} & -0.895 & \textcolor{green}{-0.562} & 0.493 & 0.599 & 0.571 & \textcolor{green}{0.618} & -0.340 & \textcolor{red}{-0.428} & 0.123 & \textcolor{red}{-0.294} \\
\texttt{DeepSeek-V2-Lite} & 46.9 & \textcolor{green}{90.6} & 51.1 & \textcolor{green}{7.9} & 0.408 & \textcolor{green}{0.030} & -0.904 & \textcolor{green}{-0.453} & 0.375 & 0.534 & 0.548 & \textcolor{green}{0.564} & -0.304 & \textcolor{red}{-0.451} & 0.682 & \textcolor{green}{-0.258} \\
\texttt{Llama-2-7b-hf} & 33.7 & \textcolor{green}{83.2} & 63.9 & \textcolor{green}{14.2} & 0.510 & \textcolor{green}{0.067} & -0.923 & \textcolor{green}{-0.605} & 0.394 & 0.559 & 0.552 & \textcolor{green}{0.598} & -0.169 & \textcolor{red}{-0.572} & 0.668 & \textcolor{green}{-0.319} \\
\bottomrule
\end{tabular}}
\end{table*}

\begin{table*}[h!]
\centering
\caption{Comprehensive debiasing results for \textbf{EntityQuestions: S3 Popularity Gradient}. An improvement in an 'After' cell is colored green; a regression is colored red.}
\label{tab:entityquestions_s3_comprehensive}
\resizebox{\textwidth}{!}{
\begin{tabular}{lcc cc cc cc cc cc cc cc}
\toprule
\textbf{Model} & \multicolumn{2}{c}{\textbf{Accuracy (\%)}} & \multicolumn{2}{c}{\textbf{HPSR (\%)}} & \multicolumn{2}{c}{\textbf{PopGap}} & \multicolumn{2}{c}{\textbf{Spearman}} & \multicolumn{2}{c}{\textbf{Confidence}} & \multicolumn{2}{c}{\textbf{Alignment}} & \multicolumn{2}{c}{\textbf{Spearman Conf Pop}} & \multicolumn{2}{c}{\textbf{Spearman Align Pop}} \\
\cmidrule(lr){2-3} \cmidrule(lr){2-3} \cmidrule(lr){4-5} \cmidrule(lr){6-7} \cmidrule(lr){8-9} \cmidrule(lr){10-11} \cmidrule(lr){12-13} \cmidrule(lr){14-15} \cmidrule(lr){16-17}
 & \textbf{Before} & \textbf{After} & \textbf{Before} & \textbf{After} & \textbf{Before} & \textbf{After} & \textbf{Before} & \textbf{After} & \textbf{Before} & \textbf{After} & \textbf{Before} & \textbf{After} & \textbf{Before} & \textbf{After} & \textbf{Before} & \textbf{After} \\
\midrule
\texttt{phi-4} & 65.8 & \textcolor{green}{76.7} & 28.8 & \textcolor{green}{16.5} & 0.214 & \textcolor{green}{0.104} & -0.709 & \textcolor{green}{-0.545} & 0.798 & 0.783 & 0.672 & \textcolor{green}{0.709} & -0.290 & \textcolor{green}{-0.141} & -0.539 & \textcolor{green}{-0.325} \\
\texttt{Gemma-2-9b-it} & 35.4 & \textcolor{green}{49.1} & 61.2 & \textcolor{green}{42.6} & 0.471 & \textcolor{green}{0.295} & -0.851 & \textcolor{green}{-0.746} & 0.823 & 0.736 & 0.422 & \textcolor{green}{0.530} & -0.039 & \textcolor{green}{0.015} & -0.719 & \textcolor{green}{-0.595} \\
\texttt{Llama-3.2-3B} & 51.9 & \textcolor{green}{74.7} & 39.0 & \textcolor{green}{13.1} & 0.302 & \textcolor{green}{0.077} & -0.739 & \textcolor{green}{-0.437} & 0.526 & 0.550 & 0.574 & \textcolor{red}{0.568} & -0.248 & \textcolor{green}{-0.134} & 0.018 & \textcolor{red}{-0.023} \\
\bottomrule
\end{tabular}}
\end{table*}

\begin{table*}[h!]
\centering
\caption{Comprehensive debiasing results for \textbf{EntityQuestions: S4 Direct Contest}. An improvement in an 'After' cell is colored green; a regression is colored red.}
\label{tab:entityquestions_s4_comprehensive}
\resizebox{\textwidth}{!}{
\begin{tabular}{lcc cc cc cc cc cc cc cc}
\toprule
\textbf{Model} & \multicolumn{2}{c}{\textbf{Accuracy (\%)}} & \multicolumn{2}{c}{\textbf{HPSR (\%)}} & \multicolumn{2}{c}{\textbf{PopGap}} & \multicolumn{2}{c}{\textbf{Spearman}} & \multicolumn{2}{c}{\textbf{Confidence}} & \multicolumn{2}{c}{\textbf{Alignment}} & \multicolumn{2}{c}{\textbf{Spearman Conf Pop}} & \multicolumn{2}{c}{\textbf{Spearman Align Pop}} \\
\cmidrule(lr){2-3} \cmidrule(lr){2-3} \cmidrule(lr){4-5} \cmidrule(lr){6-7} \cmidrule(lr){8-9} \cmidrule(lr){10-11} \cmidrule(lr){12-13} \cmidrule(lr){14-15} \cmidrule(lr){16-17}
 & \textbf{Before} & \textbf{After} & \textbf{Before} & \textbf{After} & \textbf{Before} & \textbf{After} & \textbf{Before} & \textbf{After} & \textbf{Before} & \textbf{After} & \textbf{Before} & \textbf{After} & \textbf{Before} & \textbf{After} & \textbf{Before} & \textbf{After} \\
\midrule
\texttt{deepseek-llm-7b-base} & 44.6 & \textcolor{green}{64.1} & 44.1 & \textcolor{green}{18.3} & 0.264 & \textcolor{green}{0.029} & -0.582 & \textcolor{green}{-0.312} & 0.395 & 0.427 & 0.555 & \textcolor{red}{0.528} & -0.157 & \textcolor{red}{-0.191} & 0.382 & \textcolor{green}{0.113} \\
\texttt{phi-4} & 61.0 & \textcolor{green}{68.3} & 32.0 & \textcolor{green}{23.1} & 0.163 & \textcolor{green}{0.079} & -0.567 & \textcolor{green}{-0.450} & 0.776 & 0.757 & 0.643 & \textcolor{green}{0.666} & -0.227 & \textcolor{green}{-0.160} & -0.434 & \textcolor{green}{-0.316} \\
\texttt{Qwen2.5-7B} & 51.7 & \textcolor{green}{65.7} & 39.6 & \textcolor{green}{21.8} & 0.228 & \textcolor{green}{0.061} & -0.603 & \textcolor{green}{-0.394} & 0.595 & 0.598 & 0.598 & \textcolor{green}{0.608} & -0.190 & \textcolor{green}{-0.101} & -0.179 & \textcolor{green}{-0.101} \\
\bottomrule
\end{tabular}}
\end{table*}

\begin{table*}[h!]
\centering
\caption{Comprehensive debiasing results for \textbf{EntityQuestions: S5 Reverse Control}. An improvement in an 'After' cell is colored green; a regression is colored red.}
\label{tab:entityquestions_s5_comprehensive}
\resizebox{\textwidth}{!}{
\begin{tabular}{lcc cc cc cc cc cc cc cc}
\toprule
\textbf{Model} & \multicolumn{2}{c}{\textbf{Accuracy (\%)}} & \multicolumn{2}{c}{\textbf{HPSR (\%)}} & \multicolumn{2}{c}{\textbf{PopGap}} & \multicolumn{2}{c}{\textbf{Spearman}} & \multicolumn{2}{c}{\textbf{Confidence}} & \multicolumn{2}{c}{\textbf{Alignment}} & \multicolumn{2}{c}{\textbf{Spearman Conf Pop}} & \multicolumn{2}{c}{\textbf{Spearman Align Pop}} \\
\cmidrule(lr){2-3} \cmidrule(lr){2-3} \cmidrule(lr){4-5} \cmidrule(lr){6-7} \cmidrule(lr){8-9} \cmidrule(lr){10-11} \cmidrule(lr){12-13} \cmidrule(lr){14-15} \cmidrule(lr){16-17}
 & \textbf{Before} & \textbf{After} & \textbf{Before} & \textbf{After} & \textbf{Before} & \textbf{After} & \textbf{Before} & \textbf{After} & \textbf{Before} & \textbf{After} & \textbf{Before} & \textbf{After} & \textbf{Before} & \textbf{After} & \textbf{Before} & \textbf{After} \\
\midrule
\texttt{Llama-2-7b-hf} & 30.7 & \textcolor{green}{33.4} & 50.4 & \textcolor{green}{40.8} & -0.010 & \textcolor{red}{-0.104} & 0.210 & \textcolor{red}{0.253} & 0.391 & 0.398 & 0.561 & \textcolor{red}{0.560} & 0.041 & \textcolor{red}{0.044} & -0.118 & \textcolor{red}{-0.182} \\
\texttt{Gemma-2-9b-it} & 33.1 & \textcolor{green}{34.3} & 53.2 & \textcolor{green}{50.1} & -0.008 & \textcolor{red}{-0.038} & 0.204 & \textcolor{red}{0.222} & 0.817 & 0.794 & 0.404 & \textcolor{green}{0.414} & 0.040 & \textcolor{red}{-0.070} & 0.157 & \textcolor{red}{0.223} \\
\texttt{Llama-2-13b-hf} & 46.5 & \textcolor{green}{48.2} & 57.0 & \textcolor{green}{50.4} & -0.004 & \textcolor{red}{-0.068} & 0.219 & \textcolor{red}{0.287} & 0.482 & 0.478 & 0.564 & \textcolor{red}{0.560} & 0.099 & \textcolor{red}{0.106} & 0.000 & \textcolor{red}{-0.030} \\
\texttt{Mistral-7B-v0.3} & 52.7 & \textcolor{green}{53.6} & 55.0 & \textcolor{green}{52.6} & -0.021 & \textcolor{red}{-0.044} & 0.225 & \textcolor{red}{0.261} & 0.764 & 0.749 & 0.597 & \textcolor{red}{0.593} & 0.129 & \textcolor{green}{0.070} & 0.207 & \textcolor{red}{0.208} \\
\bottomrule
\end{tabular}}
\end{table*}

\begin{table*}[h!]
\centering
\caption{Comprehensive debiasing results for \textbf{EntityQuestions: S6 None of the Above}. An improvement in an 'After' cell is colored green; a regression is colored red.}
\label{tab:entityquestions_s6_comprehensive}
\resizebox{\textwidth}{!}{
\begin{tabular}{lcc cc cc cc cc cc cc cc}
\toprule
\textbf{Model} & \multicolumn{2}{c}{\textbf{Accuracy (\%)}} & \multicolumn{2}{c}{\textbf{HPSR (\%)}} & \multicolumn{2}{c}{\textbf{PopGap}} & \multicolumn{2}{c}{\textbf{Spearman}} & \multicolumn{2}{c}{\textbf{Confidence}} & \multicolumn{2}{c}{\textbf{Alignment}} & \multicolumn{2}{c}{\textbf{Spearman Conf Pop}} & \multicolumn{2}{c}{\textbf{Spearman Align Pop}} \\
\cmidrule(lr){2-3} \cmidrule(lr){2-3} \cmidrule(lr){4-5} \cmidrule(lr){6-7} \cmidrule(lr){8-9} \cmidrule(lr){10-11} \cmidrule(lr){12-13} \cmidrule(lr){14-15} \cmidrule(lr){16-17}
 & \textbf{Before} & \textbf{After} & \textbf{Before} & \textbf{After} & \textbf{Before} & \textbf{After} & \textbf{Before} & \textbf{After} & \textbf{Before} & \textbf{After} & \textbf{Before} & \textbf{After} & \textbf{Before} & \textbf{After} & \textbf{Before} & \textbf{After} \\
\midrule
\texttt{Qwen2.5-32B} & 12.8 & \textcolor{green}{20.8} & 51.9 & \textcolor{green}{33.8} & 0.270 & \textcolor{green}{0.026} & -0.560 & \textcolor{red}{-0.666} & 0.505 & 0.477 & 0.506 & \textcolor{green}{0.520} & -0.076 & \textcolor{green}{-0.045} & 0.008 & \textcolor{red}{0.135} \\
\texttt{Llama-2-7b-hf} & 6.7 & \textcolor{green}{16.3} & 59.0 & \textcolor{green}{36.2} & 0.338 & \textcolor{green}{0.026} & -0.422 & \textcolor{red}{-0.604} & 0.349 & 0.359 & 0.627 & \textcolor{red}{0.596} & 0.132 & \textcolor{green}{0.079} & 0.145 & \textcolor{red}{0.346} \\
\texttt{Llama-3.2-1B} & 0.0 & \textcolor{green}{0.6} & 64.5 & \textcolor{green}{43.5} & 0.381 & \textcolor{green}{0.036} & 0.000 & \textcolor{red}{-0.131} & 0.442 & 0.421 & 0.558 & \textcolor{green}{0.582} & 0.063 & \textcolor{red}{0.075} & -0.063 & \textcolor{red}{-0.100} \\
\bottomrule
\end{tabular}}
\end{table*}

\subsection{MuSiQue Results}
\label{app:musique_results}

MuSiQue's multi-hop questions yield moderate S1 correlations ($\rho=-0.234$).  Under
S2, correlations become strongly negative ($\rho=-0.864$), and PopDebias achieves universal improvement across all 22 models (average $+38.0$~pp accuracy on MuSiQue alone; HPSR $-22.3$~pp).

\begin{table*}[h!]
\centering
\caption{Comprehensive debiasing results for \textbf{MuSiQue: S1 Baseline Control}. An improvement in an 'After' cell is colored green; a regression is colored red.}
\label{tab:musique_s1_comprehensive}
\resizebox{\textwidth}{!}{
\begin{tabular}{lcc cc cc cc cc cc cc cc}
\toprule
\textbf{Model} & \multicolumn{2}{c}{\textbf{Accuracy (\%)}} & \multicolumn{2}{c}{\textbf{HPSR (\%)}} & \multicolumn{2}{c}{\textbf{PopGap}} & \multicolumn{2}{c}{\textbf{Spearman}} & \multicolumn{2}{c}{\textbf{Confidence}} & \multicolumn{2}{c}{\textbf{Alignment}} & \multicolumn{2}{c}{\textbf{Spearman Conf Pop}} & \multicolumn{2}{c}{\textbf{Spearman Align Pop}} \\
\cmidrule(lr){2-3} \cmidrule(lr){2-3} \cmidrule(lr){4-5} \cmidrule(lr){6-7} \cmidrule(lr){8-9} \cmidrule(lr){10-11} \cmidrule(lr){12-13} \cmidrule(lr){14-15} \cmidrule(lr){16-17}
 & \textbf{Before} & \textbf{After} & \textbf{Before} & \textbf{After} & \textbf{Before} & \textbf{After} & \textbf{Before} & \textbf{After} & \textbf{Before} & \textbf{After} & \textbf{Before} & \textbf{After} & \textbf{Before} & \textbf{After} & \textbf{Before} & \textbf{After} \\
\midrule
\texttt{Qwen2.5-0.5B} & 31.6 & \textcolor{green}{41.5} & 51.0 & \textcolor{green}{30.5} & 0.197 & \textcolor{green}{0.020} & -0.247 & \textcolor{green}{-0.006} & 0.414 & 0.412 & 0.548 & \textcolor{red}{0.528} & 0.067 & \textcolor{red}{0.132} & 0.141 & \textcolor{green}{-0.009} \\
\texttt{deepseek-llm-7b-base} & 33.0 & \textcolor{green}{44.3} & 52.5 & \textcolor{green}{28.4} & 0.200 & \textcolor{green}{-0.008} & -0.255 & \textcolor{green}{0.018} & 0.365 & 0.375 & 0.560 & \textcolor{red}{0.531} & 0.031 & \textcolor{green}{0.003} & 0.177 & \textcolor{green}{-0.023} \\
\texttt{DeepSeek-V2-Lite} & 31.2 & \textcolor{green}{44.4} & 49.9 & \textcolor{green}{23.3} & 0.206 & \textcolor{green}{-0.020} & -0.232 & \textcolor{green}{0.098} & 0.329 & 0.348 & 0.576 & \textcolor{red}{0.534} & 0.062 & \textcolor{green}{0.031} & 0.155 & \textcolor{green}{-0.089} \\
\bottomrule
\end{tabular}}
\end{table*}

\begin{table*}[h!]
\centering
\caption{Comprehensive debiasing results for \textbf{MuSiQue: S2 Popular Trap}. An improvement in an 'After' cell is colored green; a regression is colored red.}
\label{tab:musique_s2_comprehensive}
\resizebox{\textwidth}{!}{
\begin{tabular}{lcc cc cc cc cc cc cc cc}
\toprule
\textbf{Model} & \multicolumn{2}{c}{\textbf{Accuracy (\%)}} & \multicolumn{2}{c}{\textbf{HPSR (\%)}} & \multicolumn{2}{c}{\textbf{PopGap}} & \multicolumn{2}{c}{\textbf{Spearman}} & \multicolumn{2}{c}{\textbf{Confidence}} & \multicolumn{2}{c}{\textbf{Alignment}} & \multicolumn{2}{c}{\textbf{Spearman Conf Pop}} & \multicolumn{2}{c}{\textbf{Spearman Align Pop}} \\
\cmidrule(lr){2-3} \cmidrule(lr){2-3} \cmidrule(lr){4-5} \cmidrule(lr){6-7} \cmidrule(lr){8-9} \cmidrule(lr){10-11} \cmidrule(lr){12-13} \cmidrule(lr){14-15} \cmidrule(lr){16-17}
 & \textbf{Before} & \textbf{After} & \textbf{Before} & \textbf{After} & \textbf{Before} & \textbf{After} & \textbf{Before} & \textbf{After} & \textbf{Before} & \textbf{After} & \textbf{Before} & \textbf{After} & \textbf{Before} & \textbf{After} & \textbf{Before} & \textbf{After} \\
\midrule
\texttt{Qwen2.5-0.5B} & 34.1 & \textcolor{green}{79.5} & 61.6 & \textcolor{green}{16.1} & 0.484 & \textcolor{green}{0.087} & -0.856 & \textcolor{green}{-0.591} & 0.436 & 0.556 & 0.527 & \textcolor{green}{0.574} & -0.005 & \textcolor{red}{-0.400} & 0.370 & \textcolor{green}{-0.309} \\
\texttt{Llama-3.2-1B} & 28.3 & \textcolor{green}{79.5} & 68.4 & \textcolor{green}{17.4} & 0.535 & \textcolor{green}{0.085} & -0.874 & \textcolor{green}{-0.619} & 0.408 & 0.556 & 0.551 & \textcolor{green}{0.592} & -0.002 & \textcolor{red}{-0.518} & 0.430 & \textcolor{green}{-0.374} \\
\texttt{Llama-3.2-3B} & 33.3 & \textcolor{green}{78.7} & 62.6 & \textcolor{green}{17.2} & 0.498 & \textcolor{green}{0.094} & -0.865 & \textcolor{green}{-0.615} & 0.471 & 0.576 & 0.542 & \textcolor{green}{0.601} & -0.116 & \textcolor{red}{-0.488} & 0.140 & \textcolor{red}{-0.375} \\
\bottomrule
\end{tabular}}
\end{table*}

\begin{table*}[h!]
\centering
\caption{Comprehensive debiasing results for \textbf{MuSiQue: S3 Popularity Gradient}. An improvement in an 'After' cell is colored green; a regression is colored red.}
\label{tab:musique_s3_comprehensive}
\resizebox{\textwidth}{!}{
\begin{tabular}{lcc cc cc cc cc cc cc cc}
\toprule
\textbf{Model} & \multicolumn{2}{c}{\textbf{Accuracy (\%)}} & \multicolumn{2}{c}{\textbf{HPSR (\%)}} & \multicolumn{2}{c}{\textbf{PopGap}} & \multicolumn{2}{c}{\textbf{Spearman}} & \multicolumn{2}{c}{\textbf{Confidence}} & \multicolumn{2}{c}{\textbf{Alignment}} & \multicolumn{2}{c}{\textbf{Spearman Conf Pop}} & \multicolumn{2}{c}{\textbf{Spearman Align Pop}} \\
\cmidrule(lr){2-3} \cmidrule(lr){2-3} \cmidrule(lr){4-5} \cmidrule(lr){6-7} \cmidrule(lr){8-9} \cmidrule(lr){10-11} \cmidrule(lr){12-13} \cmidrule(lr){14-15} \cmidrule(lr){16-17}
 & \textbf{Before} & \textbf{After} & \textbf{Before} & \textbf{After} & \textbf{Before} & \textbf{After} & \textbf{Before} & \textbf{After} & \textbf{Before} & \textbf{After} & \textbf{Before} & \textbf{After} & \textbf{Before} & \textbf{After} & \textbf{Before} & \textbf{After} \\
\midrule
\texttt{Qwen2.5-1.5B} & 35.6 & \textcolor{green}{63.2} & 52.2 & \textcolor{green}{19.2} & 0.389 & \textcolor{green}{0.111} & -0.647 & \textcolor{green}{-0.424} & 0.479 & 0.507 & 0.530 & \textcolor{red}{0.528} & -0.037 & \textcolor{red}{-0.098} & 0.137 & \textcolor{green}{-0.028} \\
\texttt{Gemma-2-9b-it} & 32.0 & \textcolor{green}{43.5} & 65.7 & \textcolor{green}{50.1} & 0.496 & \textcolor{green}{0.336} & -0.831 & \textcolor{green}{-0.729} & 0.839 & 0.744 & 0.384 & \textcolor{green}{0.499} & 0.062 & \textcolor{red}{0.064} & -0.712 & \textcolor{green}{-0.630} \\
\texttt{phi-1\_5} & 29.4 & \textcolor{green}{65.3} & 65.7 & \textcolor{green}{24.5} & 0.466 & \textcolor{green}{0.104} & -0.733 & \textcolor{green}{-0.476} & 0.416 & 0.457 & 0.548 & \textcolor{red}{0.521} & -0.061 & \textcolor{red}{-0.318} & 0.418 & \textcolor{green}{0.045} \\
\bottomrule
\end{tabular}}
\end{table*}

\begin{table*}[h!]
\centering
\caption{Comprehensive debiasing results for \textbf{MuSiQue: S4 Direct Contest}. An improvement in an 'After' cell is colored green; a regression is colored red.}
\label{tab:musique_s4_comprehensive}
\resizebox{\textwidth}{!}{
\begin{tabular}{lcc cc cc cc cc cc cc cc}
\toprule
\textbf{Model} & \multicolumn{2}{c}{\textbf{Accuracy (\%)}} & \multicolumn{2}{c}{\textbf{HPSR (\%)}} & \multicolumn{2}{c}{\textbf{PopGap}} & \multicolumn{2}{c}{\textbf{Spearman}} & \multicolumn{2}{c}{\textbf{Confidence}} & \multicolumn{2}{c}{\textbf{Alignment}} & \multicolumn{2}{c}{\textbf{Spearman Conf Pop}} & \multicolumn{2}{c}{\textbf{Spearman Align Pop}} \\
\cmidrule(lr){2-3} \cmidrule(lr){2-3} \cmidrule(lr){4-5} \cmidrule(lr){6-7} \cmidrule(lr){8-9} \cmidrule(lr){10-11} \cmidrule(lr){12-13} \cmidrule(lr){14-15} \cmidrule(lr){16-17}
 & \textbf{Before} & \textbf{After} & \textbf{Before} & \textbf{After} & \textbf{Before} & \textbf{After} & \textbf{Before} & \textbf{After} & \textbf{Before} & \textbf{After} & \textbf{Before} & \textbf{After} & \textbf{Before} & \textbf{After} & \textbf{Before} & \textbf{After} \\
\midrule
\texttt{Falcon-7b-instruct} & 25.1 & \textcolor{green}{50.3} & 57.0 & \textcolor{green}{23.7} & 0.375 & \textcolor{green}{0.061} & -0.528 & \textcolor{green}{-0.353} & 0.399 & 0.454 & 0.552 & \textcolor{red}{0.549} & 0.005 & \textcolor{red}{-0.313} & 0.375 & \textcolor{green}{0.016} \\
\texttt{Qwen2.5-1.5B} & 30.4 & \textcolor{green}{50.7} & 54.5 & \textcolor{green}{27.6} & 0.340 & \textcolor{green}{0.087} & -0.508 & \textcolor{green}{-0.323} & 0.472 & 0.495 & 0.530 & \textcolor{green}{0.549} & 0.022 & \textcolor{red}{-0.097} & 0.075 & \textcolor{red}{-0.100} \\
\bottomrule
\end{tabular}}
\end{table*}

\begin{table*}[h!]
\centering
\caption{Comprehensive debiasing results for \textbf{MuSiQue: S5 Reverse Control}. An improvement in an 'After' cell is colored green; a regression is colored red.}
\label{tab:musique_s5_comprehensive}
\resizebox{\textwidth}{!}{
\begin{tabular}{lcc cc cc cc cc cc cc cc}
\toprule
\textbf{Model} & \multicolumn{2}{c}{\textbf{Accuracy (\%)}} & \multicolumn{2}{c}{\textbf{HPSR (\%)}} & \multicolumn{2}{c}{\textbf{PopGap}} & \multicolumn{2}{c}{\textbf{Spearman}} & \multicolumn{2}{c}{\textbf{Confidence}} & \multicolumn{2}{c}{\textbf{Alignment}} & \multicolumn{2}{c}{\textbf{Spearman Conf Pop}} & \multicolumn{2}{c}{\textbf{Spearman Align Pop}} \\
\cmidrule(lr){2-3} \cmidrule(lr){2-3} \cmidrule(lr){4-5} \cmidrule(lr){6-7} \cmidrule(lr){8-9} \cmidrule(lr){10-11} \cmidrule(lr){12-13} \cmidrule(lr){14-15} \cmidrule(lr){16-17}
 & \textbf{Before} & \textbf{After} & \textbf{Before} & \textbf{After} & \textbf{Before} & \textbf{After} & \textbf{Before} & \textbf{After} & \textbf{Before} & \textbf{After} & \textbf{Before} & \textbf{After} & \textbf{Before} & \textbf{After} & \textbf{Before} & \textbf{After} \\
\midrule
\texttt{Qwen2.5-0.5B} & 30.9 & \textcolor{green}{34.8} & 51.7 & \textcolor{green}{41.1} & 0.016 & \textcolor{red}{-0.082} & 0.193 & \textcolor{red}{0.294} & 0.397 & 0.402 & 0.556 & \textcolor{green}{0.556} & 0.209 & \textcolor{green}{0.201} & -0.217 & \textcolor{red}{-0.292} \\
\texttt{Llama-3.2-1B} & 33.3 & \textcolor{green}{38.9} & 58.5 & \textcolor{green}{45.5} & 0.063 & \textcolor{green}{-0.058} & 0.189 & \textcolor{red}{0.315} & 0.402 & 0.408 & 0.553 & \textcolor{red}{0.551} & 0.196 & \textcolor{red}{0.197} & -0.134 & \textcolor{red}{-0.229} \\
\texttt{Qwen2.5-3B} & 41.6 & \textcolor{green}{45.0} & 56.8 & \textcolor{green}{48.5} & 0.039 & \textcolor{green}{-0.037} & 0.251 & \textcolor{red}{0.356} & 0.444 & 0.443 & 0.552 & \textcolor{red}{0.548} & 0.219 & \textcolor{red}{0.229} & -0.130 & \textcolor{red}{-0.195} \\
\bottomrule
\end{tabular}}
\end{table*}

\begin{table*}[h!]
\centering
\caption{Comprehensive debiasing results for \textbf{MuSiQue: S6 None of the Above}. An improvement in an 'After' cell is colored green; a regression is colored red.}
\label{tab:musique_s6_comprehensive}
\resizebox{\textwidth}{!}{
\begin{tabular}{lcc cc cc cc cc cc cc cc}
\toprule
\textbf{Model} & \multicolumn{2}{c}{\textbf{Accuracy (\%)}} & \multicolumn{2}{c}{\textbf{HPSR (\%)}} & \multicolumn{2}{c}{\textbf{PopGap}} & \multicolumn{2}{c}{\textbf{Spearman}} & \multicolumn{2}{c}{\textbf{Confidence}} & \multicolumn{2}{c}{\textbf{Alignment}} & \multicolumn{2}{c}{\textbf{Spearman Conf Pop}} & \multicolumn{2}{c}{\textbf{Spearman Align Pop}} \\
\cmidrule(lr){2-3} \cmidrule(lr){2-3} \cmidrule(lr){4-5} \cmidrule(lr){6-7} \cmidrule(lr){8-9} \cmidrule(lr){10-11} \cmidrule(lr){12-13} \cmidrule(lr){14-15} \cmidrule(lr){16-17}
 & \textbf{Before} & \textbf{After} & \textbf{Before} & \textbf{After} & \textbf{Before} & \textbf{After} & \textbf{Before} & \textbf{After} & \textbf{Before} & \textbf{After} & \textbf{Before} & \textbf{After} & \textbf{Before} & \textbf{After} & \textbf{Before} & \textbf{After} \\
\midrule
\texttt{DeepSeek-V2-Lite} & 68.6 & \textcolor{green}{80.7} & 22.2 & \textcolor{green}{7.8} & 0.148 & \textcolor{green}{0.005} & -0.955 & \textcolor{green}{-0.942} & 0.341 & 0.348 & 0.452 & \textcolor{red}{0.415} & -0.283 & \textcolor{green}{-0.205} & 0.762 & \textcolor{green}{0.640} \\
\texttt{Meta-Llama-3-8B} & 23.2 & \textcolor{green}{41.0} & 51.3 & \textcolor{green}{27.3} & 0.311 & \textcolor{green}{0.034} & -0.730 & \textcolor{red}{-0.850} & 0.679 & 0.581 & 0.372 & \textcolor{green}{0.446} & 0.218 & \textcolor{green}{0.180} & -0.448 & \textcolor{green}{-0.255} \\
\texttt{Qwen2.5-3B} & 6.2 & \textcolor{green}{20.4} & 65.8 & \textcolor{green}{35.5} & 0.444 & \textcolor{green}{0.034} & -0.422 & \textcolor{red}{-0.671} & 0.439 & 0.403 & 0.545 & \textcolor{green}{0.555} & 0.221 & \textcolor{green}{0.143} & -0.029 & \textcolor{red}{0.290} \\
\bottomrule
\end{tabular}}
\end{table*}

\subsection{Natural Questions Results}
\label{app:nq_results}

NQ's simpler factoid questions yield higher baseline accuracy (45–65\%).  Under S2,
$\rho=-0.871$; PopDebias improves accuracy by $+30.2$~pp on average.

\begin{table*}[h!]
\centering
\caption{Comprehensive debiasing results for \textbf{NQ: S1 Baseline Control}. An improvement in an 'After' cell is colored green; a regression is colored red.}
\label{tab:nq_s1_comprehensive}
\resizebox{\textwidth}{!}{
\begin{tabular}{lcc cc cc cc cc cc cc cc}
\toprule
\textbf{Model} & \multicolumn{2}{c}{\textbf{Accuracy (\%)}} & \multicolumn{2}{c}{\textbf{HPSR (\%)}} & \multicolumn{2}{c}{\textbf{PopGap}} & \multicolumn{2}{c}{\textbf{Spearman}} & \multicolumn{2}{c}{\textbf{Confidence}} & \multicolumn{2}{c}{\textbf{Alignment}} & \multicolumn{2}{c}{\textbf{Spearman Conf Pop}} & \multicolumn{2}{c}{\textbf{Spearman Align Pop}} \\
\cmidrule(lr){2-3} \cmidrule(lr){2-3} \cmidrule(lr){4-5} \cmidrule(lr){6-7} \cmidrule(lr){8-9} \cmidrule(lr){10-11} \cmidrule(lr){12-13} \cmidrule(lr){14-15} \cmidrule(lr){16-17}
 & \textbf{Before} & \textbf{After} & \textbf{Before} & \textbf{After} & \textbf{Before} & \textbf{After} & \textbf{Before} & \textbf{After} & \textbf{Before} & \textbf{After} & \textbf{Before} & \textbf{After} & \textbf{Before} & \textbf{After} & \textbf{Before} & \textbf{After} \\
\midrule
\texttt{Llama-2-13b-hf} & 48.5 & \textcolor{green}{57.0} & 43.3 & \textcolor{green}{26.3} & 0.155 & \textcolor{green}{0.015} & -0.322 & \textcolor{green}{-0.098} & 0.497 & 0.493 & 0.555 & \textcolor{red}{0.546} & -0.093 & \textcolor{green}{-0.008} & 0.069 & \textcolor{red}{0.081} \\
\texttt{Gemma-2-9b-it} & 32.6 & \textcolor{green}{34.5} & 50.7 & \textcolor{green}{46.5} & 0.219 & \textcolor{green}{0.177} & -0.326 & \textcolor{green}{-0.283} & 0.835 & 0.786 & 0.393 & \textcolor{green}{0.435} & 0.038 & \textcolor{red}{-0.123} & -0.277 & \textcolor{green}{-0.154} \\
\texttt{Gemma-3-1b-it} & 23.6 & \textcolor{green}{30.5} & 53.5 & \textcolor{green}{42.3} & 0.238 & \textcolor{green}{0.131} & -0.336 & \textcolor{green}{-0.278} & 0.666 & 0.624 & 0.405 & \textcolor{green}{0.441} & 0.060 & \textcolor{green}{0.026} & -0.223 & \textcolor{green}{-0.125} \\
\bottomrule
\end{tabular}}
\end{table*}

\begin{table*}[h!]
\centering
\caption{Comprehensive debiasing results for \textbf{NQ: S2 Popular Trap}. An improvement in an 'After' cell is colored green; a regression is colored red.}
\label{tab:nq_s2_comprehensive}
\resizebox{\textwidth}{!}{
\begin{tabular}{lcc cc cc cc cc cc cc cc}
\toprule
\textbf{Model} & \multicolumn{2}{c}{\textbf{Accuracy (\%)}} & \multicolumn{2}{c}{\textbf{HPSR (\%)}} & \multicolumn{2}{c}{\textbf{PopGap}} & \multicolumn{2}{c}{\textbf{Spearman}} & \multicolumn{2}{c}{\textbf{Confidence}} & \multicolumn{2}{c}{\textbf{Alignment}} & \multicolumn{2}{c}{\textbf{Spearman Conf Pop}} & \multicolumn{2}{c}{\textbf{Spearman Align Pop}} \\
\cmidrule(lr){2-3} \cmidrule(lr){2-3} \cmidrule(lr){4-5} \cmidrule(lr){6-7} \cmidrule(lr){8-9} \cmidrule(lr){10-11} \cmidrule(lr){12-13} \cmidrule(lr){14-15} \cmidrule(lr){16-17}
 & \textbf{Before} & \textbf{After} & \textbf{Before} & \textbf{After} & \textbf{Before} & \textbf{After} & \textbf{Before} & \textbf{After} & \textbf{Before} & \textbf{After} & \textbf{Before} & \textbf{After} & \textbf{Before} & \textbf{After} & \textbf{Before} & \textbf{After} \\
\midrule
\texttt{phi-2} & 48.8 & \textcolor{green}{80.7} & 49.5 & \textcolor{green}{17.5} & 0.367 & \textcolor{green}{0.087} & -0.895 & \textcolor{green}{-0.650} & 0.543 & 0.621 & 0.566 & \textcolor{green}{0.624} & -0.270 & \textcolor{red}{-0.396} & -0.150 & \textcolor{red}{-0.382} \\
\texttt{phi-4} & 63.3 & \textcolor{green}{76.7} & 35.4 & \textcolor{green}{22.1} & 0.263 & \textcolor{green}{0.141} & -0.833 & \textcolor{green}{-0.713} & 0.832 & 0.827 & 0.656 & \textcolor{green}{0.719} & -0.293 & \textcolor{green}{-0.106} & -0.649 & \textcolor{green}{-0.394} \\
\texttt{Qwen2.5-3B} & 51.2 & \textcolor{green}{78.6} & 47.0 & \textcolor{green}{19.6} & 0.345 & \textcolor{green}{0.100} & -0.886 & \textcolor{green}{-0.675} & 0.600 & 0.659 & 0.579 & \textcolor{green}{0.644} & -0.256 & \textcolor{red}{-0.295} & -0.299 & \textcolor{red}{-0.359} \\
\bottomrule
\end{tabular}}
\end{table*}

\begin{table*}[h!]
\centering
\caption{Comprehensive debiasing results for \textbf{NQ: S3 Popularity Gradient}. An improvement in an 'After' cell is colored green; a regression is colored red.}
\label{tab:nq_s3_comprehensive}
\resizebox{\textwidth}{!}{
\begin{tabular}{lcc cc cc cc cc cc cc cc}
\toprule
\textbf{Model} & \multicolumn{2}{c}{\textbf{Accuracy (\%)}} & \multicolumn{2}{c}{\textbf{HPSR (\%)}} & \multicolumn{2}{c}{\textbf{PopGap}} & \multicolumn{2}{c}{\textbf{Spearman}} & \multicolumn{2}{c}{\textbf{Confidence}} & \multicolumn{2}{c}{\textbf{Alignment}} & \multicolumn{2}{c}{\textbf{Spearman Conf Pop}} & \multicolumn{2}{c}{\textbf{Spearman Align Pop}} \\
\cmidrule(lr){2-3} \cmidrule(lr){2-3} \cmidrule(lr){4-5} \cmidrule(lr){6-7} \cmidrule(lr){8-9} \cmidrule(lr){10-11} \cmidrule(lr){12-13} \cmidrule(lr){14-15} \cmidrule(lr){16-17}
 & \textbf{Before} & \textbf{After} & \textbf{Before} & \textbf{After} & \textbf{Before} & \textbf{After} & \textbf{Before} & \textbf{After} & \textbf{Before} & \textbf{After} & \textbf{Before} & \textbf{After} & \textbf{Before} & \textbf{After} & \textbf{Before} & \textbf{After} \\
\midrule
\texttt{Gemma-2-9b-it} & 34.0 & \textcolor{green}{46.0} & 62.4 & \textcolor{green}{47.3} & 0.446 & \textcolor{green}{0.308} & -0.844 & \textcolor{green}{-0.784} & 0.850 & 0.777 & 0.399 & \textcolor{green}{0.501} & 0.016 & \textcolor{red}{0.046} & -0.720 & \textcolor{green}{-0.651} \\
\texttt{Meta-Llama-3-8B} & 60.8 & \textcolor{green}{67.3} & 31.2 & \textcolor{green}{23.7} & 0.227 & \textcolor{green}{0.160} & -0.677 & \textcolor{green}{-0.596} & 0.863 & 0.833 & 0.632 & \textcolor{green}{0.673} & -0.230 & \textcolor{green}{-0.101} & -0.560 & \textcolor{green}{-0.391} \\
\texttt{Llama-3.2-1B} & 31.8 & \textcolor{green}{61.9} & 55.5 & \textcolor{green}{21.6} & 0.406 & \textcolor{green}{0.120} & -0.712 & \textcolor{green}{-0.555} & 0.438 & 0.483 & 0.553 & \textcolor{red}{0.535} & -0.179 & \textcolor{red}{-0.299} & 0.318 & \textcolor{green}{0.040} \\
\bottomrule
\end{tabular}}
\end{table*}

\begin{table*}[h!]
\centering
\caption{Comprehensive debiasing results for \textbf{NQ: S4 Direct Contest}. An improvement in an 'After' cell is colored green; a regression is colored red.}
\label{tab:nq_s4_comprehensive}
\resizebox{\textwidth}{!}{
\begin{tabular}{lcc cc cc cc cc cc cc cc}
\toprule
\textbf{Model} & \multicolumn{2}{c}{\textbf{Accuracy (\%)}} & \multicolumn{2}{c}{\textbf{HPSR (\%)}} & \multicolumn{2}{c}{\textbf{PopGap}} & \multicolumn{2}{c}{\textbf{Spearman}} & \multicolumn{2}{c}{\textbf{Confidence}} & \multicolumn{2}{c}{\textbf{Alignment}} & \multicolumn{2}{c}{\textbf{Spearman Conf Pop}} & \multicolumn{2}{c}{\textbf{Spearman Align Pop}} \\
\cmidrule(lr){2-3} \cmidrule(lr){2-3} \cmidrule(lr){4-5} \cmidrule(lr){6-7} \cmidrule(lr){8-9} \cmidrule(lr){10-11} \cmidrule(lr){12-13} \cmidrule(lr){14-15} \cmidrule(lr){16-17}
 & \textbf{Before} & \textbf{After} & \textbf{Before} & \textbf{After} & \textbf{Before} & \textbf{After} & \textbf{Before} & \textbf{After} & \textbf{Before} & \textbf{After} & \textbf{Before} & \textbf{After} & \textbf{Before} & \textbf{After} & \textbf{Before} & \textbf{After} \\
\midrule
\texttt{Qwen2.5-1.5B} & 42.1 & \textcolor{green}{58.8} & 51.1 & \textcolor{green}{30.7} & 0.303 & \textcolor{green}{0.111} & -0.642 & \textcolor{green}{-0.492} & 0.588 & 0.587 & 0.548 & \textcolor{green}{0.574} & -0.090 & \textcolor{green}{-0.046} & -0.258 & \textcolor{green}{-0.219} \\
\texttt{phi-2} & 40.0 & \textcolor{green}{58.1} & 47.0 & \textcolor{green}{25.0} & 0.278 & \textcolor{green}{0.078} & -0.572 & \textcolor{green}{-0.424} & 0.513 & 0.523 & 0.543 & \textcolor{green}{0.560} & -0.049 & \textcolor{red}{-0.121} & -0.052 & \textcolor{red}{-0.080} \\
\texttt{Llama-3.2-1B} & 30.5 & \textcolor{green}{57.5} & 60.7 & \textcolor{green}{27.0} & 0.382 & \textcolor{green}{0.081} & -0.602 & \textcolor{green}{-0.415} & 0.438 & 0.471 & 0.550 & \textcolor{red}{0.550} & -0.036 & \textcolor{red}{-0.230} & 0.157 & \textcolor{green}{-0.086} \\
\bottomrule
\end{tabular}}
\end{table*}

\begin{table*}[h!]
\centering
\caption{Comprehensive debiasing results for \textbf{NQ: S5 Reverse Control}. An improvement in an 'After' cell is colored green; a regression is colored red.}
\label{tab:nq_s5_comprehensive}
\resizebox{\textwidth}{!}{
\begin{tabular}{lcc cc cc cc cc cc cc cc}
\toprule
\textbf{Model} & \multicolumn{2}{c}{\textbf{Accuracy (\%)}} & \multicolumn{2}{c}{\textbf{HPSR (\%)}} & \multicolumn{2}{c}{\textbf{PopGap}} & \multicolumn{2}{c}{\textbf{Spearman}} & \multicolumn{2}{c}{\textbf{Confidence}} & \multicolumn{2}{c}{\textbf{Alignment}} & \multicolumn{2}{c}{\textbf{Spearman Conf Pop}} & \multicolumn{2}{c}{\textbf{Spearman Align Pop}} \\
\cmidrule(lr){2-3} \cmidrule(lr){2-3} \cmidrule(lr){4-5} \cmidrule(lr){6-7} \cmidrule(lr){8-9} \cmidrule(lr){10-11} \cmidrule(lr){12-13} \cmidrule(lr){14-15} \cmidrule(lr){16-17}
 & \textbf{Before} & \textbf{After} & \textbf{Before} & \textbf{After} & \textbf{Before} & \textbf{After} & \textbf{Before} & \textbf{After} & \textbf{Before} & \textbf{After} & \textbf{Before} & \textbf{After} & \textbf{Before} & \textbf{After} & \textbf{Before} & \textbf{After} \\
\midrule
\texttt{Llama-2-7b-hf} & 31.1 & \textcolor{green}{35.1} & 53.2 & \textcolor{green}{41.0} & 0.004 & \textcolor{red}{-0.106} & 0.189 & \textcolor{red}{0.244} & 0.409 & 0.411 & 0.550 & \textcolor{red}{0.545} & 0.072 & \textcolor{green}{0.064} & -0.129 & \textcolor{red}{-0.228} \\
\texttt{Qwen2.5-32B} & 58.7 & \textcolor{green}{60.8} & 60.8 & \textcolor{green}{57.2} & 0.020 & \textcolor{green}{-0.015} & 0.150 & \textcolor{red}{0.214} & 0.674 & 0.657 & 0.646 & \textcolor{red}{0.638} & 0.155 & \textcolor{green}{0.137} & 0.112 & \textcolor{red}{0.128} \\
\texttt{DeepSeek-V2-Lite} & 45.5 & \textcolor{green}{48.0} & 60.8 & \textcolor{green}{46.4} & 0.026 & \textcolor{red}{-0.106} & 0.150 & \textcolor{red}{0.283} & 0.353 & 0.361 & 0.540 & \textcolor{red}{0.539} & 0.113 & \textcolor{green}{0.110} & -0.116 & \textcolor{red}{-0.250} \\
\texttt{Llama-3.2-3B} & 49.9 & \textcolor{green}{52.1} & 61.2 & \textcolor{green}{52.5} & 0.035 & \textcolor{red}{-0.047} & 0.175 & \textcolor{red}{0.283} & 0.533 & 0.520 & 0.578 & \textcolor{red}{0.570} & 0.124 & \textcolor{red}{0.152} & 0.026 & \textcolor{green}{0.017} \\
\bottomrule
\end{tabular}}
\end{table*}

\begin{table*}[h!]
\centering
\caption{Comprehensive debiasing results for \textbf{NQ: S6 None of the Above}. An improvement in an 'After' cell is colored green; a regression is colored red.}
\label{tab:nq_s6_comprehensive}
\resizebox{\textwidth}{!}{
\begin{tabular}{lcc cc cc cc cc cc cc cc}
\toprule
\textbf{Model} & \multicolumn{2}{c}{\textbf{Accuracy (\%)}} & \multicolumn{2}{c}{\textbf{HPSR (\%)}} & \multicolumn{2}{c}{\textbf{PopGap}} & \multicolumn{2}{c}{\textbf{Spearman}} & \multicolumn{2}{c}{\textbf{Confidence}} & \multicolumn{2}{c}{\textbf{Alignment}} & \multicolumn{2}{c}{\textbf{Spearman Conf Pop}} & \multicolumn{2}{c}{\textbf{Spearman Align Pop}} \\
\cmidrule(lr){2-3} \cmidrule(lr){2-3} \cmidrule(lr){4-5} \cmidrule(lr){6-7} \cmidrule(lr){8-9} \cmidrule(lr){10-11} \cmidrule(lr){12-13} \cmidrule(lr){14-15} \cmidrule(lr){16-17}
 & \textbf{Before} & \textbf{After} & \textbf{Before} & \textbf{After} & \textbf{Before} & \textbf{After} & \textbf{Before} & \textbf{After} & \textbf{Before} & \textbf{After} & \textbf{Before} & \textbf{After} & \textbf{Before} & \textbf{After} & \textbf{Before} & \textbf{After} \\
\midrule
\texttt{Gemma-2-2b-it} & 67.2 & \textcolor{green}{78.1} & 23.4 & \textcolor{green}{11.0} & 0.158 & \textcolor{green}{0.017} & -0.951 & \textcolor{green}{-0.950} & 0.529 & 0.513 & 0.514 & \textcolor{green}{0.521} & -0.064 & \textcolor{red}{-0.143} & -0.124 & \textcolor{green}{0.038} \\
\texttt{phi-2} & 3.0 & \textcolor{green}{11.7} & 63.0 & \textcolor{green}{40.9} & 0.355 & \textcolor{green}{0.070} & -0.284 & \textcolor{red}{-0.524} & 0.487 & 0.454 & 0.505 & \textcolor{green}{0.532} & 0.142 & \textcolor{red}{0.173} & -0.040 & \textcolor{green}{0.036} \\
\texttt{Meta-Llama-3-8B} & 10.8 & \textcolor{green}{26.5} & 59.9 & \textcolor{green}{36.1} & 0.349 & \textcolor{green}{0.061} & -0.520 & \textcolor{red}{-0.724} & 0.770 & 0.646 & 0.262 & \textcolor{green}{0.372} & 0.190 & \textcolor{red}{0.213} & -0.317 & \textcolor{green}{-0.282} \\
\texttt{Llama-3.2-3B} & 0.0 & \textcolor{green}{2.3} & 69.4 & \textcolor{green}{45.5} & 0.461 & \textcolor{green}{0.082} & 0.000 & \textcolor{red}{-0.243} & 0.540 & 0.479 & 0.460 & \textcolor{green}{0.529} & 0.183 & \textcolor{green}{0.135} & -0.183 & \textcolor{red}{-0.186} \\
\bottomrule
\end{tabular}}
\end{table*}

\subsection{WebQuestions Results}
\label{app:webq_results}

WebQuestions shows consistent patterns with the other datasets: $\rho=-0.875$ under S2,
$+27.9$~pp accuracy gain with PopDebias, and the same S5 reversal and S6 abstention
failures.

\begin{table*}[h!]
\centering
\caption{Comprehensive debiasing results for \textbf{WebQ: S1 Baseline Control}. An improvement in an 'After' cell is colored green; a regression is colored red.}
\label{tab:webq_s1_comprehensive}
\resizebox{\textwidth}{!}{
\begin{tabular}{lcc cc cc cc cc cc cc cc}
\toprule
\textbf{Model} & \multicolumn{2}{c}{\textbf{Accuracy (\%)}} & \multicolumn{2}{c}{\textbf{HPSR (\%)}} & \multicolumn{2}{c}{\textbf{PopGap}} & \multicolumn{2}{c}{\textbf{Spearman}} & \multicolumn{2}{c}{\textbf{Confidence}} & \multicolumn{2}{c}{\textbf{Alignment}} & \multicolumn{2}{c}{\textbf{Spearman Conf Pop}} & \multicolumn{2}{c}{\textbf{Spearman Align Pop}} \\
\cmidrule(lr){2-3} \cmidrule(lr){2-3} \cmidrule(lr){4-5} \cmidrule(lr){6-7} \cmidrule(lr){8-9} \cmidrule(lr){10-11} \cmidrule(lr){12-13} \cmidrule(lr){14-15} \cmidrule(lr){16-17}
 & \textbf{Before} & \textbf{After} & \textbf{Before} & \textbf{After} & \textbf{Before} & \textbf{After} & \textbf{Before} & \textbf{After} & \textbf{Before} & \textbf{After} & \textbf{Before} & \textbf{After} & \textbf{Before} & \textbf{After} & \textbf{Before} & \textbf{After} \\
\midrule
\texttt{Mistral-7B-v0.3} & 56.7 & \textcolor{green}{58.5} & 45.2 & \textcolor{green}{40.6} & 0.129 & \textcolor{green}{0.081} & -0.305 & \textcolor{green}{-0.196} & 0.873 & 0.845 & 0.599 & \textcolor{green}{0.614} & -0.109 & \textcolor{red}{-0.150} & -0.266 & \textcolor{green}{-0.195} \\
\texttt{Meta-Llama-3-8B} & 54.6 & \textcolor{green}{56.6} & 45.6 & \textcolor{green}{40.2} & 0.136 & \textcolor{green}{0.083} & -0.295 & \textcolor{green}{-0.174} & 0.832 & 0.799 & 0.616 & \textcolor{green}{0.636} & -0.098 & \textcolor{green}{-0.095} & -0.283 & \textcolor{green}{-0.173} \\
\texttt{phi-1\_5} & 35.9 & \textcolor{green}{48.7} & 52.1 & \textcolor{green}{27.5} & 0.207 & \textcolor{green}{-0.015} & -0.321 & \textcolor{green}{-0.030} & 0.432 & 0.446 & 0.555 & \textcolor{red}{0.539} & -0.023 & \textcolor{red}{-0.033} & 0.089 & \textcolor{green}{-0.047} \\
\bottomrule
\end{tabular}}
\end{table*}

\begin{table*}[h!]
\centering
\caption{Comprehensive debiasing results for \textbf{WebQ: S2 Popular Trap}. An improvement in an 'After' cell is colored green; a regression is colored red.}
\label{tab:webq_s2_comprehensive}
\resizebox{\textwidth}{!}{
\begin{tabular}{lcc cc cc cc cc cc cc cc}
\toprule
\textbf{Model} & \multicolumn{2}{c}{\textbf{Accuracy (\%)}} & \multicolumn{2}{c}{\textbf{HPSR (\%)}} & \multicolumn{2}{c}{\textbf{PopGap}} & \multicolumn{2}{c}{\textbf{Spearman}} & \multicolumn{2}{c}{\textbf{Confidence}} & \multicolumn{2}{c}{\textbf{Alignment}} & \multicolumn{2}{c}{\textbf{Spearman Conf Pop}} & \multicolumn{2}{c}{\textbf{Spearman Align Pop}} \\
\cmidrule(lr){2-3} \cmidrule(lr){2-3} \cmidrule(lr){4-5} \cmidrule(lr){6-7} \cmidrule(lr){8-9} \cmidrule(lr){10-11} \cmidrule(lr){12-13} \cmidrule(lr){14-15} \cmidrule(lr){16-17}
 & \textbf{Before} & \textbf{After} & \textbf{Before} & \textbf{After} & \textbf{Before} & \textbf{After} & \textbf{Before} & \textbf{After} & \textbf{Before} & \textbf{After} & \textbf{Before} & \textbf{After} & \textbf{Before} & \textbf{After} & \textbf{Before} & \textbf{After} \\
\midrule
\texttt{Qwen2.5-32B} & 53.7 & \textcolor{green}{73.1} & 43.7 & \textcolor{green}{24.6} & 0.322 & \textcolor{green}{0.147} & -0.884 & \textcolor{green}{-0.752} & 0.741 & 0.743 & 0.593 & \textcolor{green}{0.672} & -0.316 & \textcolor{green}{-0.255} & -0.655 & \textcolor{green}{-0.482} \\
\texttt{Falcon-7b-instruct} & 28.6 & \textcolor{green}{74.5} & 68.5 & \textcolor{green}{21.9} & 0.502 & \textcolor{green}{0.100} & -0.910 & \textcolor{green}{-0.712} & 0.393 & 0.525 & 0.548 & \textcolor{green}{0.583} & -0.051 & \textcolor{red}{-0.596} & 0.729 & \textcolor{green}{-0.192} \\
\texttt{Gemma-2-2b-it} & 32.9 & \textcolor{green}{63.2} & 64.1 & \textcolor{green}{33.6} & 0.472 & \textcolor{green}{0.194} & -0.910 & \textcolor{green}{-0.817} & 0.633 & 0.659 & 0.497 & \textcolor{green}{0.595} & -0.215 & \textcolor{red}{-0.387} & -0.548 & \textcolor{red}{-0.616} \\
\bottomrule
\end{tabular}}
\end{table*}

\begin{table*}[h!]
\centering
\caption{Comprehensive debiasing results for \textbf{WebQ: S3 Popularity Gradient}. An improvement in an 'After' cell is colored green; a regression is colored red.}
\label{tab:webq_s3_comprehensive}
\resizebox{\textwidth}{!}{
\begin{tabular}{lcc cc cc cc cc cc cc cc}
\toprule
\textbf{Model} & \multicolumn{2}{c}{\textbf{Accuracy (\%)}} & \multicolumn{2}{c}{\textbf{HPSR (\%)}} & \multicolumn{2}{c}{\textbf{PopGap}} & \multicolumn{2}{c}{\textbf{Spearman}} & \multicolumn{2}{c}{\textbf{Confidence}} & \multicolumn{2}{c}{\textbf{Alignment}} & \multicolumn{2}{c}{\textbf{Spearman Conf Pop}} & \multicolumn{2}{c}{\textbf{Spearman Align Pop}} \\
\cmidrule(lr){2-3} \cmidrule(lr){2-3} \cmidrule(lr){4-5} \cmidrule(lr){6-7} \cmidrule(lr){8-9} \cmidrule(lr){10-11} \cmidrule(lr){12-13} \cmidrule(lr){14-15} \cmidrule(lr){16-17}
 & \textbf{Before} & \textbf{After} & \textbf{Before} & \textbf{After} & \textbf{Before} & \textbf{After} & \textbf{Before} & \textbf{After} & \textbf{Before} & \textbf{After} & \textbf{Before} & \textbf{After} & \textbf{Before} & \textbf{After} & \textbf{Before} & \textbf{After} \\
\midrule
\texttt{Llama-3.2-1B} & 42.0 & \textcolor{green}{70.2} & 50.7 & \textcolor{green}{20.2} & 0.360 & \textcolor{green}{0.096} & -0.776 & \textcolor{green}{-0.540} & 0.497 & 0.517 & 0.560 & \textcolor{red}{0.551} & -0.269 & \textcolor{red}{-0.294} & -0.054 & \textcolor{green}{-0.050} \\
\texttt{DeepSeek-V2-Lite} & 46.4 & \textcolor{green}{71.8} & 40.5 & \textcolor{green}{14.4} & 0.302 & \textcolor{green}{0.071} & -0.691 & \textcolor{green}{-0.425} & 0.437 & 0.470 & 0.542 & \textcolor{red}{0.528} & -0.119 & \textcolor{red}{-0.251} & 0.228 & \textcolor{green}{0.007} \\
\texttt{deepseek-llm-7b-base} & 44.9 & \textcolor{green}{75.5} & 47.3 & \textcolor{green}{16.0} & 0.337 & \textcolor{green}{0.068} & -0.757 & \textcolor{green}{-0.457} & 0.412 & 0.449 & 0.538 & \textcolor{red}{0.503} & -0.170 & \textcolor{red}{-0.328} & 0.411 & \textcolor{green}{0.012} \\
\bottomrule
\end{tabular}}
\end{table*}

\begin{table*}[h!]
\centering
\caption{Comprehensive debiasing results for \textbf{WebQ: S4 Direct Contest}. An improvement in an 'After' cell is colored green; a regression is colored red.}
\label{tab:webq_s4_comprehensive}
\resizebox{\textwidth}{!}{
\begin{tabular}{lcc cc cc cc cc cc cc cc}
\toprule
\textbf{Model} & \multicolumn{2}{c}{\textbf{Accuracy (\%)}} & \multicolumn{2}{c}{\textbf{HPSR (\%)}} & \multicolumn{2}{c}{\textbf{PopGap}} & \multicolumn{2}{c}{\textbf{Spearman}} & \multicolumn{2}{c}{\textbf{Confidence}} & \multicolumn{2}{c}{\textbf{Alignment}} & \multicolumn{2}{c}{\textbf{Spearman Conf Pop}} & \multicolumn{2}{c}{\textbf{Spearman Align Pop}} \\
\cmidrule(lr){2-3} \cmidrule(lr){2-3} \cmidrule(lr){4-5} \cmidrule(lr){6-7} \cmidrule(lr){8-9} \cmidrule(lr){10-11} \cmidrule(lr){12-13} \cmidrule(lr){14-15} \cmidrule(lr){16-17}
 & \textbf{Before} & \textbf{After} & \textbf{Before} & \textbf{After} & \textbf{Before} & \textbf{After} & \textbf{Before} & \textbf{After} & \textbf{Before} & \textbf{After} & \textbf{Before} & \textbf{After} & \textbf{Before} & \textbf{After} & \textbf{Before} & \textbf{After} \\
\midrule
\texttt{Qwen2.5-1.5B} & 41.5 & \textcolor{green}{57.0} & 51.1 & \textcolor{green}{33.5} & 0.291 & \textcolor{green}{0.120} & -0.641 & \textcolor{green}{-0.515} & 0.661 & 0.638 & 0.539 & \textcolor{green}{0.584} & -0.105 & \textcolor{green}{-0.101} & -0.490 & \textcolor{green}{-0.343} \\
\texttt{Qwen2.5-3B} & 43.4 & \textcolor{green}{58.7} & 51.1 & \textcolor{green}{33.6} & 0.291 & \textcolor{green}{0.117} & -0.630 & \textcolor{green}{-0.462} & 0.666 & 0.654 & 0.553 & \textcolor{green}{0.598} & -0.127 & \textcolor{green}{-0.080} & -0.483 & \textcolor{green}{-0.391} \\
\texttt{Qwen2.5-7B} & 46.9 & \textcolor{green}{56.1} & 50.3 & \textcolor{green}{38.0} & 0.278 & \textcolor{green}{0.163} & -0.645 & \textcolor{green}{-0.544} & 0.743 & 0.707 & 0.555 & \textcolor{green}{0.600} & -0.162 & \textcolor{green}{-0.112} & -0.579 & \textcolor{green}{-0.446} \\
\bottomrule
\end{tabular}}
\end{table*}

\begin{table*}[h!]
\centering
\caption{Comprehensive debiasing results for \textbf{WebQ: S5 Reverse Control}. An improvement in an 'After' cell is colored green; a regression is colored red.}
\label{tab:webq_s5_comprehensive}
\resizebox{\textwidth}{!}{
\begin{tabular}{lcc cc cc cc cc cc cc cc}
\toprule
\textbf{Model} & \multicolumn{2}{c}{\textbf{Accuracy (\%)}} & \multicolumn{2}{c}{\textbf{HPSR (\%)}} & \multicolumn{2}{c}{\textbf{PopGap}} & \multicolumn{2}{c}{\textbf{Spearman}} & \multicolumn{2}{c}{\textbf{Confidence}} & \multicolumn{2}{c}{\textbf{Alignment}} & \multicolumn{2}{c}{\textbf{Spearman Conf Pop}} & \multicolumn{2}{c}{\textbf{Spearman Align Pop}} \\
\cmidrule(lr){2-3} \cmidrule(lr){2-3} \cmidrule(lr){4-5} \cmidrule(lr){6-7} \cmidrule(lr){8-9} \cmidrule(lr){10-11} \cmidrule(lr){12-13} \cmidrule(lr){14-15} \cmidrule(lr){16-17}
 & \textbf{Before} & \textbf{After} & \textbf{Before} & \textbf{After} & \textbf{Before} & \textbf{After} & \textbf{Before} & \textbf{After} & \textbf{Before} & \textbf{After} & \textbf{Before} & \textbf{After} & \textbf{Before} & \textbf{After} & \textbf{Before} & \textbf{After} \\
\midrule
\texttt{Falcon-7b-instruct} & 23.8 & \textcolor{green}{29.5} & 46.6 & \textcolor{green}{33.2} & -0.037 & \textcolor{red}{-0.153} & 0.195 & \textcolor{red}{0.265} & 0.390 & 0.401 & 0.564 & \textcolor{red}{0.558} & 0.062 & \textcolor{green}{0.062} & -0.171 & \textcolor{red}{-0.298} \\
\texttt{Qwen2.5-1.5B} & 53.0 & \textcolor{green}{56.0} & 60.0 & \textcolor{green}{54.6} & 0.006 & \textcolor{red}{-0.049} & 0.176 & \textcolor{red}{0.277} & 0.644 & 0.626 & 0.609 & \textcolor{red}{0.605} & 0.165 & \textcolor{red}{0.166} & 0.049 & \textcolor{red}{0.092} \\
\texttt{DeepSeek-V2-Lite} & 52.4 & \textcolor{green}{54.6} & 58.5 & \textcolor{green}{49.9} & -0.014 & \textcolor{red}{-0.099} & 0.208 & \textcolor{red}{0.350} & 0.417 & 0.417 & 0.553 & \textcolor{red}{0.550} & 0.217 & \textcolor{red}{0.230} & -0.129 & \textcolor{red}{-0.232} \\
\bottomrule
\end{tabular}}
\end{table*}

\begin{table*}[h!]
\centering
\caption{Comprehensive debiasing results for \textbf{WebQ: S6 None of the Above}. An improvement in an 'After' cell is colored green; a regression is colored red.}
\label{tab:webq_s6_comprehensive}
\resizebox{\textwidth}{!}{
\begin{tabular}{lcc cc cc cc cc cc cc cc}
\toprule
\textbf{Model} & \multicolumn{2}{c}{\textbf{Accuracy (\%)}} & \multicolumn{2}{c}{\textbf{HPSR (\%)}} & \multicolumn{2}{c}{\textbf{PopGap}} & \multicolumn{2}{c}{\textbf{Spearman}} & \multicolumn{2}{c}{\textbf{Confidence}} & \multicolumn{2}{c}{\textbf{Alignment}} & \multicolumn{2}{c}{\textbf{Spearman Conf Pop}} & \multicolumn{2}{c}{\textbf{Spearman Align Pop}} \\
\cmidrule(lr){2-3} \cmidrule(lr){2-3} \cmidrule(lr){4-5} \cmidrule(lr){6-7} \cmidrule(lr){8-9} \cmidrule(lr){10-11} \cmidrule(lr){12-13} \cmidrule(lr){14-15} \cmidrule(lr){16-17}
 & \textbf{Before} & \textbf{After} & \textbf{Before} & \textbf{After} & \textbf{Before} & \textbf{After} & \textbf{Before} & \textbf{After} & \textbf{Before} & \textbf{After} & \textbf{Before} & \textbf{After} & \textbf{Before} & \textbf{After} & \textbf{Before} & \textbf{After} \\
\midrule
\texttt{Llama-3.1-8B} & 16.1 & \textcolor{green}{31.6} & 56.3 & \textcolor{green}{27.4} & 0.412 & \textcolor{green}{0.039} & -0.644 & \textcolor{red}{-0.793} & 0.672 & 0.530 & 0.343 & \textcolor{green}{0.469} & 0.365 & \textcolor{green}{0.129} & -0.437 & \textcolor{green}{-0.091} \\
\texttt{zephyr-7b-beta} & 20.8 & \textcolor{green}{40.9} & 54.7 & \textcolor{green}{26.4} & 0.405 & \textcolor{green}{0.037} & -0.713 & \textcolor{red}{-0.854} & 0.862 & 0.672 & 0.277 & \textcolor{green}{0.443} & 0.260 & \textcolor{green}{0.063} & -0.592 & \textcolor{green}{-0.436} \\
\texttt{phi-1\_5} & 18.1 & \textcolor{green}{40.9} & 53.3 & \textcolor{green}{23.1} & 0.369 & \textcolor{green}{0.030} & -0.669 & \textcolor{red}{-0.859} & 0.409 & 0.394 & 0.537 & \textcolor{red}{0.513} & 0.248 & \textcolor{green}{0.063} & 0.281 & \textcolor{red}{0.582} \\
\bottomrule
\end{tabular}}
\end{table*}


\section{Calibration Analysis}
\label{app:calibration}

\begin{figure*}[t]
\centering
\begin{subfigure}[b]{0.48\textwidth}
\includegraphics[width=\textwidth]{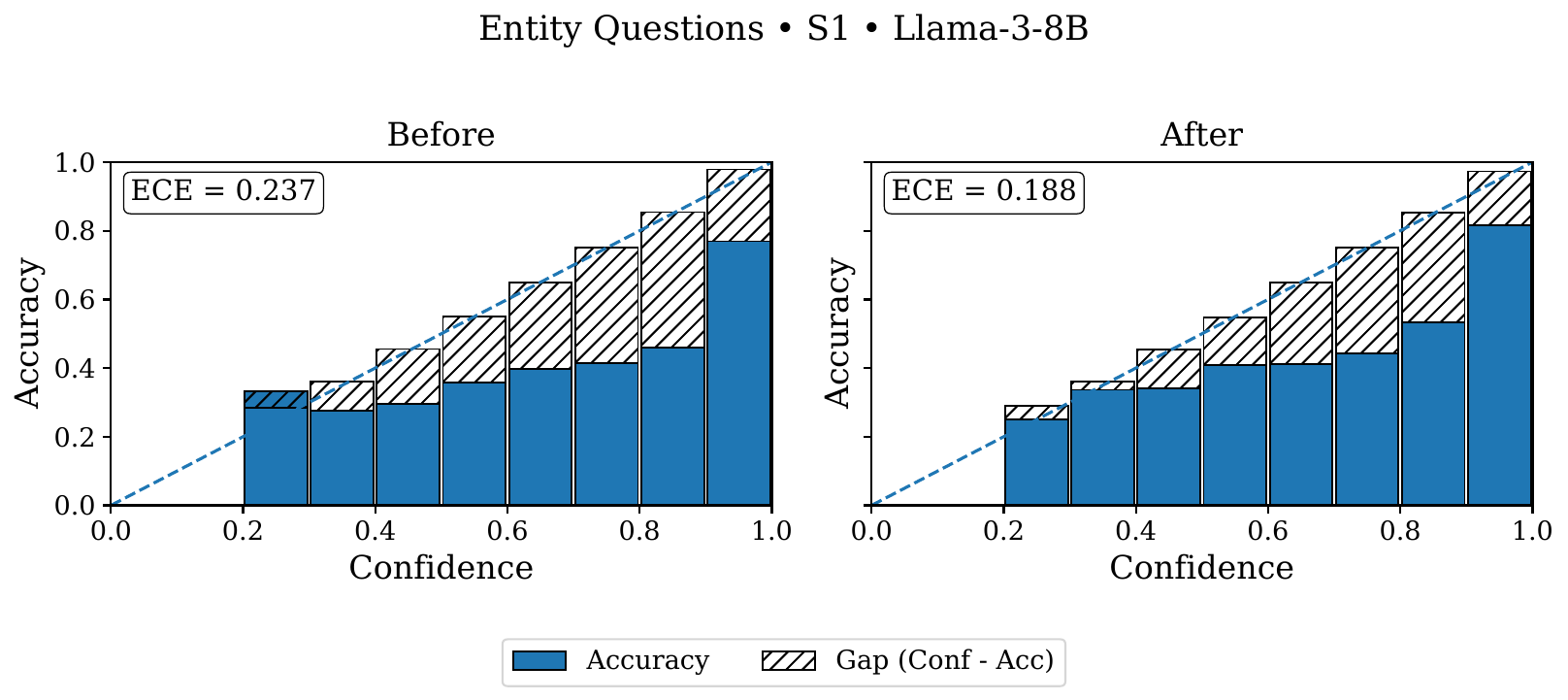}
\caption{Llama-3-8B: ECE 0.237 → 0.188}
\end{subfigure}
\begin{subfigure}[b]{0.48\textwidth}
\includegraphics[width=\textwidth]{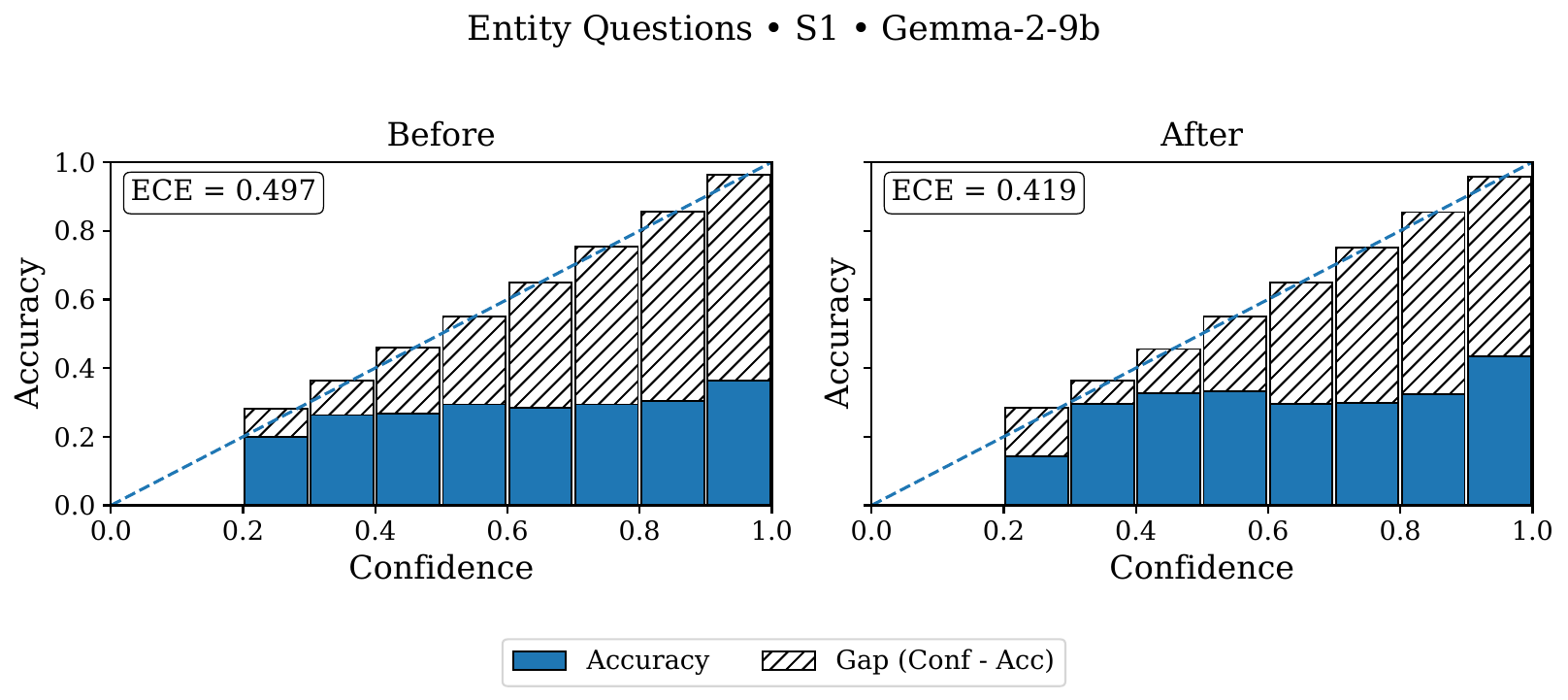}
\caption{Gemma-2-9b: ECE 0.497 → 0.419}
\end{subfigure}
\caption{\textbf{S1 (Baseline) reliability diagrams.}}
\label{fig:reliability_s1}
\end{figure*}

\begin{figure*}[t]
\centering
\begin{subfigure}[b]{0.48\textwidth}
\includegraphics[width=\textwidth]{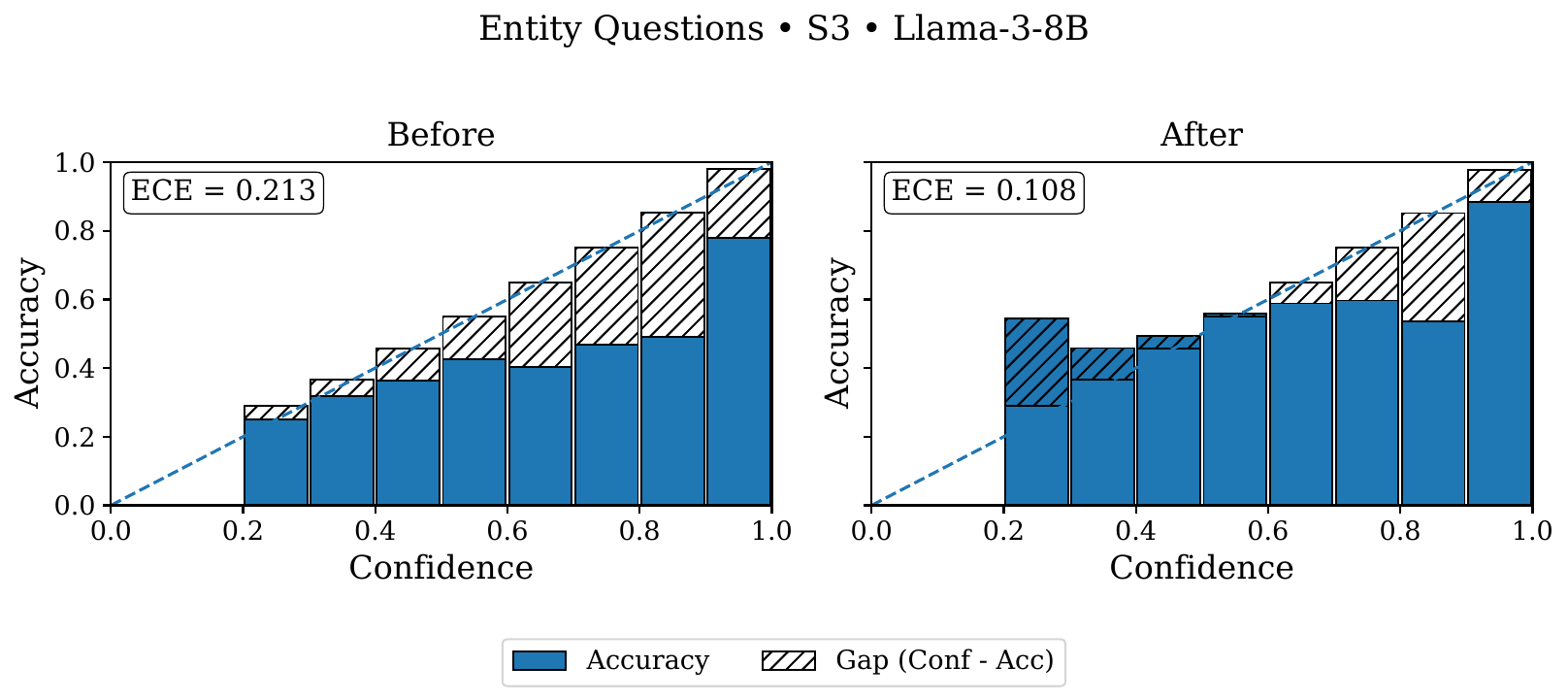}
\caption{Llama-3-8B: ECE 0.213 → 0.108}
\end{subfigure}
\begin{subfigure}[b]{0.48\textwidth}
\includegraphics[width=\textwidth]{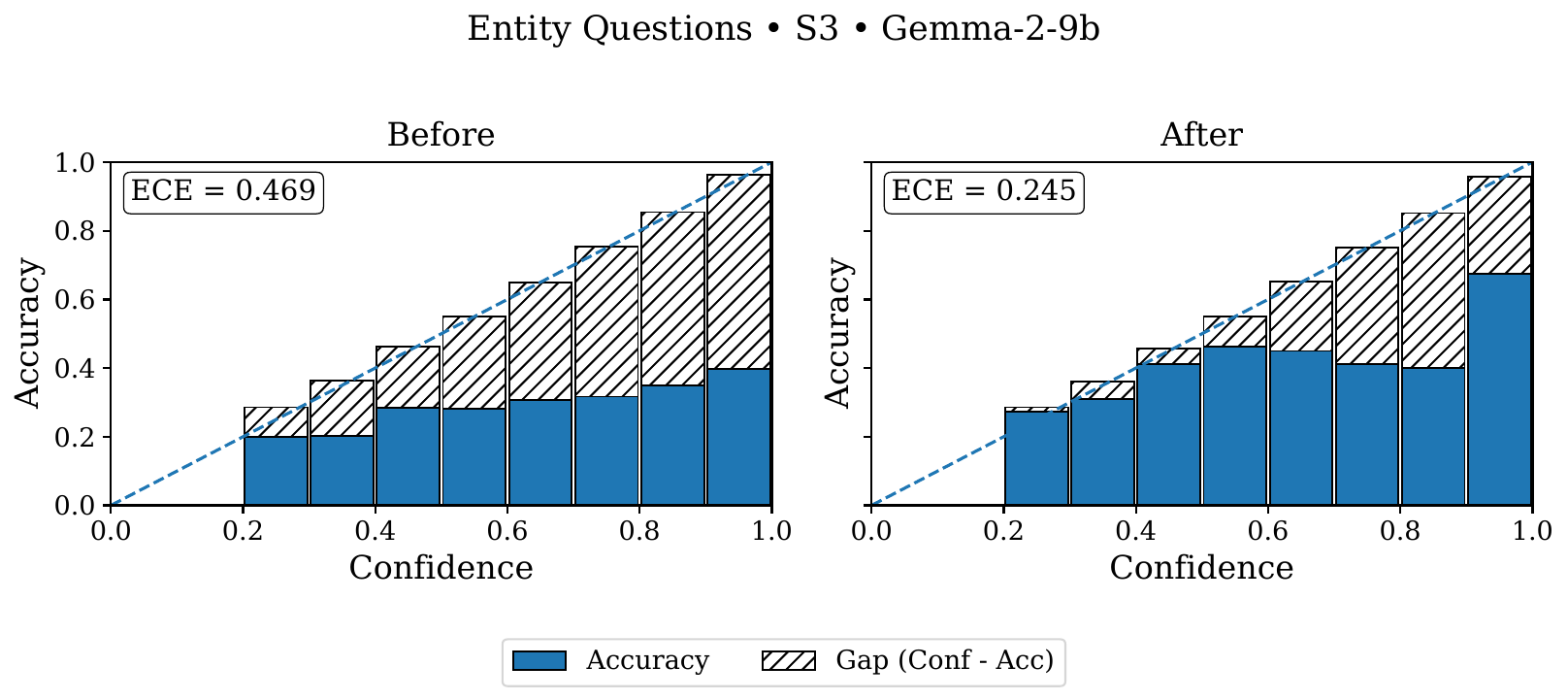}
\caption{Gemma-2-9b: ECE 0.469 → 0.245}
\end{subfigure}
\caption{\textbf{S3 (Gradient) reliability diagrams.}}
\label{fig:reliability_s3}
\end{figure*}

\begin{figure*}[t]
\centering
\begin{subfigure}[b]{0.48\textwidth}
\includegraphics[width=\textwidth]{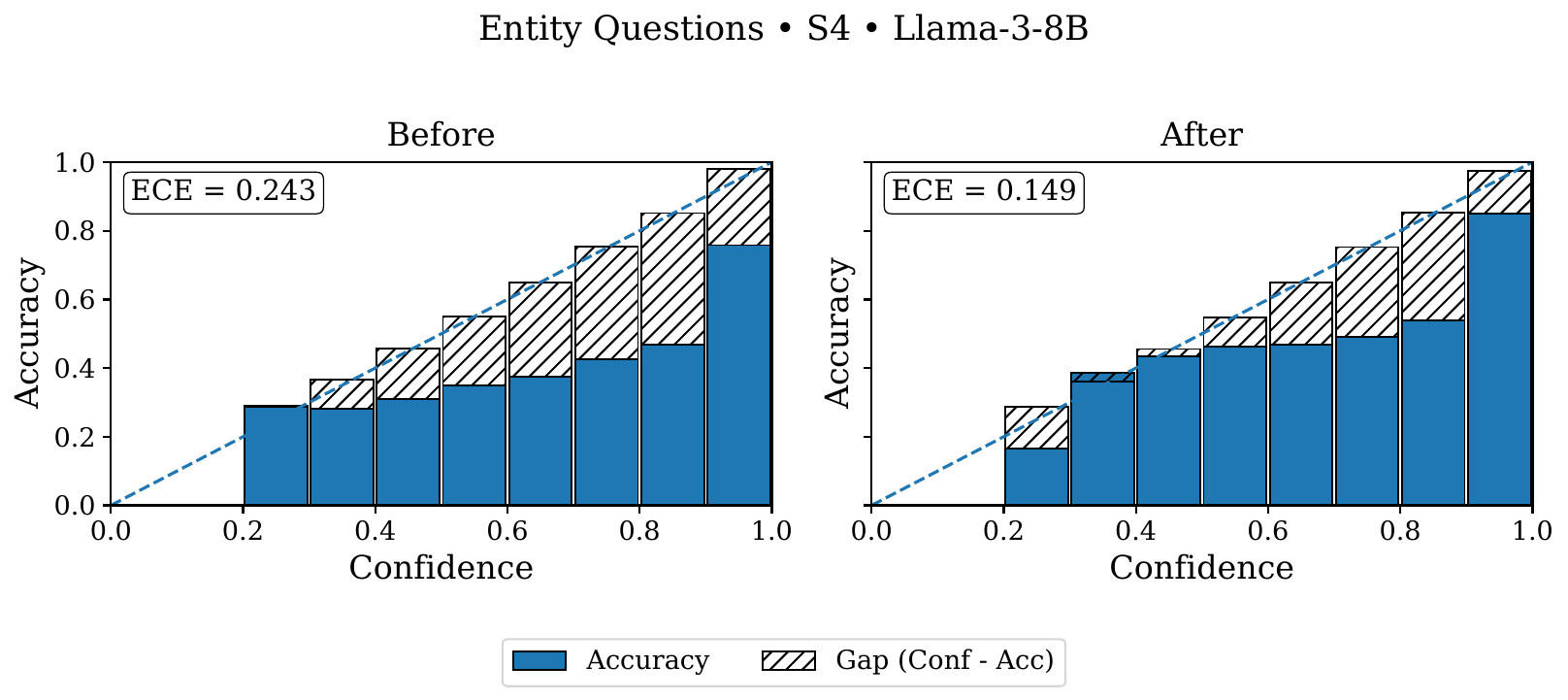}
\caption{Llama-3-8B: ECE 0.243 → 0.149}
\end{subfigure}
\begin{subfigure}[b]{0.48\textwidth}
\includegraphics[width=\textwidth]{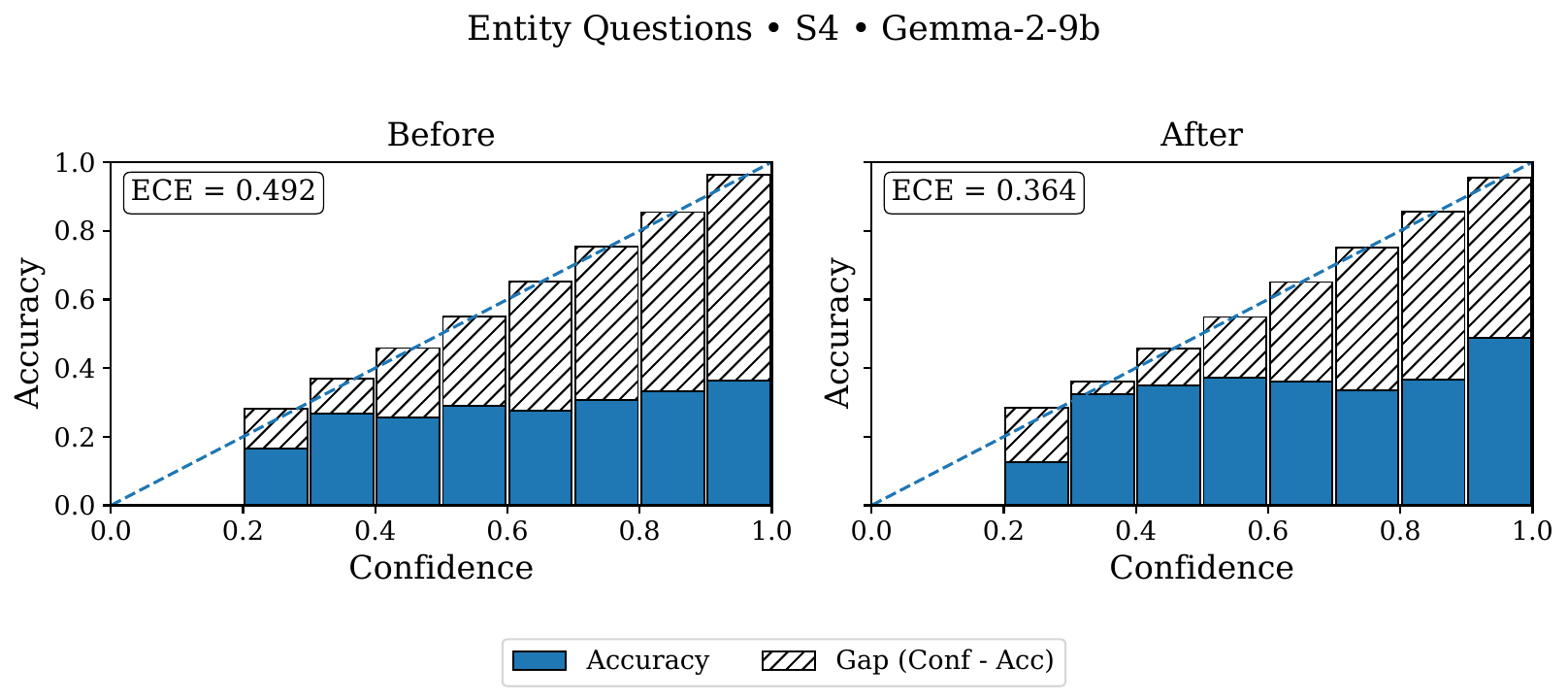}
\caption{Gemma-2-9b: ECE 0.492 → 0.364}
\end{subfigure}
\caption{\textbf{S4 (Direct Contest) reliability diagrams.}}
\label{fig:reliability_s4}
\end{figure*}

\begin{figure*}[t]
\centering
\begin{subfigure}[b]{0.48\textwidth}
\includegraphics[width=\textwidth]{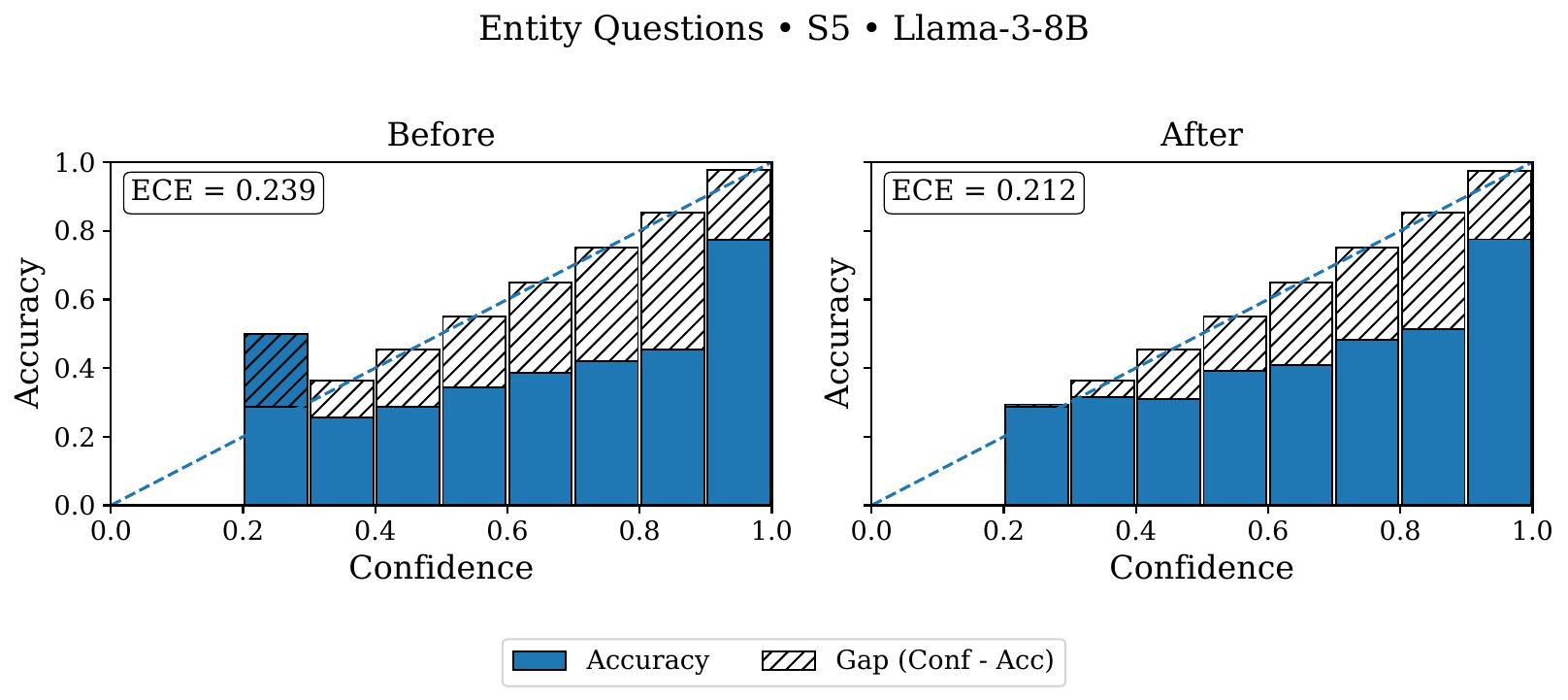}
\caption{Llama-3-8B: ECE 0.239 → 0.212}
\end{subfigure}
\begin{subfigure}[b]{0.48\textwidth}
\includegraphics[width=\textwidth]{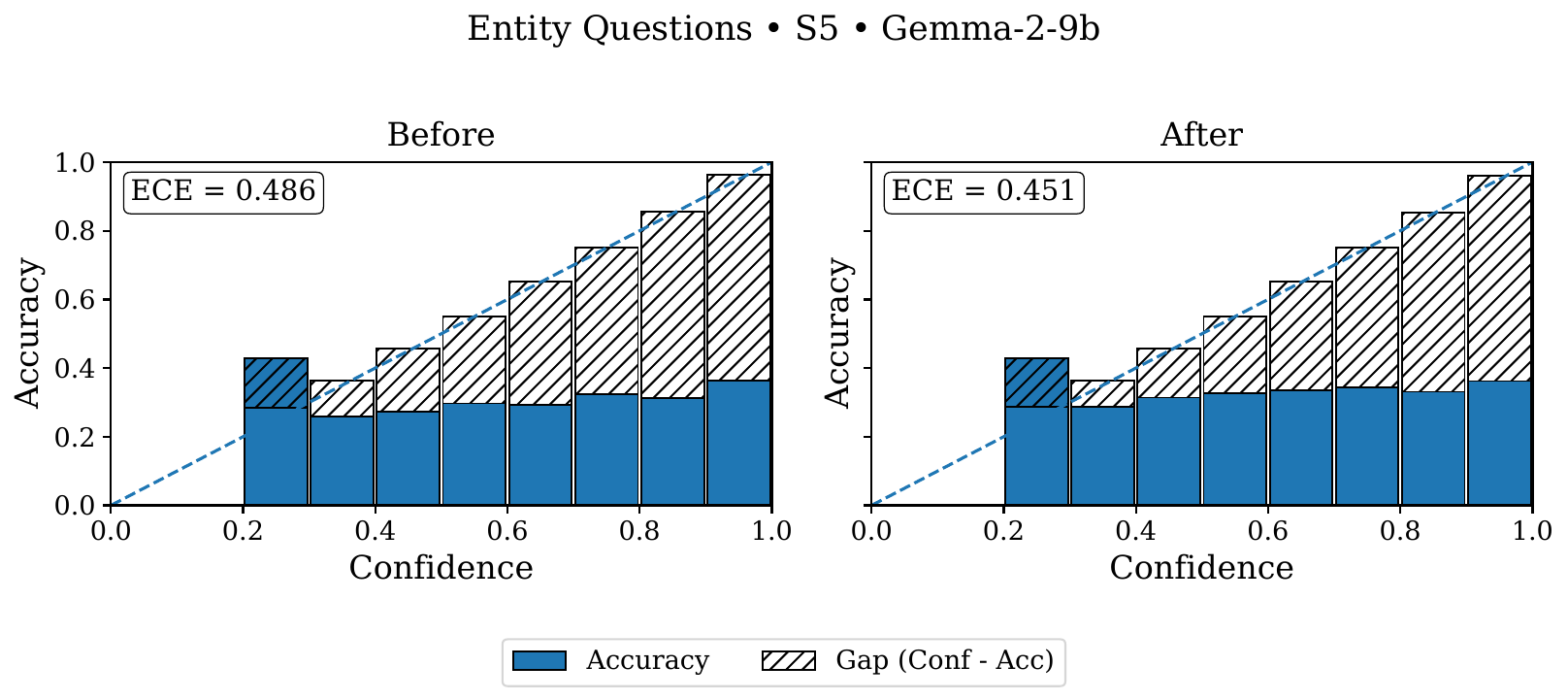}
\caption{Gemma-2-9b: ECE 0.486 → 0.451}
\end{subfigure}
\caption{\textbf{S5 (Reverse Control) reliability diagrams.} Minimal ECE reduction
validates that PopDebias targets popularity-induced overconfidence specifically.}
\label{fig:reliability_s5}
\end{figure*}

\begin{figure*}[t]
\centering
\begin{subfigure}[b]{0.48\textwidth}
\includegraphics[width=\textwidth]{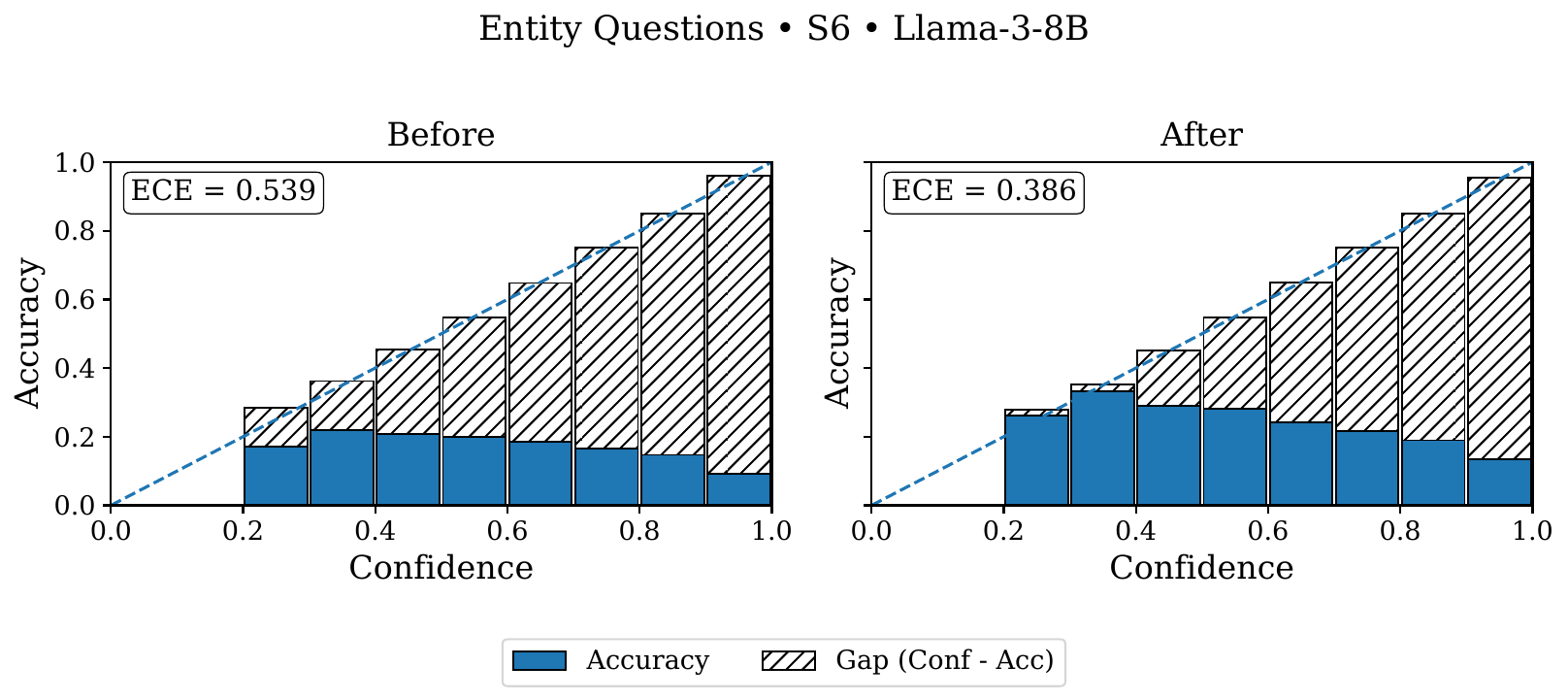}
\caption{Llama-3-8B: ECE 0.539 → 0.386}
\end{subfigure}
\begin{subfigure}[b]{0.48\textwidth}
\includegraphics[width=\textwidth]{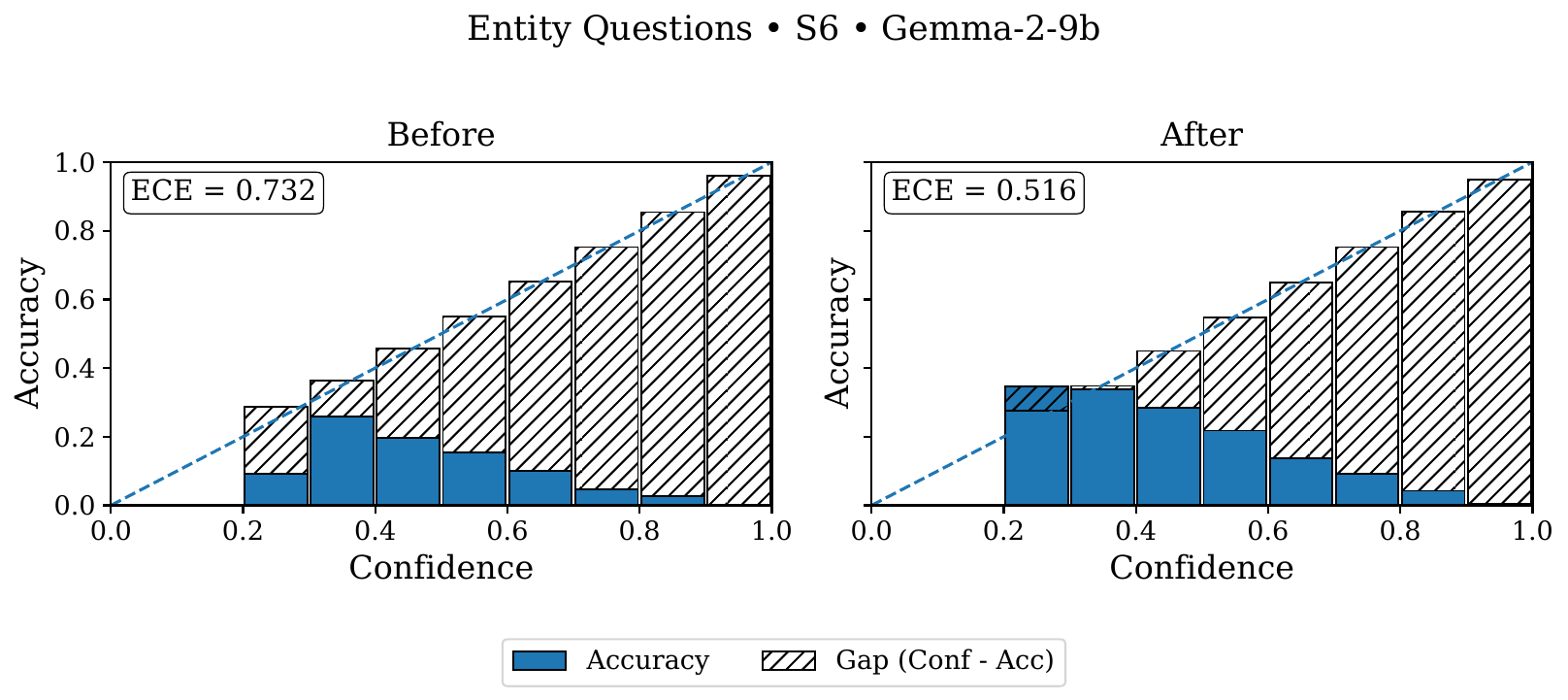}
\caption{Gemma-2-9b: ECE 0.732 → 0.516}
\end{subfigure}
\caption{\textbf{S6 (None of the Above) reliability diagrams.} Despite the highest
initial ECE, PopDebias still achieves 28–30\% calibration improvement.}
\label{fig:reliability_s6}
\end{figure*}

Figures~\ref{fig:reliability_s1}--\ref{fig:reliability_s6} show reliability diagrams
across all six strategies on EntityQuestions.

\paragraph{S1 (Baseline).} Even with random distractors, models exhibit moderate
overconfidence (ECE = 0.237–0.497); PopDebias reduces ECE by 16–21\%.

\paragraph{S2 (Popular Trap).} The most severe miscalibration; 60–67\% ECE reduction
after debiasing.

\paragraph{S3 (Gradient) and S4 (Direct Contest).} 48–49\% and 26–39\% ECE reduction,
respectively.

\paragraph{S5 (Reverse Control).} Smallest ECE reduction (7–11\%) when correct answers
are more popular than distractors, validating that PopDebias primarily corrects
popularity-induced overconfidence, not general confidence.

\paragraph{S6 (None of the Above).} Despite the challenge of epistemic uncertainty,
PopDebias reduces ECE by 28–30\%; however, absolute ECE remains high (0.386–0.516),
indicating that abstention requires mechanisms beyond popularity debiasing.

\section{Full Debiasing Results (All 22 Models)}
\label{sec:full_results_for_Result}

Table~\ref{tab:comprehensive_all_strategies_full} presents complete
``Before $\to$ After'' results for all 22 models across all four datasets and all six
strategies.  Asterisks~(*) denote statistically significant accuracy improvements
($p<0.05$, McNemar's test).

\paragraph{Key findings.}
(1)~\textbf{Universal improvement:} PopDebias improves all 22 models under adversarial
conditions (S2--S4, S6).  (2)~\textbf{Largest gains under S2:} average $+24.7$~pp across all four datasets,
with individual gains up to $+54.1$~pp.  (3)~\textbf{No harm under S5:} performance
is maintained or slightly improved when popularity aligns with truth.
(4)~\textbf{Consistent \HPSR{} reduction} across all conditions.
(5)~\textbf{Scale-independent:} benefits appear from 0.5B to 32B parameters.

\begin{table*}[ht]
\centering
\caption{\textbf{Comprehensive Debiasing Results Across All Strategies (S1--S6).} 
We report Accuracy (\%), Correlation ($\rho$), and High-Pop Selection Rate (HPSR, \%) 
in ``Before $\to$ After'' format. 
Asterisks (*) denote statistically significant accuracy improvements. Full per-model tables, including confidence and calibration metrics, are reported in Appendix~\ref{sec:full_results_for_Result} (EntityQuestions: Tables~\ref{tab:entityquestions_s2_comprehensive}--\ref{tab:entityquestions_s6_comprehensive}; MuSiQue: Tables~\ref{tab:musique_s1_comprehensive}--\ref{tab:musique_s6_comprehensive}; Natural Questions: Tables~\ref{tab:nq_s1_comprehensive}--\ref{tab:nq_s6_comprehensive}; WebQuestions: Tables~\ref{tab:webq_s1_comprehensive}--\ref{tab:webq_s6_comprehensive}).}
\label{tab:comprehensive_all_strategies_full}
\resizebox{\textwidth}{!}{
\begin{tabular}{l|ccc|ccc|ccc|ccc}
\toprule
& \multicolumn{3}{c|}{\textbf{NQ}} & \multicolumn{3}{c|}{\textbf{EntityQuestions}} & \multicolumn{3}{c|}{\textbf{MuSiQue}} & \multicolumn{3}{c}{\textbf{WebQuestions}} \\
\cmidrule(lr){2-4} \cmidrule(lr){5-7} \cmidrule(lr){8-10} \cmidrule(lr){11-13}
\textbf{Model} & \textbf{Acc} & \textbf{$\rho$} & \textbf{HPSR} & \textbf{Acc} & \textbf{$\rho$} & \textbf{HPSR} & \textbf{Acc} & \textbf{$\rho$} & \textbf{HPSR} & \textbf{Acc} & \textbf{$\rho$} & \textbf{HPSR} \\
\midrule
\multicolumn{13}{c}{\textbf{S1: Baseline}} \\
\midrule
DeepSeek-V2-Lite & 41.7 $\to$ \textbf{54.0*} & -0.32 $\to$ -0.02 & 48.1 $\to$ 22.5 & 43.4 $\to$ \textbf{54.0*} & -0.28 $\to$ 0.01 & 44.4 $\to$ 22.5 & 31.2 $\to$ \textbf{44.4*} & -0.23 $\to$ 0.10 & 49.9 $\to$ 23.3 & 50.0 $\to$ \textbf{56.8*} & -0.34 $\to$ -0.01 & 46.6 $\to$ 28.6 \\
Falcon-7B & 24.8 $\to$ \textbf{40.6*} & -0.33 $\to$ -0.13 & 53.6 $\to$ 26.0 & 26.7 $\to$ \textbf{39.2*} & -0.28 $\to$ -0.06 & 50.6 $\to$ 25.2 & 27.1 $\to$ \textbf{37.3*} & -0.26 $\to$ -0.07 & 48.4 $\to$ 27.3 & 27.8 $\to$ \textbf{42.5*} & -0.26 $\to$ 0.05 & 50.3 $\to$ 23.4 \\
Gemma-2-9B & 32.6 $\to$ 34.5 & -0.33 $\to$ -0.28 & 50.7 $\to$ 46.5 & 32.7 $\to$ \textbf{36.0*} & -0.27 $\to$ -0.23 & 47.7 $\to$ 42.1 & 29.3 $\to$ \textbf{31.7*} & -0.20 $\to$ -0.16 & 48.2 $\to$ 42.8 & 36.0 $\to$ \textbf{38.7*} & -0.31 $\to$ -0.22 & 49.1 $\to$ 42.7 \\
Llama-2-7B & 31.2 $\to$ \textbf{43.4*} & -0.35 $\to$ -0.14 & 51.0 $\to$ 27.6 & 31.7 $\to$ \textbf{43.2*} & -0.29 $\to$ -0.05 & 48.6 $\to$ 24.9 & 29.3 $\to$ \textbf{40.3*} & -0.26 $\to$ -0.05 & 49.7 $\to$ 26.7 & 32.2 $\to$ \textbf{44.6*} & -0.29 $\to$ 0.01 & 49.4 $\to$ 24.6 \\
Llama-3.1-8B & 62.1 $\to$ \textbf{65.0*} & -0.29 $\to$ -0.20 & 37.6 $\to$ 32.7 & 59.2 $\to$ \textbf{63.7*} & -0.25 $\to$ -0.09 & 35.6 $\to$ 27.5 & 42.8 $\to$ \textbf{50.1*} & -0.21 $\to$ -0.08 & 40.8 $\to$ 30.9 & 53.3 $\to$ \textbf{57.8*} & -0.27 $\to$ -0.11 & 45.9 $\to$ 37.5 \\
\bottomrule
\end{tabular}
}

\end{table*}

\end{document}